\pdfoutput=1
\PassOptionsToPackage{backref=page,breaklinks=true}{hyperref}

\documentclass[11pt]{article}

\usepackage{ACL2023}
\usepackage{times}
\usepackage{latexsym}

\usepackage[T1]{fontenc}

\usepackage[utf8]{inputenc}

\usepackage{microtype}

\usepackage{inconsolata}

\usepackage{float}
\usepackage{longtable}
\usepackage{amsfonts}

\usepackage{booktabs}

\usepackage{bbding} 
\usepackage{xspace}
\newcommand{\icon}[2][0.4cm]{\begin{minipage}{#1}\includegraphics[width=#1]{#2}\end{minipage}}
\newcommand{\bfull}[0]{\icon{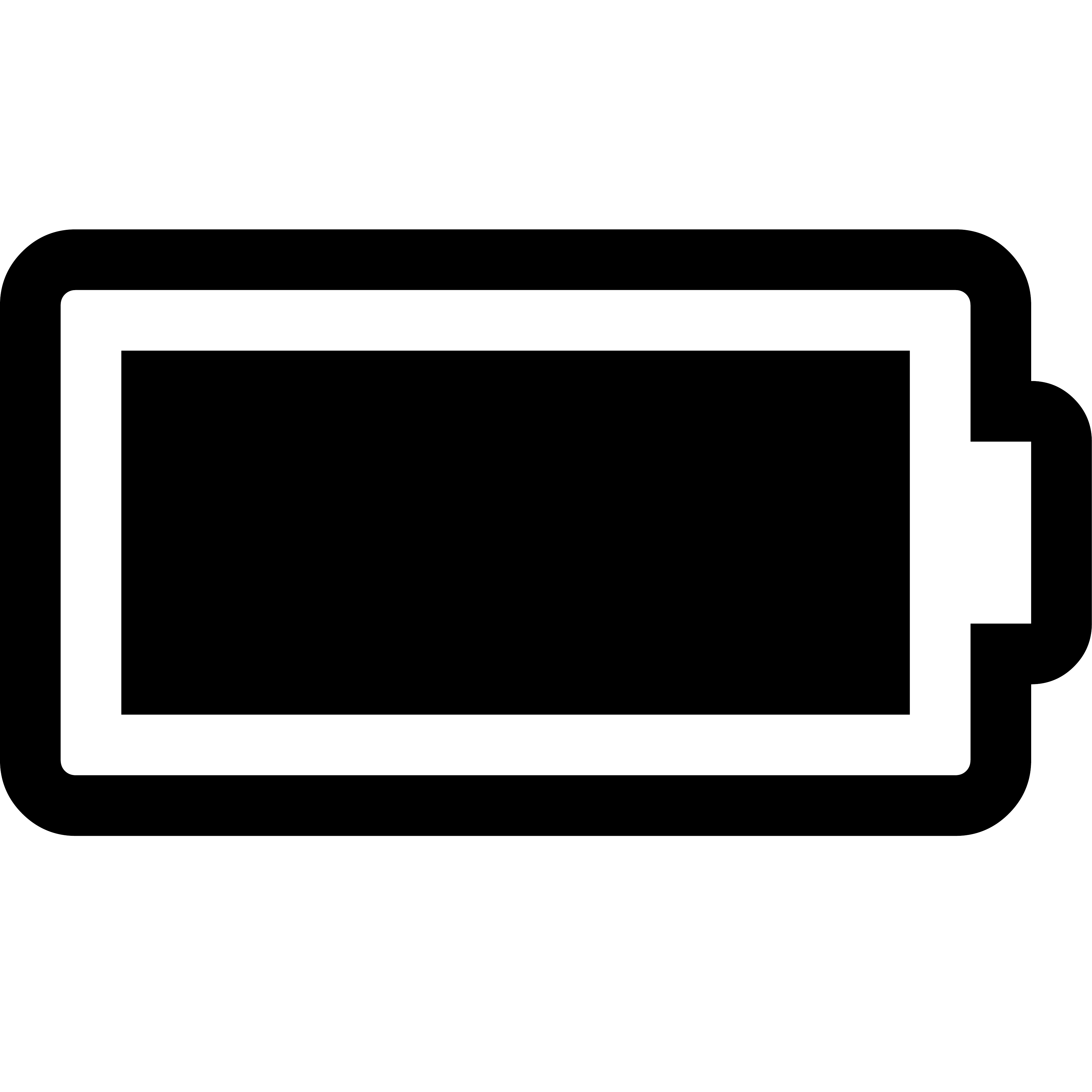}\xspace}
\newcommand{\bempty}[0]{\icon{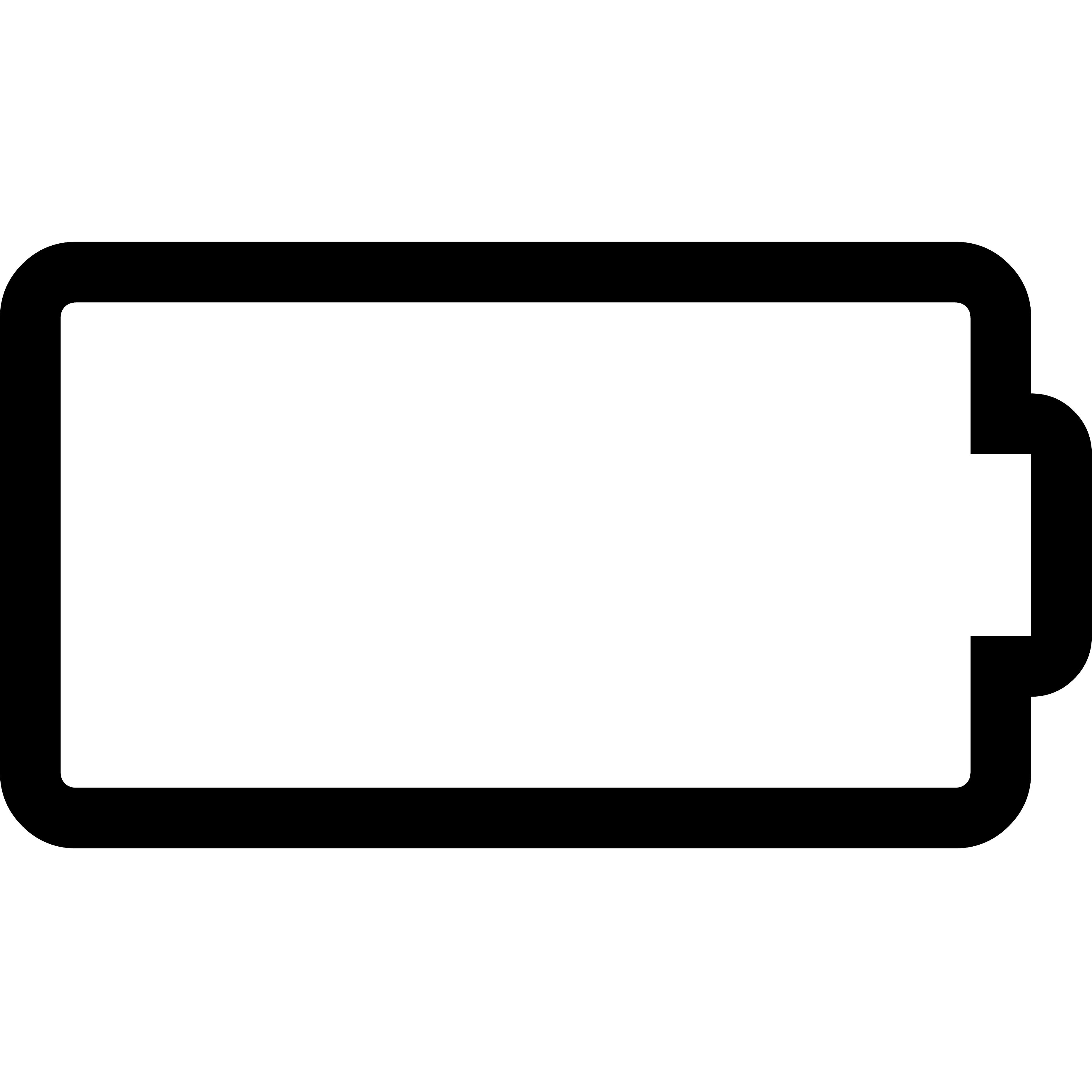}\xspace}
\newcommand{\bhalf}[0]{\icon{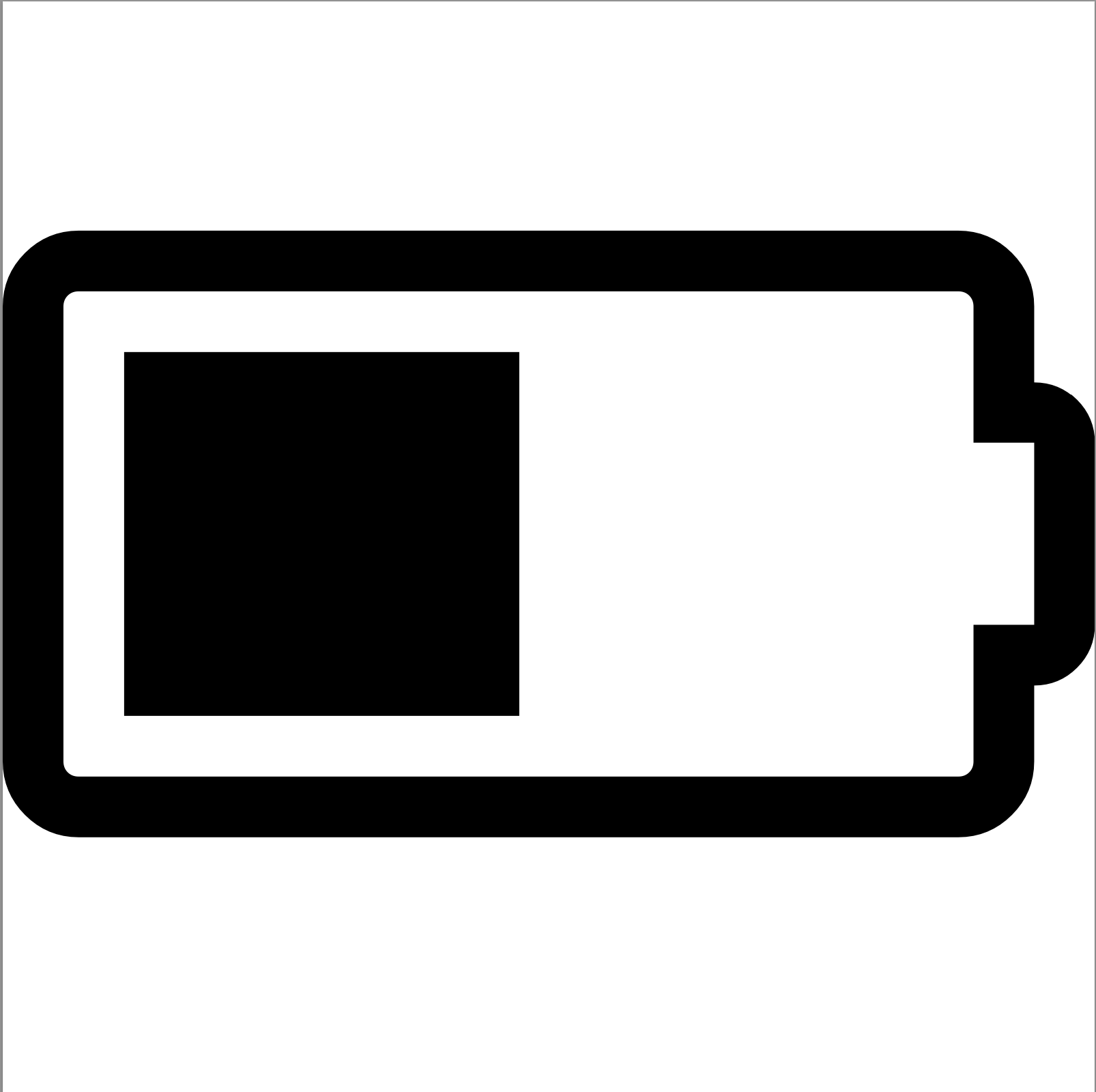}\xspace}

\usepackage{colortbl}
\definecolor{mygray}{gray}{.9}

\usepackage{tikz}
\usepackage[edges]{forest}
\definecolor{hidden-draw}{RGB}{5, 4, 36}
\definecolor{hidden-blue}{RGB}{194,232,247}
\definecolor{hidden-orange}{RGB}{243,202,120}
\definecolor{hidden-yellow}{RGB}{242,244,193}
\definecolor{tree-level-1}{RGB}{245,20,85}
\definecolor{tree-level-2}{RGB}{246,86,118}
\definecolor{tree-level-3}{RGB}{248,177,193}
\definecolor{tree-leaf}{RGB}{176,230,198}
\definecolor{taxonomyColor}{RGB}{200,200,200} 
\newcommand{\taxonomyColorName}[0]{light gray\xspace}

\definecolor{acquisitionColor}{RGB}{243,202,120} 
\newcommand{\acquisitionColorName}[0]{hidden orange\xspace}

\definecolor{analysisColor}{RGB}{173,216,230} 
\newcommand{\analysisColorName}[0]{pale blue\xspace}

\definecolor{Self}{RGB}{255,0,128}
\definecolor{Ensemble}{RGB}{0,127,255}
\definecolor{Iterative}{RGB}{153,51,255}

\definecolor{exemplar1}{RGB}{136,98,148}
\definecolor{exemplar2}{RGB}{148,210,242}
\definecolor{knowledge1}{RGB}{249,219,152}
\definecolor{knowledge2}{RGB}{255,245,220}

\usetikzlibrary{arrows.meta, positioning}

\usepackage{subcaption}
\usepackage{pgfplots}
\pgfplotsset{compat=1.17}
\usetikzlibrary{pgfplots.groupplots}

\usepackage{amsmath}
\usepackage{footnote}
\newcommand{\CFD}[1]{\textcolor{blue}{#1}}

\newcommand{\SmallHeading}[1]{\noindent\textbf{#1.}\quad}
\newcommand{\SmallHeadingQuestion}[1]{\noindent\textbf{#1?}\quad}

\usepackage{changepage}
\usepackage{enumitem}
\usepackage{mathtools}

\usepackage{kantlipsum}
\usepackage[breakable, theorems, skins]{tcolorbox}
\tcbset{enhanced}
\DeclareRobustCommand{\mybox}[2][gray!20]{%
\begin{tcolorbox}[   
        breakable,
        left=0pt,
        right=0pt,
        top=0pt,
        bottom=0pt,
        colback=#1,
        colframe=#1,
        width=\dimexpr0.39\textwidth\relax, 
        enlarge left by=0mm,
        boxsep=5pt,
        arc=0pt,outer arc=0pt,
        ]
        #2
\end{tcolorbox}
}

\usepackage{array}

\usepackage[framemethod=TikZ]{mdframed}
\usepackage[framemethod=TikZ]{mdframed}

\usepackage{xcolor}
\usepackage{lipsum} 

\mdfdefinestyle{examplestyle}{
  linecolor=blue,           
  outerlinewidth=1pt,       
  roundcorner=10pt,         
  innertopmargin=10pt,      
  innerbottommargin=10pt,   
  innerrightmargin=5pt,    
  innerleftmargin=5pt,     
  backgroundcolor=gray!10,  
  shadow=false,             
  shadowsize=6pt,           
  shadowcolor=black!20,     
  frametitlerule=true,      
  frametitlerulewidth=1pt,  
  frametitlerulecolor=blue, 
  frametitlebackgroundcolor=gray!20, 
  frametitlefont=\bfseries, 
  frametitlealignment=\centering 
}
\newmdenv[style=examplestyle]{exampleblock}

\usepackage{adjustbox}

\usepackage{textcomp}  
\usepackage{scalerel}  

\newcommand{\thumbsup}{\scalerel*{\includegraphics{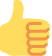}}{\textrm{\textbigcircle}}\xspace}

\newcommand{\Appendix}{App.\xspace}

\definecolor{starcolor}{RGB}{255,215,0} 

\newcommand{\solidstar}{\tikz{\node[star, star point ratio=2.25, fill=starcolor, scale=0.5] {};}}

\newcommand{\hollowstar}{\tikz{\node[star, star point ratio=2.25, draw=starcolor, scale=0.5] {};}}

\newcommand{\onestar}{\solidstar\hollowstar\hollowstar\hollowstar\hollowstar}
\newcommand{\twostars}{\solidstar\solidstar\hollowstar\hollowstar\hollowstar}
\newcommand{\threestars}{\solidstar\solidstar\solidstar\hollowstar\hollowstar}
\newcommand{\fourstars}{\solidstar\solidstar\solidstar\solidstar\hollowstar}
\newcommand{\fivestars}{\solidstar\solidstar\solidstar\solidstar\solidstar}

\usepackage{soul}
\usepackage{cleveref}
\crefname{figure}{Figure}{Figures}
\crefname{table}{Table}{Tables}
\crefname{appendix}{App}{Apps}
\crefname{section}{Section}{Sections}
\crefname{equation}{Eq.}{Eqs.}

\usepackage{pgfplots}
\usepgfplotslibrary{fillbetween}
\usetikzlibrary{patterns, arrows.meta} 

\title{The Odyssey of Commonsense Causality: \\From Foundational Benchmarks to Cutting-Edge Reasoning}

\author{Shaobo Cui \\
  EPFL, Switzerland \\
  \texttt{shaobo.cui@epfl.ch} \\\And
  Zhijing Jin \\
  MPI \& ETH Zürich \\
  \texttt{jinzhi@ethz.ch} \\\And
  Bernhard Sch{\"o}lkopf \\
  MPI \& ETH Zürich \\
  \texttt{bs@tue.mpg.de} \\\And
  Boi Faltings \\
  EPFL, Switzerland\\ 
  \texttt{boi.faltings@epfl.ch}
   \\
  }

\begin{document}

\maketitle

\begin{abstract}
Understanding commonsense causality is a unique mark of intelligence for humans. It helps people understand the principles of the real world better and benefits the decision-making process related to causation. For instance, commonsense causality is crucial in judging whether a defendant's action causes the plaintiff's loss in determining legal liability. Despite its significance, a systematic exploration of this topic is notably lacking. Our comprehensive survey bridges this gap by focusing on taxonomies, benchmarks, acquisition methods, qualitative reasoning, and quantitative measurements in commonsense causality, synthesizing insights from over 200 representative articles. Our work aims to provide a systematic overview, update scholars on recent advancements, provide a pragmatic guide for beginners, and highlight promising future research directions in this vital field. 
\end{abstract}

\section{Introduction} \label{sec:introduction}
\begin{quote}
    \textit{We do not have knowledge of a thing until we have grasped its why, that is to say, its cause.}
    \hspace*{\fill}--- \textit{Aristotle, 384--322 BC}
\end{quote}
Causality~\cite{fisher-1936-design,rubin-1974-estimating,holland-1986-statistics,granger-1988-some,pearl-2009-causality,pearl-mackenzie-2018-book} has been a cornerstone concept spanning both scientific and philosophical spheres since Aristotle's era~\cite{hocutt-1974-aristotle}. 
Commonsense causality encapsulates our intuition of how the occurrence of one event, fact, process, state, or object~(the cause) plays a role in bringing about or contributing to the happening of another event, fact, process, state, or object~(the effect).
For example, we know that a rainy morning precipitates traffic congestion or that eating too much leads to weight gain. 
This innate comprehension of cause-and-effect dynamics is frequently termed ``commonsense causality''. 
It has applications across fields such as medical diagnosis~\cite{richens-etal-2020-improving}, psychology~\cite{matute-etal-2015-illusions,eronen-2020-causal}, behavioral science~\cite{grunbaum-1952-causality}, economics~\cite{bronfenbrenner-1981-causality,hoover-2006-causality}, and legal systems~\cite{williams-1961-causation,summers-2018-common} ~(see more applications in \Appendix~\ref{appendix:applications}).

\begin{figure}
\centering
\hspace{-1.5cm}
\resizebox{1.09\linewidth}{!}{
\input{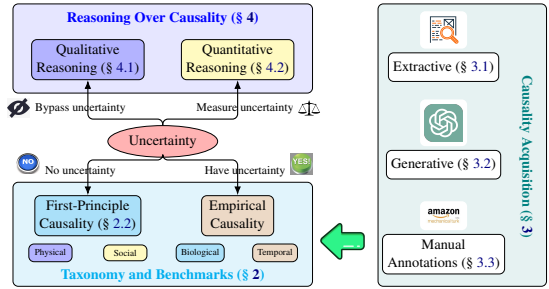}
}
\caption{Different aspects of commonsense causality and their link to different sections of this survey. }
\label{fig:intro:example}
\end{figure}
\begin{figure*}[ht!]
    \hspace{-3em}
    \centering
    \resizebox{1.03\textwidth}{!}{
    \input{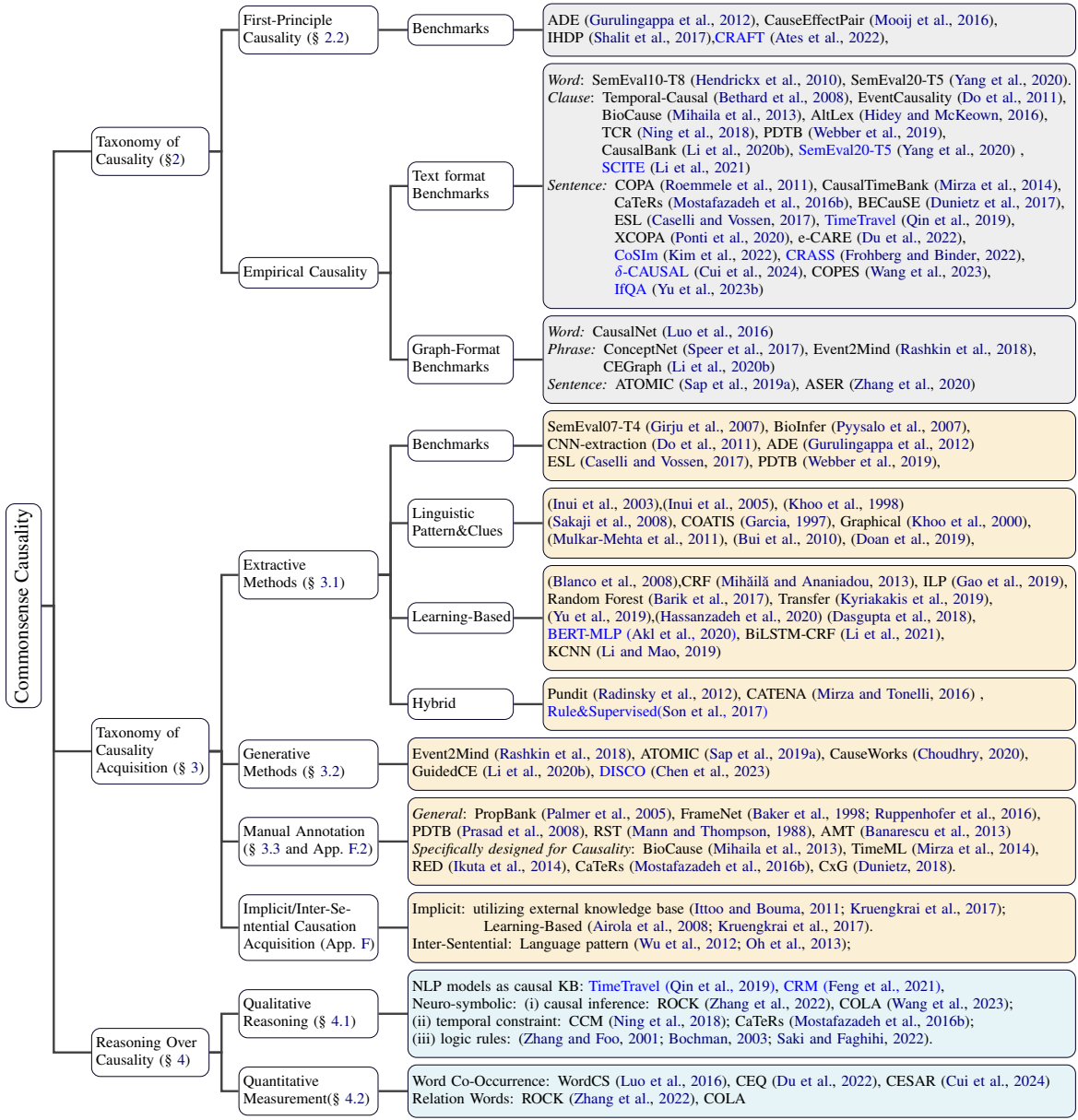}
    }
    \caption{Taxonomy of commonsense causality in various aspects. The benchmarks, datasets, and methods in \CFD{blue color} are about counterfactual. Leaf nodes with different colors are associated with different sections of this survey. }
    \label{fig:categorization_full}
\end{figure*}

Despite its significance, the field still lacks a comprehensive overview of commonsense causality. 
While there are several survey papers on causal inference~\cite{yao-etal-2021-survey,zeng-wang-2022-survey,feder-etal-2022-causal} and commonsense knowledge~\cite{storks-etal-2019-commonsense, bhargava-ng-2022-commonsense}, a comprehensive overview of the intersection of these two domains~---~commonsense causality~---~remains missing. 
The importance of this gap has been further highlighted by recent advancements in large language models (LLMs)~\cite{openai-etal-2023-gpt4,touvron-etal-2023-llama}, which underscore commonsense causality as a pivotal reasoning capability for models. This emerging focus accentuates the urgent need for an in-depth overview.
To fill this blank, we conduct an extensive and up-to-date survey of commonsense causality, with comprehensive coverage of its taxonomy, benchmarks, acquisition methods, as well as qualitative and quantitative reasoning approaches.

We start by presenting a taxonomy of commonsense causality based on different types of commonsense knowledge~(e.g., physical, social, biological, and temporal commonsense) and different levels of uncertainty~(\S~\ref{sec:taxonomy}). 
Leveraging this taxonomy, we methodically categorize 37 existing benchmarks to provide a structured overview. 
Following this, we discuss three main approaches to acquiring benchmarks conducive to commonsense causality research: 
extractive~(\S~\ref{sec:acquisition:extraction}), generative~(\S~\ref{sec:acquisition:generative}), and manual annotation methods~(\S~\ref{sec:acquisition:manual}). Beyond introducing each approach, we also systematically compare the merits and demerits of these three approaches, providing insights for future work on commonsense causality acquisition.

Furthermore, we classify the existing causality reasoning methods into two categories based on their way of managing the intrinsic uncertainty within commonsense causality. The first type is qualitative approaches~(\S~\ref{sec:analysis:qualitative}), which simplify causal reasoning as a classification task and bypass the uncertainty. The second type is quantitative approaches~(\S~\ref{sec:analysis:quantitative}), which employ metrics to measure causal strength, thereby quantifying the uncertainty. 
This classification not only aids in understanding the diverse methodologies but also highlights the varied strategies employed to tackle uncertainty in commonsense causality reasoning.

Lastly, we suggest several promising directions in the field of commonsense causality in \S~\ref{sec:future}. These topics include the exploration of contextual nuances, the analysis of complex structures, the measurement of probabilistic causality, the understanding of temporal dynamics, and the integration of multimodal data. This exploration aims to offer a roadmap for future research.

The contributions of our survey are threefold: 
\begin{itemize}
    \item We present the first comprehensive overview of commonsense causality, synthesizing insights from over 200 representative papers to provide a broad perspective on this topic. 
    \item We methodically review existing benchmarks, acquisition approaches, and reasoning methods by establishing an overall taxonomy, thus offering a useful road map for this field. 
    \item We propose potential research directions for future works and provide a pragmatic handbook for researchers, along with substantial appendices covering a wide range of related topics and preliminary knowledge.\footnote{
    Due to the page limit, we present a main overview of commonsense causality research in the main text. We also provide extensive supplementary information in~\cref{appendix:applications,appendix:background,appendix:survey,appendix:taxonomy,appendix:uncertainty,appendix:acquisition,appendix:detailed_benchmarks,appendix:lm,appendix:concepts,appendix:linguistic,appendix:causal_inference,appendix:handbook},
covering 
applications, 
preliminary knowledge, 
related survey works, 
other taxonomies, 
details of uncertainty, acquisition methods, and benchmarks, concepts of causality,
NLP techniques, 
linguistic causality, 
causal inference, 
and handbook for beginners.
} 
\end{itemize}

\SmallHeading{Paper Selection} Our review focused on articles related to commonsense causality from leading peer-reviewed venues in NLP and AI research, such as ACL, EMNLP, NAACL, AAAI, NeurIPS, ICLR, ICML, and IJCAI. We utilized a keyword-based selection strategy, prioritizing papers featuring terms like "causality", "acquisition", "causal reasoning", and "commonsense" in their titles or abstracts. Additionally, we explored GitHub repositories related to causal NLP papers to complement our search. There are also some papers from the philosophy community that help illustrate the concepts related to causality. 

\SmallHeading{The Scope of This Survey} Determining the precise end line of the scope for this survey presents a significant challenge: the domain of commonsense reasoning encompasses a vast area, within which causality plays a crucial role across a substantial portion. Nevertheless, each dataset and reasoning methods covered in this survey explicitly incorporates the concept of causality and commonsense, either through its designation or its inherent characteristics. Exclusions are made for datasets that focus on non-causal reasoning, such as Social Chemistry 101~\cite{forbes-etal-2020-social}, datasets pertaining to generic logical reasoning (e.g., ProofWriter), among others, which constitute a separate category.

\section{Taxonomy and Benchmarks} \label{sec:taxonomy}
\newcommand{\physicalCSC}[0]{PhysC\xspace}
\newcommand{\socialCSC}[0]{SocC\xspace}
\newcommand{\temporalCSC}[0]{TempC\xspace}
\newcommand{\biologicalCSC}[0]{BioC\xspace}
\newcommand{\economicCSC}[0]{EconC\xspace}
\newcommand{\generalCSC}[0]{*\xspace}

\begin{table}[!ht]
  \centering
  \resizebox{\columnwidth}{!}{
    \begin{tabular}{lp{1.5cm}lp{1.5cm}p{0.5cm}p{1.0cm}}
    \toprule
     Benchmarks & Annotation Unit & \#Overall & \#Causal  & C.F.\footnotemark[1]  & Type \\
    \midrule
    \rowcolor{mygray} \multicolumn{6}{c}{\textit{First-principle causality}}\\
    CauseEffectPairs~\scriptsize{~\cite{mooij-etal-2016-distinguishing}} & Variable & 108 & 108 & \bempty & \generalCSC \\
    IHDP~\scriptsize{~\cite{shalit-etal-2017-estimating}} & Variable & 2,000 & 2,000 & \bhalf & \biologicalCSC \\
    CRAFT~\scriptsize{~\cite{ates-etal-2022-craft}} & Video & 58,000 & - & \bfull & \physicalCSC \\
    \rowcolor{mygray} \multicolumn{6}{c}{\textit{Empirical causality in text format}}\\
    Temporal-Causal ~\scriptsize{\cite{bethard-etal-2008-building}} &  Clause & 1,000  & 271  & \bempty  & \temporalCSC \\
    CW~\scriptsize{\cite{ferguson-sanford-2008-anomalies}} &  Clause & 128  & 128  & \bfull  & \generalCSC \\
    SemEval07-T4 ~\scriptsize{\cite{girju-etal-2007-semeval}} &  Phrase & 220  & 114  & \bempty & \generalCSC \\
    SemEval10-T8 ~\scriptsize{\cite{hendrickx-etal-2010-semeval}} &  Phrase & 10,717  & 1,331  & \bempty & \generalCSC \\
    COPA~\scriptsize{\cite{roemmele-etal-2011-choice}} &  Sentence & 2,000 & 1,000  & \bempty & \generalCSC \\
    EventCausality ~\scriptsize{\cite{do-etal-2011-minimally}} &  Clause & 583  & 583  & \bempty  & \generalCSC \\
    BioCause~\scriptsize{\cite{mihuailua-etal-2013-biocause}} &  Clause & 851  & 851  & \bempty & \biologicalCSC \\
    CausalTimeBank~\scriptsize{\cite{mirza-etal-2014-annotating}} &  Sentence & 318  & 318  & \bempty & \temporalCSC \\
    CBND~~\scriptsize{\cite{boue-etal-2015-causal}} & Sentence & 120 & 120 & \bempty & \biologicalCSC \\
    CaTeRs~\scriptsize{\cite{mostafazadeh-etal-2016-caters}} &  Sentence & 2,502 & 308  & \bempty & \temporalCSC  \\
    AltLex~\scriptsize{\cite{hidey-mckeown-2016-identifying}} &  Clause & 44,240 & 4,595  & \bempty & \generalCSC \\
    BECauSE~\scriptsize{\cite{dunietz-etal-2017-corpus}} &  Sentence & 729 & 554  & \bempty & \generalCSC \\
    ESL~\scriptsize{\cite{caselli-vossen-2017-event}} &  Sentence & 2,608 & 2,608  & \bempty  & \temporalCSC \\
    TCR~\scriptsize{\cite{ning-etal-2018-joint}} &  Clause & 172 & 172  & \bempty  & \temporalCSC \\
    SocialIQa~\scriptsize{\cite{sap-etal-2019-social}} & Sentence & 37,588 & - & \bempty & \socialCSC \\
    PDTB~\scriptsize{\cite{webber-etal-2019-penn}} &  Clause & 7,991 & 7,991  & \bempty & \generalCSC \\
    TimeTravel~\scriptsize{~\cite{qin-etal-2019-counterfactual}} & Sentence & 109,964 & 29,849 & \bhalf  & \generalCSC\\
    GLUCOSE~\scriptsize{~\cite{mostafazadeh-etal-2020-glucose}} & Clause & 670K & 670K & \bempty & \socialCSC \\
    XCOPA~\scriptsize{~\cite{ponti-etal-2020-xcopa}} & Sentence & 11,000 & 11,000 & \bempty & \generalCSC \\
    CausalBank~\scriptsize{~\cite{li-etal-2021-guided}} & Clause & 314M & 314M & \bempty & \generalCSC \\
    SemEval20-T5~\scriptsize{\cite{yang-etal-2020-semeval}} &  Clause & 25,501  & 25,501  & \bfull  & \generalCSC\\
    e-CARE~\scriptsize{\cite{du-etal-2022-e}} & Sentence & 21,324 & 21,324  & \bempty & \physicalCSC \\
    CoSIm~\scriptsize{\cite{kim-etal-2022-cosim}} & Image\&Text & 3,500 & 3,500 & \bfull & \generalCSC \\
    CRASS~\scriptsize{\cite{frohberg-binder-2022-crass}} & Sentence & 274 & 274 & \bfull & \generalCSC  \\
    COPES~\scriptsize{~\cite{wang-etal-2023-cola}} & Sentence & 1,360 & 1,360 & \bempty &  \generalCSC \\
    IfQA~\scriptsize{~\cite{yu-etal-2023-ifqa}} & Sentence & 3,800 & 3,800 & \bfull &  \socialCSC \\
    CW-extended~\scriptsize{~\cite{li-etal-2023-counterfactual}} & Sentence & 10,848 & 10,848 & \bfull &  \generalCSC \\
    CausalQuest~\scriptsize{~\cite{ceraolo-etal-2024-causalquest}} & Sentence & 13,500 & 13,500 & \bhalf & \generalCSC \\
    $\delta$-CAUSAL~\scriptsize{~\cite{cui-etal-2024-exploring}} & Sentence & 11,245 & 11,245 & \bhalf & \generalCSC \\
    \rowcolor{mygray} \multicolumn{6}{c}{\textit{Empirical commonsense causality in knowledge graph format}}\\ 
    CausalNet~\scriptsize{\cite{luo-etal-2016-commonsense}} &  Word & 11M & 11M  & \bempty & \generalCSC \\
    ConceptNet~\scriptsize{~\cite{speer-etal-2017-conceptnet}} & Phrase & 473,000 & - & \bempty & \generalCSC \\
    Event2Mind~\scriptsize{~\cite{rashkin-etal-2018-event2mind}} & Phrase & 25,000 & - & \bempty & \socialCSC \\
    ATOMIC~\scriptsize{~\cite{sap-etal-2019-atomic}} & Sentence & 877K & - & \bhalf & \socialCSC  \\
    ASER~\scriptsize{~\cite{zhang-etal-2020-aser}} & Sentence & 64M & 494K & \bempty  & \generalCSC \\
    CauseNet~\scriptsize{~\cite{heindorf-etal-2020-causenet}} & Word & 11M & 11M & \bempty & \generalCSC  \\    
    CEGraph~\scriptsize{~\cite{li-etal-2021-guided}} & Phrase & 89.1M & 89.1M & \bempty & \generalCSC \\
    \bottomrule
  \end{tabular}
  }
  \caption{Overview of commonsense causality datasets. A more detailed version is present in \Appendix~\ref{appendix:detailed_benchmarks}. }
  \label{tab:benchmarks}
\end{table}
\footnotetext[2]{\textit{C.F.} denotes whether the dataset contains \textit{counterfactual} reasoning, which can vary from no counterfactuals~(\bempty), a subset being counterfactuals~(\bhalf), to all counterfactuals~(\bfull).
For the commonsense type (\textit{Type}), * means that the dataset covers multiple commonsense types. }

\setcounter{footnote}{2}

Different classification criteria lead to different taxonomies for commonsense causality. We build our criteria based on commonsense types~(\S~\ref{sec:taxonomy:type}) and uncertainty levels~(\S~\ref{sec:taxonomy:uncertainty}). This section corresponds to the context marked in \colorbox{taxonomyColor!60}{\taxonomyColorName color} in Figure~\ref{fig:categorization_full}. 

\subsection{Classification by Commonsense Types} \label{sec:taxonomy:type}
According to the commonsense types~(\Appendix~\ref{appendix:background:commonsense}) on which causality is built, commonsense causality can be roughly classified into four categories:
(i) \textit{Physical causality}~(\physicalCSC) refers to the commonsense cause-effect relationships grounded in the physical world. \physicalCSC usually covers domains such as physics, chemistry, and environmental science, with datasets such as CRAFT~\cite{ates-etal-2022-craft} and e-CARE~\cite{du-etal-2022-e};
(ii) \textit{Social causality}~(\socialCSC) involves the understanding of social norms, cultures, human behavior, intents, and reactions. For instance, criticism~(cause) leads to depression~(effect) in a social context. \socialCSC covers domains like law, culture, education, psychology, etc. Typical examples are ATOMIC~\cite{sap-etal-2019-atomic}, GLUCOSE~\cite{mostafazadeh-etal-2020-glucose}, and IfQA~\cite{yu-etal-2023-ifqa}; 
(iii) \textit{Biological causality}~(\biologicalCSC) relates to cause-effect pairs that govern biological processes and phenomena such as a healthy diet contributes to longevity. Typical benchmarks include BioCause~\cite{mihuailua-etal-2013-biocause}, CBND~\cite{boue-etal-2015-causal}, etc; 
(iv) \textit{Temporal causality}~(\temporalCSC) involves the sequential understanding that the cause must precede the effect in time~\cite{imbens-etal-2022-long,van-etal-2023-estimating}. This type includes Temporal-Causal~\cite{bethard-etal-2008-building}, CausalTimeBank~\cite{mirza-etal-2014-annotating}, CaTeRs~\cite{mostafazadeh-etal-2016-caters}, etc.

\subsection{Classification by Uncertainty Levels} \label{sec:taxonomy:uncertainty}
\SmallHeading{Sources of Uncertainty} Generally, commonsense causality usually involves unobserved facts and uncertainties.  For instance, the claim that ``eating a healthy diet and exercising regularly'' leads to ``a long life'' does not reveal/consider the influence of other factors including genetics, access to healthcare, accidents, and so on. Based on the criteria of causal sufficiency and necessity~(\Appendix~\ref{appendix:concepts}), there are two kinds of uncertainties in commonsense causality~\cite{yarlett-ramscar-2019-uncertainty}: 
(i) \textit{Factual uncertainties} refers to uncertainties caused by insufficient information. This is pervasive in commonsense causality since the knowledge humans possess is always incomplete. For instance, the claim that ``rain makes roads slippery'' does not reveal detailed information about the type of roads~(asphalt, concrete, gravel, earth, chip seal, cobblestones, pervious concrete, etc.) and the intensity of the rain. The missing of these important information influences the validity of causality; 
(ii) \textit{Causal uncertainties} concerns uncertainties due to unstable observation about the cause-effect relation. One example is the claim that ``smoking leads to lung cancer''. Although there is overwhelming evidence that smokers have a high incidence of lung cancer, there are always some people who smoke a lot but do not develop lung cancer.  See more factual and causal uncertainty details in \Appendix~\ref{appendix:uncertainty}. 

\SmallHeading{Categorization by Levels of Uncertainty}
Depending on the level of uncertainty, commonsense causality can be categorized into two types: first-principle causality and empirical causality: 
\begin{itemize}
    \item \textbf{First-principle causality} refers to causal relationships grounded in established laws, such as the link between mass and gravity. Usually, first-principle causality is based on fully observed, well-defined, proven settings based on definite physical or mathematical facts. 
    \item \textbf{Empirical causality} is prone to suffer from various sources of uncertainties. For instance, it is common knowledge that stepping on a banana peel causes one to slip. However, the validity of this causal relationship is influenced by factors such as the condition of the banana peel(e.g., fresh or dried, factual uncertainties), the condition of the roads~(is stepping on the banana peel is the real cause or the wet or oily surface of the road is the true causes, causal uncertainties). 
\end{itemize}

Existing benchmarks, categorized by the two criteria aforementioned, are summarized in Table~\ref{tab:benchmarks}. Further classifications based on skill sets and entity types are detailed in \Appendix~\ref{appendix:taxonomy}.

\section{Causality Acquisition} \label{sec:acquisition} 
Common methods for acquiring commonsense causality benchmarks are categorized into three main approaches: extractive methods~(\S~\ref{sec:acquisition:extraction}), generative methods~(\S~\ref{sec:acquisition:generative}), and manual annotation methods~(\S~\ref{sec:acquisition:manual}). These methods are summarized in Figure~\ref{fig:categorization_full} with a \colorbox{acquisitionColor!60}{\acquisitionColorName background color}

\subsection{Extractive Methods} \label{sec:acquisition:extraction}
\begin{table}[htb!]
    \centering
    \resizebox{0.99\linewidth}{!}{
    \begin{tabular}{p{2cm}p{6.4cm}} 
        \toprule %
        \textbf{Form} & \textbf{Connectives} \\
        \midrule
        \rowcolor{mygray} \multicolumn{2}{c}{\textit{\textbf{Cause-Effect Connectives}}}\\
        Cause-Effect &  as, because, cause, since, bring about, due to, lead to,   owing to, resulting in\\
        Consequence & accordingly, as a result, consequently, for this reason, hence, so, therefore, thus\\
        Reason & in light of, given that, on account of, by reason of, for the sake of, inasmuch as, seeing that\\
        Intention & so that, in order to, so as to, with the aim of, for the purpose of, with this in mind, in hopes of\\
        Conditions & if...then, provided that, assuming that, as long as, unless, in the event that\\
        Source & arises from, stems from, comes from, originates from\\
        \rowcolor{mygray} \multicolumn{2}{c}{\textit{\textbf{Counterfactual Connectives}}}\\
        Hypothetical & had...then, if it hadn't been for, had it not been for, if only\\
        Negation & were it not for, but for, if it weren't for, without, in the absence of, lacking \\
        \bottomrule %
    \end{tabular}
    }
    \caption{Common causality-related connectives. The presence of these connectives usually implies the existence of causal relations, which is commonly used in extractive methods. }
    \label{tab:causal_connectives}
\end{table}

\SmallHeading{Benchmarks} The automatic extraction methods are based on annotated domain corpus: open-source text and standard benchmarks. The open-source corpus generally refers to the content available on web pages or Wikipedia. 
The standard benchmarks cover a variety of datasets such as SemEval07-T4~\cite{girju-etal-2007-semeval}, CNN-extraction~\cite{do-etal-2011-minimally}, ESL~\cite{caselli-vossen-2017-event} and PDTB~\cite{webber-etal-2019-penn} from the general domain, as well as BioInfer~\cite{pyysalo-etal-2007-bioinfer} and ADE~\cite{gurulingappa-etal-2012-development} from the biomedical domain.
A detailed description of these benchmarks is presented in \Appendix~\ref{appendix:detailed_benchmarks}. 

\begin{figure} [htp!]
  \centering
  \resizebox{0.99\linewidth}{!}{
  \usetikzlibrary{positioning}
\begin{tikzpicture}
  \begin{scope}[local bounding box=blocks]
  \node[draw, rectangle, rounded corners, minimum height=2em, minimum width=3em, fill=blue!20] (before) at (-0.5,0) {The explosion};
  \node[below=1.9em of before] {Noun phrase};
  
  \node[draw, rectangle, rounded corners, minimum height=2em, text width=3em, align=center, fill=green!20] (altlex) at (2.9,0) {made\\forced\\caused};
  \node[below=1em of altlex, text width=10em, align=center] {Alternative lexicalization verbs~(AltLex)};
  
  \node[draw, rectangle, rounded corners, minimum height=2em, minimum width=3cm, fill=red!20] (after) at (7.5,0) {people (to) evacuate the building.};
  \node[below=1.9em of after] {Object complement};
    \end{scope}
  
  \draw[rounded corners, dotted, very thick] ([xshift=-0.3cm,yshift=-0.3cm]blocks.south west) rectangle ([xshift=0.3cm,yshift=0.3cm]blocks.north east);
\end{tikzpicture}
  }
  \caption{A template of pattern matching from AltLex~\cite{hidey-mckeown-2016-identifying}.}
  \label{fig:pattern-altlex}
\end{figure}
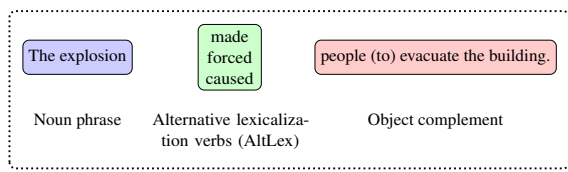

\SmallHeading{Linguistic Pattern Matching Methods}
The methods for extracting causality from text by linguistic pattern matching can be either \emph{clue}-based or \emph{rule}-based. 
(i) The clue-based approach~\cite{sakaji-etal-2008-extracting,cao-etal-2014-mining} relies on hand-crafted or automatically generated clues to detect the presence of causation. For instance, the presence of the words ``cause'' or ``accordingly'' always indicates causality. We list common causal connectives in Table~\ref{tab:causal_connectives};  
(ii) The pattern/rule-based approach~\cite{girju-2003-automatic,cole-etal-2006-lightweight,ishii-etal-2010-causal} predefines a specific semantic format for extracting causality from text. 
One common format is a \textit{noun phrase}, a \textit{causation verb}~(see \Appendix~\ref{appendix:linguistic:verbs} for a detailed list of causation verbs), and another \textit{noun phrase or an object complement}.  We provide an example sentence within this format in Figure~\ref{fig:pattern-altlex}. 

\SmallHeading{Machine and Deep Learning-Based Methods} 
Machine learning-based methods use traditional machine learning models like Support Vector Machines~(SVMs)~\cite{cortes-vapnik-1965-support} or Decision Trees~(DTs)~\cite{quinlan-1986-induction} to detect the presence of causal relationships. The handcrafted or automatically generated textual features, e.g., dependency parsing features, causal patterns~\cite{girju-2003-automatic,blanco-etal-2008-causal}, the presence of causatives, causal connectives~\cite{zhao-etal-2016-event} are taken as the input features to the machine learning models, which are then trained to learn the causal extractor. 
In addition to the conventional machine learning techniques, with the recent success of deep neural networks in various tasks, the deep learning models especially the pre-trained language models provide a more powerful engine for causality extraction.

\subsection{Generative Methods} \label{sec:acquisition:generative}
The rapid advance of generative language models like T5~\cite{raffel-etal-2020-exploring} and ChatGPT~\cite{openai-etal-2023-gpt4} enables the LLMs to be useful tools for generating reliable cause-effect pairs~\cite{kim-etal-2023-soda}. 
\citet{rashkin-etal-2018-event2mind} utilize an encoder-decoder structure for generating intents/reactions over a range of daily events, which contains a variety of causal relationships. CauseWorks~\cite{choudhry-2020-narrative} is a generative method that converts causal graphs into textual narratives of causal relationships. \citet{li-etal-2021-guided} firstly utilize pattern matching to build a causal graph CausalBank, and then employ a Sequence-to-Sequence model to generate the textual cause-effect pairs.\footnote{Note that although some works~\cite{madaan-etal-2021-generate,robeer-etal-2021-generating-realistic,wu-etal-2021-polyjuice,calderon-etal-2022-docogen,chen-etal-2023-disco} focusing on counterfactual generation, some of them are more on the side of adversarial/fake samples generation instead of the counterfactual meaning in causal reasoning. 
} 

\subsection{Manual Annotation} \label{sec:acquisition:manual}
Apart from the automatic extraction strategies, manual annotation is also an important approach for collecting commonsense causality benchmarks. There are plenty of general annotation schemes in semantic parsing that introduce the causation as \emph{one of} the semantic relations to be annotated. Some representative schemes include PropBank~\cite{palmer-etal-2005-proposition}, FrameNet~\cite{baker-etal-1998-berkeley,ruppenhofer-etal-2016-framenet}, PDTB~\cite{prasad-etal-2008-penn}, RST~\cite{mann-thompson-1988-rhetorical}, AMT~\cite{banarescu-etal-2013-abstract} and so on. 
Besides general schemes, there are schemes designed exclusively for annotating causal relations. For example, BioCause~\cite{mihuailua-etal-2013-biocause}, TimeML~\cite{mirza-etal-2014-annotating}, RED~\cite{ikuta-etal-2014-challenges}, CaTeRs~\cite{mostafazadeh-etal-2016-caters}, and CxG~\cite{dunietz-2018-annotating} all fall into this framework. More discussion on annotation schemes is in \Appendix~\ref{appendix:acquisition:annotation}. 

\subsection{Comparison of Data Acquisition Methods} \label{sec:acquisition:comparison}
The summary of the pros and cons of these acquisition methods is presented in Table~\ref{tab:acquisition_comparison}. Generally, compared to extractive and generative methods, manual annotation provides the highest quality data and is more explainable. However, it suffers from cost and efficiency issues and thus lacks scalability and coverage. We refer to \Appendix~\ref{appendix:acquisition:comparison} for a more detailed comparison. 
\begin{table}[h]
    \centering
    \resizebox{0.5\textwidth}{!}{
    \begin{tabular}{lcccc} 
        \toprule 
        \textbf{Method} & \textbf{Accuracy} & \textbf{Cost} & \textbf{Coverage} & \textbf{Explainablity}\\
        \midrule
        Extractive &  \fourstars{}  &  \fivestars{} &   \fourstars{} & \fourstars{} \\
        Generative & \threestars{} & \fourstars{} &  \threestars{} & \onestar{} \\
        Manual Annotation & \fivestars{} & \twostars{} & \fivestars{} & \fivestars{} \\
        \bottomrule 
    \end{tabular}
    }
    \caption{Comparison of different commonsense causality acquisition methods. The more solid stars, the better.
    }
    \label{tab:acquisition_comparison}
\end{table}

The aforementioned methods are mainly targeted at explicit causality acquisition and they are more centered on causality inside the sentence. However, causality is not always explicit and may appear in different sentences.  
More details about implicit causal relationships and inter-sentential causality can be found in \Appendix~\ref{appendix:acquisition:other_causality}.

\section{Reasoning Over Causality} \label{sec:analysis}

This section reviews qualitative and quantitative causal reasoning approaches for addressing uncertainty in commonsense causality, as discussed in \S~\ref{sec:taxonomy:uncertainty}. Qualitative methods (\S~\ref{sec:analysis:qualitative}) treat causal reasoning as a 0/1 classification task, while quantitative methods (\S~\ref{sec:analysis:quantitative}) quantify causality strength numerically. This section relates to the content highlighted in \colorbox{analysisColor}{\analysisColorName color} in Figure~\ref{fig:categorization_full}.

\subsection{Bypassing Uncertainty by Qualitative Causal Reasoning} \label{sec:analysis:qualitative}
\SmallHeading{Scaling NLP Models as Causal Knowledge Bases}
The evolution of commonsense reasoning is in parallel with the advancement of NLP models. NLP models can be used as the causal knowledge bases that are distilled from the training data or pre-training corpora. 
NLP models experienced four stages of development: 
(i) Statistical Methods: The initial approach in NLP analyzes patterns and linguistic correlation of text resources to identify causal relationships. They are solely based on term co-occurrence and thus suffer from complex causal structures; 
(ii) Deep Learning Methods: Methods based on neural network architectures, especially recurrent neural networks and later transformers, are more capable of capturing contextual information. Consequently, they show substantial improvements in the identification and analysis of causal relationships in text;  
(iii) Pre-Trained Language Models: Language models like BERT~\cite{devlin-etal-2019-bert} and GPT~\cite{brown-etal-2020-language} that are trained on large corpora expand the reasoning ability drastically. When fine-tuned for causal/counterfactual reasoning tasks, they can not only identify the causal relationship but also comprehend the subtleties inherent in commonsense causality such as implicit causality, temporal constraints, etc; 
(iv) LLMs~\cite{openai-etal-2023-gpt4,jiang-etal-2023-mistral,touvron-etal-2023-llama,mesnard-etal-2024-gemma}: We are now in the era of LLMs employed with prompting techniques~\cite{wei-etal-2022-chain,yu-etal-2023-alert,alkhamissi-etal-2023-opt}. 
They enable more accurate understanding, predictions, and explanations of causal and counterfactual scenarios.

A detailed chronological overview of these advancements and their impact on causal reasoning is provided in Figure~\ref{fig:nlp_timeline} and \Appendix~\ref{appendix:lm}.  

\SmallHeading{Neuro-Symbolic Methods}
Neuro-symbolic methods represent an innovative approach to computational reasoning, overcoming the limitations of traditional NLP models that struggle with complex, non-linear causal relationships. These methods leverage the synergy of neural networks and symbolic logic, blending the pattern-recognition prowess of the former with the explicit, interpretable reasoning of the latter. We categorize these neuro-symbolic strategies into three distinct subcategories:

\begin{itemize}
\item \textit{Reasoning with Causal Inference Rules:} Techniques like ROCK~\cite{zhang-etal-2022-rock} and COLA~\cite{wang-etal-2023-cola} employ the concept of Average Treatment Effect (ATE) to assess the likelihood of one event causing another. ATE is instrumental in quantifying the effect of a treatment on an outcome, represented as $P(E_i \to E_j) = p(E_i \prec E_j) - p(\neg E_i \prec E_j)$. Furthermore, \citet{jin-etal-2023-cladder} integrates causal inference steps into chain-of-thought reasoning, a method pioneered by \citet{wei-etal-2022-chain}. Preliminary of causal inference is elaborated in \Appendix~\ref{appendix:causal_inference}. 
\item \textit{Explicitly Incorporating Temporal Constraints}: Recognizing that the cause must precede the effect in time~--~a fundamental principle in science~--~methods like those proposed by \citet{ning-etal-2018-joint} introduce temporal constraints. These constraints aid in causal reasoning, reformulating the problem as an integer linear programming challenge. 
\item \textit{Integrating Logic Rules}: This approach~\cite{zhang-foo-2001-epdl,bochman-2003-logic,saki-faghihi-2022-fundamental} involves embedding logic rules directly into the reasoning mechanism, thereby enhancing the model's ability to handle complex, logically-driven tasks and presenting better explainability. 
\end{itemize}

\subsection{Measuring Uncertainty by Quantitative Causal Reasoning} \label{sec:analysis:quantitative}
While qualitative causal reasoning focuses on distinguishing true cause-effect relationships from erroneous ones, it faces challenges due to uncertainties and the defeasible nature of commonsense causality~\cite{marcos-2021-study,cui-etal-2024-exploring}. Quantitative approaches aim to address these challenges by measuring the likelihood of a cause leading to an effect, thus providing a nuanced understanding of causality. Existing methods for quantitative causal reasoning can be roughly categorized into two types. 

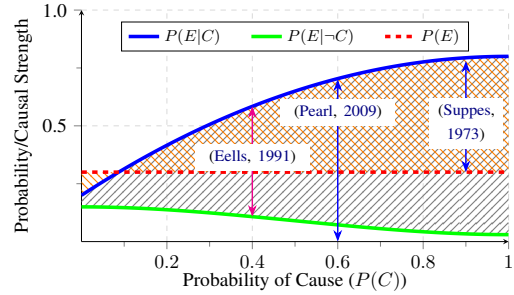
\begin{figure}[htp!] 
  \centering
  \resizebox{0.9\linewidth}{!}{
  \begin{tikzpicture}
\begin{axis}[
    axis lines=middle,
    width=10cm, 
    height=6.18cm,
    xmin=0, xmax=1, ymin=0, ymax=1,
    xlabel=Probability of Cause ($P(C)$),
    ylabel=Probability/Causal Strength,
    xlabel style={at={(axis description cs:0.5,-0.1)}, anchor=north},
    ylabel style={at={(axis description cs:-0.1,0.5)}, rotate=90, anchor=south},
    xticklabel style={/pgf/number format/fixed},
    yticklabel style={/pgf/number format/fixed},
    grid=major,
    grid style={dashed, gray!30},
    ytick={0.0,0.5,1.0},
    yticklabels={0.0,0.5,1.0},
    tick align=outside,
    minor tick num=1,
    enlargelimits=false,
    legend style={
        at={(0.5,0.95)}, 
        anchor=north, 
        legend columns=3, 
        font=\small, 
        /tikz/every even column/.append style={column sep=0.5cm}
    }
]
\addplot[name path=quadratic, domain=0:1, samples=100, smooth, line width=2pt, blue] {-0.6*x^2 + 1.2*x + 0.2};
\addlegendentry{$P(E \vert C)$}

\addplot[name path=cubic, domain=0:1, samples=100, smooth, line width=2pt, green] {0.24*x^3 - 0.36*x^2 + 0.15};
\addlegendentry{$P(E \vert \neg C)$}

\addplot[name path=horizontal, dashed, line width=2pt, red] coordinates {(0,0.3) (1,0.3)};
\addlegendentry{$P(E)$}

\addplot[pattern=north east lines, pattern color=gray] fill between[of=quadratic and cubic];

\addplot[pattern=north west lines, pattern color=orange] fill between[of=horizontal and quadratic];

\draw [<->, >=Stealth, thick, magenta] (axis cs:0.4,0.584) -- (axis cs:0.4,0.10776);
\node [align=center, fill=white, text width=2cm, anchor=south] at (axis cs:0.4,0.305) {\small{\cite{eells-1991-probabilistic}}};

\draw [<->, >=Stealth, thick, blue] (axis cs:0.6,0.7) -- (axis cs:0.6,0);
\node [align=center, fill=white, text width=2cm, anchor=south] at (axis cs:0.6,0.5) {\small{\cite{pearl-2009-causality}}};

\draw [<->, >=Stealth, thick, blue] (axis cs:0.9,0.78) -- (axis cs:0.9,0.3);
\node [align=center, fill=white, text width=1.2cm, anchor=south] at (axis cs:0.9,0.4) {\small{\cite{suppes-1973-probabilistic}}};

\end{axis}
\end{tikzpicture}
  }
  \caption{Comparison of different causal strength metrics~\cite{suppes-1973-probabilistic,eells-1991-probabilistic,pearl-2009-causality}. }
  \label{fig:metric_comparison}
\end{figure}

\SmallHeading{Measurement Based on  Event Probability}
This body of work adopts a probabilistic perspective on causality, positing that a cause \textit{increases} the likelihood of an effect occurring. This perspective is framed by two principal probability constraints\footnote{These probabilistic constraints clear off the challenges of \textit{imperfect regularity} and \textit{irrelevance} but still struggle with the challenges of \textit{asymmetry} and \textit{spurious regularities}. More details can be referred to \cite{hitchcock-1997-probabilistic}. }: 
\begin{equation}
    \begin{cases}
        P(E \vert C) > P(E) \\
        P(E \vert C) > P(E \vert \neg C)
    \end{cases}
\end{equation}
where $C$ represents the cause, $E$ denotes the effect, and $\neg C$ signifies any event other than $C$. These constraints argue that the presence of $C$ elevates the likelihood of $E$ compared to the absence of $C$ or the presence of any alternative event $\neg C$.
We summarized several key metrics developed from these two constraints in Table~\ref{tab:metrics}. 
\begin{table}[htp!]
  \centering 
  \begin{tabular}{p{2.8cm}p{3.5cm}}
    \toprule
     & Formulation  \\
    \midrule
    \cite{good-1961-causal} & \(\log \frac{1 - P(E \vert \neg C)}{1 - P(E \vert C)}\) \\
    \cite{suppes-1973-probabilistic} & $P(E \vert C) - P(E)$ \\
    \cite{eells-1991-probabilistic} & $P(E \vert C) - P(E \vert \neg C)$ \\
    \cite{pearl-2009-causality} & $P(E \vert C)$
    \\ \bottomrule
  \end{tabular}
  \caption{Probabilistic causal strength metrics. }
  \label{tab:metrics}
\end{table}
Although these metrics appear intuitive and easy to understand at first glance, they are actually difficult to characterize in practice for the following two reasons. 
Firstly, accurately estimating the conditional probabilities $P(E \vert C)$ and $P(E \vert \neg C)$ is challenging due to linguistic variability. Secondly, the solution space for $\neg C$ is vast and cannot be exhaustively explored.
The comparison of these causal strength metrics is illustrated in Figure~\ref{fig:metric_comparison}.

\SmallHeading{Measurement Based on Word Co-occurrences} 
This approach conceptualizes the causal strength between two events as the cumulative effect of word-level causal strengths of word pairs within these events. The word-level causal strength is measured based on the frequency of word co-occurrences. One example metric CEQ~\cite{luo-etal-2016-commonsense}, which estimates the sentence-level causality by synthesizing the word-level causality. 
\begin{equation}
\resizebox{\linewidth}{!}{
$
    \mathcal{CS}_{\text{CEQ}}(E_1, E_2) = \frac{1}{N_{E_1} + N_{E_2}} \sum_{w_i \in E_1, w_j \in E_2} \text{cs}(w_i, w_j)
$
}
\end{equation}
where $N_{E_1}$ and $N_{E_2}$ are respectively the number of words within the sentences corresponding to the events $E_1$ and $E_2$. $\text{cs}(w_i, w_j)$ is the causal strength between the word $w_i$ and $w_j$. This word-level causal strength is derived based on the estimation from a large-scale web corpus proposed in \cite{luo-etal-2016-commonsense}. 
In contrast to the simple average of word-level causal strengths in CEQ, CESAR~\cite{cui-etal-2024-exploring} adopt a weighted aggregation strategy to emphasize word pairs with strong causal indicators, such as ``CO\textsubscript{2}'' and ``warming'':
\begin{equation}
\label{eq:cesar}
      \mathcal{CS}_{\text{CESAR}}(C, E) = \sum_{e_i \in C} \sum_{e_j \in E} a_{ij}\frac{|e_i^T e_j|}{\Vert e_i\rVert\Vert e_j\rVert}
\end{equation}
where $e_i$ and $e_j$ are the causal embeddings for tokens in $C$ and $E$, respectively. And $a_{ij}$ is the weighting factor. These causal embeddings are generated by a BERT encoder model that is trained on a causal reasoning dataset, which incorporates considerations of uncertainty.

\SmallHeading{Comparison of Qualitative and Quantitative Causal Reasoning Approaches} \label{sec:analysis:comparison}
We compare the qualitative and quantitative causal reasoning methods from their objectives, applications, merits, and limitations. Please see details in Table~\ref{tab:comparison:qualitative_vs_quantitative}. 

\begin{table}[ht]
\newcommand{\graycline}[0]{\arrayrulecolor{gray!50}\cline{0-2}\arrayrulecolor{black}}
\centering
\resizebox{\linewidth}{!}{
    \begin{tabular}{p{2cm}p{3.8cm}p{3.8cm}}
    \toprule
    \textbf{Aspect} & \textbf{Qualitative Reasoning} & \textbf{Quantitative Reasoning} \\
    \midrule
    Objectives & To identify the causal relationship between variables. & To provide precise estimates of causal effects.   \\ \graycline
    Merits & (i) Intuitive understanding;  (ii) Easy to use. & (i) Precise estimation;  (ii) Good comparability across different cause-effect pairs; \\ \graycline
    Limitations & (i) Lack of precision;  (ii) Oversimplication, e.g., confounders are not considered; & (i) Challenges in estimating probabilistic terms like $P(E \vert C)$, $P(E \vert \neg C)$, etc; (ii) Need for large amount of high-quality causality data.  \\ \graycline
    Applications &(i) Identification of potential causal relationship between variables; (ii) Simple decision-making tasks where actions are determined by straightforward cause-and-effect determination. & (i) Quantification of uncertainty factor; (ii) Robust decision making; (iii) Comparable analysis for fine-grained causality.  \\
    \bottomrule
    \end{tabular}
}
\caption{Comparison of qualitative and quantitative causal reasoning approaches.}
\label{tab:comparison:qualitative_vs_quantitative}
\end{table}

\section{Future Research Directions} \label{sec:future}\SmallHeading{Contextual Nuances: Exploring Context-Dependent Commonsense Causality}
Contextual commonsense causality refers to the phenomenon where cause-effect relationships are valid within specific contexts but may not apply universally. 
For instance, while exercise typically benefits health, it can pose risks for individuals with heart conditions, potentially leading to severe consequences. This variability underscores the importance of understanding the contextual dynamics influencing causality. \citet{dupre-1984-probabilistic} introduced the concept of contextual-unanimity causality to capture these contextual nuances:
\begin{equation}
\resizebox{0.99\linewidth}{!}{
$
\sum_{B \in \mathbb{B}} P(E \vert C, B) \times P(B) > \sum_{B \in \mathbb{B}} P(E \vert \neg C, B) \times P(B)
$
}
\end{equation}
where $\mathbb{B}$ represents the set of all potential conditions, contexts, or backgrounds. According to this formulation, the presence of $C$ should increase the average likelihood of $E$ conditional on all conceivable contexts $B$. 
Although this formula provides us with the basic idea of describing contextual causality, it contains several quantities that are difficult to obtain. More work is needed in the future to address these issues: 
(i) Estimation of $P(B)$ and $\mathbb{B}$: Identifying a comprehensive set of conditions $\mathbb{B}$ and characterizing $P(B)$ precisely to minimize contextual unpredictability in commonsense cause-effect relationships;
(ii) Partial Contextual Models: 
Instead of accounting for all possible contexts $\mathbb{B}$, these partial contextual models focus on a subset of contexts $\mathbb{B}' \subseteq \mathbb{B}$ that are deemed most relevant or have the most significant impact on the cause-effect relationship. The objective is to find an optimal $\mathbb{B'}$ such that the model balances between accuracy~(in terms of explaining the causality between $C$ and $E$) and simplicity (minimizing the size of $\mathbb{B'}$). This can be formalized as an optimization problem:
\begin{equation}
    \max_{\mathbb{B'} \subseteq \mathbb{B}} \biggl\{ \sum_{B' \in \mathbb{B'}} P(E \vert C, B') \times P(B') - \lambda \cdot |\mathbb{B'}| \biggr\}
\end{equation}
where $\lambda$ is a regularization parameter that controls the trade-off between the model's complexity (the number of contexts considered) and its explanatory power for commonsense causality.

\SmallHeading{Unveiling Complex Structures: Understanding Complex Commonsense Causality}
In the domain of commonsense causality, reality often extends beyond simple, direct cause-and-effect relationships to encompass richer, more intricate structures such as confounders, colliders, causal chains, and cyclic causality. Such complex causal frameworks, detailed further in \Appendix~\ref{appendix:causal_inference:concepts}, underscore the intricate nature of commonsense causality where multiple variables interact to influence outcomes. Promising topics in this domain include
(i) Development of Complex Structure Commonsense Causality Benchmarks: Creating comprehensive benchmarks that capture the richness of complex structural commonsense causality is the cornerstone of our study for understanding the complexity of real-world causal relationships;
(ii) Theoretical Frameworks for Complex Structures Analysis: More efforts should be put into developing theoretical frameworks that are capable of modeling these sophisticated structures. 
For example, the confounders $C_{ij}$~---~variables that influence both the cause $X_i$ and the effect $Y_j$~---~can be identified by a structural equation model: $C_{ij} = f(X_i, Y_j)$. 

\SmallHeading{Temporal Dynamics: Unraveling the Role of Time in Commonsense Causality} 
Temporal dynamics are fundamental to causality, requiring that causes must precede effects. Despite its apparent simplicity, temporal dynamics offer rich future research avenues:
(i) Optimal Timing for Intervention: This research aims to determine the best times for interventions that prevent negative outcomes, using causal insights to proactively mitigate risks; 
(ii) Temporal Patterns of Causal Effects: This direction studies how the impact/effect of a cause varies over time, from immediate, mid-term to long-term effects. This research is vital for informed decision-making, allowing for consideration of an action's extended consequences in the long run.

\SmallHeading{Beyond Binary: Expanding Probabilistic Perspectives in Causality Measurement}
As highlighted in Section~\ref{sec:taxonomy:uncertainty}, commonsense causality transcends deterministic frameworks, embodying inherent uncertainties. To navigate and quantify these uncertainties, we suggest two promising research directions that employ a probabilistic perspective:
(i) Probabilistic Graphical Models: Developing probabilistic graphical models, such as Bayesian Networks~\cite{heckerman-2008-tutorial} or Markov Random Fields, to model probabilistic commonsense causality. The focus would be on characterizing conditional probability distributions $P(E \vert C)$ that quantify the probabilities of cause-effect relationships; 
(ii) Dynamic Probabilistic Causal Models with Temporality: This path delves into dynamic causal models that integrate the dimension of time, thereby enhancing the understanding of how causation probabilities evolve over time. This direction might entail the use of differential equations or discrete-time models that estimate $P(E_t \vert C_{t - \delta})$~---~the probability of an effect $E$ at time $t$ given a cause $C$ at a preceding time $t-\delta$.

\SmallHeading{Expanding Horizons: Advancing Multimodal Approaches for Commonsense Causality} 
Multimodal commonsense causality refers to cause-effect pairs whose entities are converted beyond text such as audio, image, and video. 
The burgeoning availability of multimodal data coupled with advancements in multimodal models~\cite{lu-etal-2019-vilbert, chen-etal-2020-uniter, li-etal-2020-oscar} has made the study of commonsense causality both more urgent and achievable.
Here we provide several prospective research topics: 
(i) Advancing Acquisition and Reasoning for Multimodal Commonsense Causality: This topic focuses on developing refined methodologies for the collection and analysis of multimodal data to identify and reason cause-effect relationships within commonsense knowledge; 
(ii) Cross-Modal Cause-Effect Pair Alignment: It focuses on synchronizing cause-effect pairs across modalities. For example, the cause is a text narrator about deforestation in the Amazon rainforest, while the effect is in videos of trucks carrying logs and the resulting habitat loss for indigenous species.
Key challenges involve creating techniques for cross-modal representation and developing robust evaluation metrics for alignment accuracy.

\section{Conclusion} \label{sec:conclusion}In this survey, we present an overview of commonsense causality, including its taxonomy, benchmarks, and data acquisition methods, along with qualitative and quantitative reasoning approaches. 
Furthermore, we shed light on several future promising research directions. 
Our work, drawing on insights from over 200 articles, aims to provide a thorough understanding of commonsense causality in the era of LLMs. 
Additionally, we include a pragmatic handbook in \Appendix~\ref{appendix:handbook} for researchers interested in further exploration of this field. 

\section*{Limitations} \label{sec:limitations}In this study, we provide a survey of commonsense causality in the context of natural language processing. We try our best to provide a bird's-eye view of commonsense causality in an 8-page paper. Notwithstanding our best efforts, this paper still has some limitations. 
Firstly, it is difficult to cover every aspect of commonsense causality due to the page limit. We choose to focus on specific subtopics including benchmarks, acquisition, qualitative reasoning, and quantitative measurement while the other areas receive less attention. Besides, we focus more on papers already being published while not capturing the unpublished works. Notwithstanding our best efforts and an extraordinarily detailed appendix, some relevant work may be unintentionally omitted. 
Furthermore, commonsense causality is an interdisciplinary area requiring expertise in linguistics, psychology, philosophy, and NLP. It is difficult to delve into each area in a survey paper. We are compelled to engage in prioritization and compromise. We place a greater emphasis on the NLP domain, with the employed methodologies predominantly leaning towards the realm of NLP.
\section*{Ethical Considerations} \label{sec:ethical}As a survey paper on a commonly addressed NLP task, there are no foreseeable major ethical concerns. 
All the investigated benchmarks or methods are clearly cited and used in their intended purpose. 
A minor concern is that while we analyzed the benchmarks, we found that some dataset papers did not provide licenses for using their data, which may cause concerns about ethical usage. Besides, for a broad topic like commonsense causality, oversimplification for certain theories or resources is likely to happen due to the limitation of coverage as well as the concerns raised in the previous limitation section. 

\bibliography{acl2023}

\begin{thebibliography}{225}
\expandafter\ifx\csname natexlab\endcsname\relax\def\natexlab#1{#1}\fi

\bibitem[{Airola et~al.(2008)Airola, Pyysalo, Bj{\"o}rne, Pahikkala, Ginter, and Salakoski}]{airola-etal-2008-graph}
Antti Airola, Sampo Pyysalo, Jari Bj{\"o}rne, Tapio Pahikkala, Filip Ginter, and Tapio Salakoski. 2008.
\newblock \href {https://aclanthology.org/W08-0601} {A graph kernel for protein-protein interaction extraction}.
\newblock In \emph{Proceedings of the Workshop on Current Trends in Biomedical Natural Language Processing}, pages 1--9, Columbus, Ohio. Association for Computational Linguistics.

\bibitem[{Akl et~al.(2020)Akl, Mariko, and Labidurie}]{abi-etal-2020-semeval}
Hanna~Abi Akl, Dominique Mariko, and Estelle Labidurie. 2020.
\newblock \href {http://arxiv.org/abs/2005.08519} {Semeval-2020 task 5: Detecting counterfactuals by disambiguation}.
\newblock \emph{CoRR}, abs/2005.08519.

\bibitem[{Alkhamissi et~al.(2023)Alkhamissi, Verma, Yu, Jin, Celikyilmaz, and Diab}]{alkhamissi-etal-2023-opt}
Badr Alkhamissi, Siddharth Verma, Ping Yu, Zhijing Jin, Asli Celikyilmaz, and Mona Diab. 2023.
\newblock \href {https://doi.org/10.18653/v1/2023.nlrse-1.10} {{OPT}-{R}: Exploring the role of explanations in finetuning and prompting for reasoning skills of large language models}.
\newblock In \emph{Proceedings of the 1st Workshop on Natural Language Reasoning and Structured Explanations (NLRSE)}, pages 128--138, Toronto, Canada. Association for Computational Linguistics.

\bibitem[{Andreas and Guenther(2021)}]{andreas-guenther-2021-regularity}
Holger Andreas and Mario Guenther. 2021.
\newblock {Regularity and Inferential Theories of Causation}.
\newblock In Edward~N. Zalta, editor, \emph{The {Stanford} Encyclopedia of Philosophy}, {F}all 2021 edition. Metaphysics Research Lab, Stanford University.

\bibitem[{Asghar(2016)}]{asghar-2016-automatic}
Nabiha Asghar. 2016.
\newblock \href {http://arxiv.org/abs/1605.07895} {Automatic extraction of causal relations from natural language texts: {A} comprehensive survey}.
\newblock \emph{CoRR}, abs/1605.07895.

\bibitem[{Asr and Demberg(2012)}]{asr-demberg-2012-implicitness}
Fatemeh~Torabi Asr and Vera Demberg. 2012.
\newblock \href {https://api.semanticscholar.org/CorpusID:60367} {Implicitness of discourse relations}.
\newblock In \emph{International Conference on Computational Linguistics}.

\bibitem[{Ates et~al.(2022)Ates, Ate{\c{s}}o{\u{g}}lu, Yi{\u{g}}it, Kesen, Kobas, Erdem, Erdem, Goksun, and Yuret}]{ates-etal-2022-craft}
Tayfun Ates, M.~Ate{\c{s}}o{\u{g}}lu, {\c{C}}a{\u{g}}atay Yi{\u{g}}it, Ilker Kesen, Mert Kobas, Erkut Erdem, Aykut Erdem, Tilbe Goksun, and Deniz Yuret. 2022.
\newblock \href {https://doi.org/10.18653/v1/2022.findings-acl.205} {{CRAFT}: A benchmark for causal reasoning about forces and in{T}eractions}.
\newblock In \emph{Findings of the Association for Computational Linguistics: ACL 2022}, pages 2602--2627, Dublin, Ireland. Association for Computational Linguistics.

\bibitem[{Baker et~al.(1998)Baker, Fillmore, and Lowe}]{baker-etal-1998-berkeley}
Collin~F. Baker, Charles~J. Fillmore, and John~B. Lowe. 1998.
\newblock \href {https://aclanthology.org/C98-1013} {The {B}erkeley {F}rame{N}et project}.
\newblock In \emph{{COLING} 1998 Volume 1: The 17th International Conference on Computational Linguistics}.

\bibitem[{Banarescu et~al.(2013)Banarescu, Bonial, Cai, Georgescu, Griffitt, Hermjakob, Knight, Koehn, Palmer, and Schneider}]{banarescu-etal-2013-abstract}
Laura Banarescu, Claire Bonial, Shu Cai, Madalina Georgescu, Kira Griffitt, Ulf Hermjakob, Kevin Knight, Philipp Koehn, Martha Palmer, and Nathan Schneider. 2013.
\newblock \href {https://aclanthology.org/W13-2322} {{A}bstract {M}eaning {R}epresentation for sembanking}.
\newblock In \emph{Proceedings of the 7th Linguistic Annotation Workshop and Interoperability with Discourse}, pages 178--186, Sofia, Bulgaria. Association for Computational Linguistics.

\bibitem[{Barik et~al.(2017)Barik, Marsi, and {\"{O}}zt{\"{u}}rk}]{barik-etal-2017-extracting}
Biswanath Barik, Erwin Marsi, and Pinar {\"{O}}zt{\"{u}}rk. 2017.
\newblock \href {https://doi.org/10.1007/978-3-319-59569-6\_13} {Extracting causal relations among complex events in natural science literature}.
\newblock In \emph{Natural Language Processing and Information Systems - 22nd International Conference on Applications of Natural Language to Information Systems, {NLDB} 2017, Li{\`{e}}ge, Belgium, June 21-23, 2017, Proceedings}, volume 10260 of \emph{Lecture Notes in Computer Science}, pages 131--137. Springer.

\bibitem[{Bethard et~al.(2008)Bethard, Corvey, Klingenstein, and Martin}]{bethard-etal-2008-building}
Steven Bethard, William Corvey, Sara Klingenstein, and James~H. Martin. 2008.
\newblock \href {http://www.lrec-conf.org/proceedings/lrec2008/pdf/229_paper.pdf} {Building a corpus of temporal-causal structure}.
\newblock In \emph{Proceedings of the Sixth International Conference on Language Resources and Evaluation ({LREC}'08)}, Marrakech, Morocco. European Language Resources Association (ELRA).

\bibitem[{Bhargava and Ng(2022)}]{bhargava-ng-2022-commonsense}
Prajjwal Bhargava and Vincent Ng. 2022.
\newblock \href {https://doi.org/10.1609/aaai.v36i11.21496} {Commonsense knowledge reasoning and generation with pre-trained language models: {A} survey}.
\newblock In \emph{Thirty-Sixth {AAAI} Conference on Artificial Intelligence, {AAAI} 2022, Thirty-Fourth Conference on Innovative Applications of Artificial Intelligence, {IAAI} 2022, The Twelveth Symposium on Educational Advances in Artificial Intelligence, {EAAI} 2022 Virtual Event, February 22 - March 1, 2022}, pages 12317--12325. {AAAI} Press.

\bibitem[{Blanco et~al.(2008)Blanco, Castell, and Moldovan}]{blanco-etal-2008-causal}
Eduardo Blanco, Nuria Castell, and Dan Moldovan. 2008.
\newblock \href {http://www.lrec-conf.org/proceedings/lrec2008/pdf/87_paper.pdf} {Causal relation extraction}.
\newblock In \emph{Proceedings of the Sixth International Conference on Language Resources and Evaluation ({LREC}'08)}, Marrakech, Morocco. European Language Resources Association (ELRA).

\bibitem[{Bochman(2003)}]{bochman-2003-logic}
Alexander Bochman. 2003.
\newblock \href {http://ijcai.org/Proceedings/03/Papers/020.pdf} {A logic for causal reasoning}.
\newblock In \emph{IJCAI-03, Proceedings of the Eighteenth International Joint Conference on Artificial Intelligence, Acapulco, Mexico, August 9-15, 2003}, pages 141--146. Morgan Kaufmann.

\bibitem[{Bollen and Pearl(2013)}]{bollen-pearl-2013-eight}
Kenneth~A Bollen and Judea Pearl. 2013.
\newblock \href {https://ftp.cs.ucla.edu/pub/stat_ser/r393-reprint.pdf} {Eight myths about causality and structural equation models}.
\newblock In \emph{Handbook of causal analysis for social research}, pages 301--328. Springer.

\bibitem[{Bommasani et~al.(2021)Bommasani, Hudson, Adeli, Altman, Arora, von Arx, Bernstein, Bohg, Bosselut, Brunskill, Brynjolfsson, Buch, Card, Castellon, Chatterji, Chen, Creel, Davis, Demszky, Donahue, Doumbouya, Durmus, Ermon, Etchemendy, Ethayarajh, Fei{-}Fei, Finn, Gale, Gillespie, Goel, Goodman, Grossman, Guha, Hashimoto, Henderson, Hewitt, Ho, Hong, Hsu, Huang, Icard, Jain, Jurafsky, Kalluri, Karamcheti, Keeling, Khani, Khattab, Koh, Krass, Krishna, Kuditipudi, and et~al.}]{bommasani-etal-2021-opportunities}
Rishi Bommasani, Drew~A. Hudson, Ehsan Adeli, Russ~B. Altman, Simran Arora, Sydney von Arx, Michael~S. Bernstein, Jeannette Bohg, Antoine Bosselut, Emma Brunskill, Erik Brynjolfsson, Shyamal Buch, Dallas Card, Rodrigo Castellon, Niladri~S. Chatterji, Annie~S. Chen, Kathleen Creel, Jared~Quincy Davis, Dorottya Demszky, Chris Donahue, Moussa Doumbouya, Esin Durmus, Stefano Ermon, John Etchemendy, Kawin Ethayarajh, Li~Fei{-}Fei, Chelsea Finn, Trevor Gale, Lauren Gillespie, Karan Goel, Noah~D. Goodman, Shelby Grossman, Neel Guha, Tatsunori Hashimoto, Peter Henderson, John Hewitt, Daniel~E. Ho, Jenny Hong, Kyle Hsu, Jing Huang, Thomas Icard, Saahil Jain, Dan Jurafsky, Pratyusha Kalluri, Siddharth Karamcheti, Geoff Keeling, Fereshte Khani, Omar Khattab, Pang~Wei Koh, Mark~S. Krass, Ranjay Krishna, Rohith Kuditipudi, and et~al. 2021.
\newblock \href {http://arxiv.org/abs/2108.07258} {On the opportunities and risks of foundation models}.
\newblock \emph{CoRR}, abs/2108.07258.

\bibitem[{Bottou et~al.(2013)Bottou, Peters, Candela, Charles, Chickering, Portugaly, Ray, Simard, and Snelson}]{bottou-etal-2013-counterfactual}
L{\'{e}}on Bottou, Jonas Peters, Joaquin~Qui{\~{n}}onero Candela, Denis~Xavier Charles, Max Chickering, Elon Portugaly, Dipankar Ray, Patrice~Y. Simard, and Ed~Snelson. 2013.
\newblock \href {https://doi.org/10.5555/2567709.2567766} {Counterfactual reasoning and learning systems: the example of computational advertising}.
\newblock \emph{J. Mach. Learn. Res.}, 14(1):3207--3260.

\bibitem[{Boué et~al.(2015)Boué, Talikka, Westra, Hayes, Di~Fabio, Park, Schlage, Sewer, Fields, Ansari, Martin, Veljkovic, Kenney, Peitsch, and Hoeng}]{boue-etal-2015-causal}
Stéphanie Boué, Marja Talikka, Jurjen~Willem Westra, William Hayes, Anselmo Di~Fabio, Jennifer Park, Walter~K. Schlage, Alain Sewer, Brett Fields, Sam Ansari, Florian Martin, Emilija Veljkovic, Renee Kenney, Manuel~C. Peitsch, and Julia Hoeng. 2015.
\newblock \href {https://doi.org/10.1093/database/bav030} {{Causal biological network database: a comprehensive platform of causal biological network models focused on the pulmonary and vascular systems}}.
\newblock \emph{Database}, 2015:bav030.

\bibitem[{Bronfenbrenner(1981)}]{bronfenbrenner-1981-causality}
Martin Bronfenbrenner. 1981.
\newblock \href {https://api.semanticscholar.org/CorpusID:157213664} {Causality in economics. john hicks}.
\newblock \emph{Economic Development and Cultural Change}, 29:860--863.

\bibitem[{Brown et~al.(2020)Brown, Mann, Ryder, Subbiah, Kaplan, Dhariwal, Neelakantan, Shyam, Sastry, Askell, Agarwal, Herbert{-}Voss, Krueger, Henighan, Child, Ramesh, Ziegler, Wu, Winter, Hesse, Chen, Sigler, Litwin, Gray, Chess, Clark, Berner, McCandlish, Radford, Sutskever, and Amodei}]{brown-etal-2020-language}
Tom~B. Brown, Benjamin Mann, Nick Ryder, Melanie Subbiah, Jared Kaplan, Prafulla Dhariwal, Arvind Neelakantan, Pranav Shyam, Girish Sastry, Amanda Askell, Sandhini Agarwal, Ariel Herbert{-}Voss, Gretchen Krueger, Tom Henighan, Rewon Child, Aditya Ramesh, Daniel~M. Ziegler, Jeffrey Wu, Clemens Winter, Christopher Hesse, Mark Chen, Eric Sigler, Mateusz Litwin, Scott Gray, Benjamin Chess, Jack Clark, Christopher Berner, Sam McCandlish, Alec Radford, Ilya Sutskever, and Dario Amodei. 2020.
\newblock \href {https://proceedings.neurips.cc/paper/2020/hash/1457c0d6bfcb4967418bfb8ac142f64a-Abstract.html} {Language models are few-shot learners}.
\newblock In \emph{Advances in Neural Information Processing Systems 33: Annual Conference on Neural Information Processing Systems 2020, NeurIPS 2020, December 6-12, 2020, virtual}.

\bibitem[{Bui et~al.(2010)Bui, Nuall{\'{a}}in, Boucher, and Sloot}]{bui-etal-2010-extracting}
Quoc{-}Chinh Bui, Breannd{\'{a}}n~{\'{O}} Nuall{\'{a}}in, Charles~A. Boucher, and Peter M.~A. Sloot. 2010.
\newblock \href {https://doi.org/10.1186/1471-2105-11-101} {Extracting causal relations on {HIV} drug resistance from literature}.
\newblock \emph{{BMC} Bioinform.}, 11:101.

\bibitem[{Calderon et~al.(2022)Calderon, Ben-David, Feder, and Reichart}]{calderon-etal-2022-docogen}
Nitay Calderon, Eyal Ben-David, Amir Feder, and Roi Reichart. 2022.
\newblock \href {https://doi.org/10.18653/v1/2022.acl-long.533} {{D}o{C}o{G}en: {D}omain counterfactual generation for low resource domain adaptation}.
\newblock In \emph{Proceedings of the 60th Annual Meeting of the Association for Computational Linguistics (Volume 1: Long Papers)}, pages 7727--7746, Dublin, Ireland. Association for Computational Linguistics.

\bibitem[{Cao et~al.(2022)Cao, Williamson, and Choi}]{cao-etal-2022-cognitive}
Angela Cao, Gregor Williamson, and Jinho~D. Choi. 2022.
\newblock \href {https://aclanthology.org/2022.law-1.18} {A cognitive approach to annotating causal constructions in a cross-genre corpus}.
\newblock In \emph{Proceedings of the 16th Linguistic Annotation Workshop (LAW-XVI) within LREC2022}, pages 151--159, Marseille, France. European Language Resources Association.

\bibitem[{Cao et~al.(2014)Cao, Zhang, Guo, and Guo}]{cao-etal-2014-mining}
Yanan Cao, Peng Zhang, Jing Guo, and Li~Guo. 2014.
\newblock \href {https://doi.org/10.1016/j.procs.2014.05.043} {Mining large-scale event knowledge from web text}.
\newblock In \emph{Proceedings of the International Conference on Computational Science, {ICCS} 2014, Cairns, Queensland, Australia, 10-12 June, 2014}, volume~29 of \emph{Procedia Computer Science}, pages 478--487. Elsevier.

\bibitem[{Caselli and Vossen(2017)}]{caselli-vossen-2017-event}
Tommaso Caselli and Piek Vossen. 2017.
\newblock \href {https://doi.org/10.18653/v1/W17-2711} {The event {S}tory{L}ine corpus: A new benchmark for causal and temporal relation extraction}.
\newblock In \emph{Proceedings of the Events and Stories in the News Workshop}, pages 77--86, Vancouver, Canada. Association for Computational Linguistics.

\bibitem[{Ceraolo et~al.(2024)Ceraolo, Kharlapenko, Reymond, Mihalcea, Sachan, Sch{\"{o}}lkopf, and Jin}]{ceraolo-etal-2024-causalquest}
Roberto Ceraolo, Dmitrii Kharlapenko, Am{\'{e}}lie Reymond, Rada Mihalcea, Mrinmaya Sachan, Bernhard Sch{\"{o}}lkopf, and Zhijing Jin. 2024.
\newblock \href {https://doi.org/10.48550/ARXIV.2405.20318} {Causalquest: Collecting natural causal questions for {AI} agents}.
\newblock \emph{CoRR}, abs/2405.20318.

\bibitem[{Chalkidis et~al.(2020)Chalkidis, Fergadiotis, Malakasiotis, Aletras, and Androutsopoulos}]{chalkidis-etal-2020-legal}
Ilias Chalkidis, Manos Fergadiotis, Prodromos Malakasiotis, Nikolaos Aletras, and Ion Androutsopoulos. 2020.
\newblock \href {https://doi.org/10.18653/v1/2020.findings-emnlp.261} {{LEGAL}-{BERT}: The muppets straight out of law school}.
\newblock In \emph{Findings of the Association for Computational Linguistics: EMNLP 2020}, pages 2898--2904, Online. Association for Computational Linguistics.

\bibitem[{Chen et~al.(2020)Chen, Li, Yu, Kholy, Ahmed, Gan, Cheng, and Liu}]{chen-etal-2020-uniter}
Yen{-}Chun Chen, Linjie Li, Licheng Yu, Ahmed~El Kholy, Faisal Ahmed, Zhe Gan, Yu~Cheng, and Jingjing Liu. 2020.
\newblock \href {https://doi.org/10.1007/978-3-030-58577-8\_7} {{UNITER:} universal image-text representation learning}.
\newblock In \emph{Computer Vision - {ECCV} 2020 - 16th European Conference, Glasgow, UK, August 23-28, 2020, Proceedings, Part {XXX}}, volume 12375 of \emph{Lecture Notes in Computer Science}, pages 104--120. Springer.

\bibitem[{Chen et~al.(2023)Chen, Gao, Bosselut, Sabharwal, and Richardson}]{chen-etal-2023-disco}
Zeming Chen, Qiyue Gao, Antoine Bosselut, Ashish Sabharwal, and Kyle Richardson. 2023.
\newblock \href {https://doi.org/10.18653/v1/2023.acl-long.302} {{DISCO}: Distilling counterfactuals with large language models}.
\newblock In \emph{Proceedings of the 61st Annual Meeting of the Association for Computational Linguistics (Volume 1: Long Papers)}, pages 5514--5528, Toronto, Canada. Association for Computational Linguistics.

\bibitem[{Chiolero(2019)}]{chiolero-2019-causality}
Arnaud Chiolero. 2019.
\newblock \href {https://doi.org/10.2105/AJPH.2019.305282} {Causality in public health: one word is not enough}.
\newblock \emph{American Journal of Public Health}, 109(10):1319.

\bibitem[{Cho et~al.(2014)Cho, van Merri{\"e}nboer, Gulcehre, Bahdanau, Bougares, Schwenk, and Bengio}]{cho-etal-2014-learning}
Kyunghyun Cho, Bart van Merri{\"e}nboer, Caglar Gulcehre, Dzmitry Bahdanau, Fethi Bougares, Holger Schwenk, and Yoshua Bengio. 2014.
\newblock \href {https://doi.org/10.3115/v1/D14-1179} {Learning phrase representations using {RNN} encoder{--}decoder for statistical machine translation}.
\newblock In \emph{Proceedings of the 2014 Conference on Empirical Methods in Natural Language Processing ({EMNLP})}, pages 1724--1734, Doha, Qatar. Association for Computational Linguistics.

\bibitem[{Choudhry(2020)}]{choudhry-2020-narrative}
Arjun Choudhry. 2020.
\newblock \href {https://vtechworks.lib.vt.edu/handle/10919/98670} {\emph{Narrative Generation to Support Causal Exploration of Directed Graphs}}.
\newblock Ph.D. thesis, Virginia Tech.

\bibitem[{Chowdhery et~al.(2023)Chowdhery, Narang, Devlin, Bosma, Mishra, Roberts, Barham, Chung, Sutton, Gehrmann, Schuh, Shi, Tsvyashchenko, Maynez, Rao, Barnes, Tay, Shazeer, Prabhakaran, Reif, Du, Hutchinson, Pope, Bradbury, Austin, Isard, Gur{-}Ari, Yin, Duke, Levskaya, Ghemawat, Dev, Michalewski, Garcia, Misra, Robinson, Fedus, Zhou, Ippolito, Luan, Lim, Zoph, Spiridonov, Sepassi, Dohan, Agrawal, Omernick, Dai, Pillai, Pellat, Lewkowycz, Moreira, Child, Polozov, Lee, Zhou, Wang, Saeta, Diaz, Firat, Catasta, Wei, Meier{-}Hellstern, Eck, Dean, Petrov, and Fiedel}]{chowdhery-etal-2023-palm}
Aakanksha Chowdhery, Sharan Narang, Jacob Devlin, Maarten Bosma, Gaurav Mishra, Adam Roberts, Paul Barham, Hyung~Won Chung, Charles Sutton, Sebastian Gehrmann, Parker Schuh, Kensen Shi, Sasha Tsvyashchenko, Joshua Maynez, Abhishek Rao, Parker Barnes, Yi~Tay, Noam Shazeer, Vinodkumar Prabhakaran, Emily Reif, Nan Du, Ben Hutchinson, Reiner Pope, James Bradbury, Jacob Austin, Michael Isard, Guy Gur{-}Ari, Pengcheng Yin, Toju Duke, Anselm Levskaya, Sanjay Ghemawat, Sunipa Dev, Henryk Michalewski, Xavier Garcia, Vedant Misra, Kevin Robinson, Liam Fedus, Denny Zhou, Daphne Ippolito, David Luan, Hyeontaek Lim, Barret Zoph, Alexander Spiridonov, Ryan Sepassi, David Dohan, Shivani Agrawal, Mark Omernick, Andrew~M. Dai, Thanumalayan~Sankaranarayana Pillai, Marie Pellat, Aitor Lewkowycz, Erica Moreira, Rewon Child, Oleksandr Polozov, Katherine Lee, Zongwei Zhou, Xuezhi Wang, Brennan Saeta, Mark Diaz, Orhan Firat, Michele Catasta, Jason Wei, Kathy Meier{-}Hellstern, Douglas Eck, Jeff Dean, Slav Petrov, and Noah Fiedel.
  2023.
\newblock \href {http://jmlr.org/papers/v24/22-1144.html} {Palm: Scaling language modeling with pathways}.
\newblock \emph{J. Mach. Learn. Res.}, 24:240:1--240:113.

\bibitem[{Cole et~al.(2006)Cole, Royal, Valtorta, Huhns, and Bowles}]{cole-etal-2006-lightweight}
Stephen~V Cole, Matthew~D Royal, Marco~G Valtorta, Michael~N Huhns, and John~B Bowles. 2006.
\newblock \href {https://ieeexplore.ieee.org/document/1629336} {A lightweight tool for automatically extracting causal relationships from text}.
\newblock In \emph{Proceedings of the IEEE SoutheastCon 2006}, pages 125--129. IEEE.

\bibitem[{Cortes and Vapnik(1995)}]{cortes-vapnik-1965-support}
Corinna Cortes and Vladimir Vapnik. 1995.
\newblock \href {https://doi.org/10.1007/BF00994018} {Support-vector networks}.
\newblock \emph{Mach. Learn.}, 20(3):273--297.

\bibitem[{Cui et~al.(2024)Cui, Milikic, Feng, Ismayilzada, Paul, Bosselut, and Faltings}]{cui-etal-2024-exploring}
Shaobo Cui, Lazar Milikic, Yiyang Feng, Mete Ismayilzada, Debjit Paul, Antoine Bosselut, and Boi Faltings. 2024.
\newblock \href {https://aclanthology.org/2024.findings-acl.384} {Exploring defeasibility in causal reasoning}.
\newblock In \emph{Findings of the Association for Computational Linguistics ACL 2024}, pages 6433--6452, Bangkok, Thailand and virtual meeting. Association for Computational Linguistics.

\bibitem[{Dasgupta et~al.(2018)Dasgupta, Saha, Dey, and Naskar}]{dasgupta-etal-2018-automatic-extraction}
Tirthankar Dasgupta, Rupsa Saha, Lipika Dey, and Abir Naskar. 2018.
\newblock \href {https://doi.org/10.18653/v1/W18-5035} {Automatic extraction of causal relations from text using linguistically informed deep neural networks}.
\newblock In \emph{Proceedings of the 19th Annual {SIG}dial Meeting on Discourse and Dialogue}, pages 306--316, Melbourne, Australia. Association for Computational Linguistics.

\bibitem[{Davis(2023)}]{davis-2023-benchmark}
Ernest Davis. 2023.
\newblock \href {https://doi.org/10.1145/3615355} {Benchmarks for automated commonsense reasoning: A survey}.
\newblock \emph{ACM Comput. Surv.}, 56(4).

\bibitem[{Devlin et~al.(2019)Devlin, Chang, Lee, and Toutanova}]{devlin-etal-2019-bert}
Jacob Devlin, Ming-Wei Chang, Kenton Lee, and Kristina Toutanova. 2019.
\newblock \href {https://doi.org/10.18653/v1/N19-1423} {{BERT}: Pre-training of deep bidirectional transformers for language understanding}.
\newblock In \emph{Proceedings of the 2019 Conference of the North {A}merican Chapter of the Association for Computational Linguistics: Human Language Technologies, Volume 1 (Long and Short Papers)}, pages 4171--4186, Minneapolis, Minnesota. Association for Computational Linguistics.

\bibitem[{Diao et~al.(2023)Diao, Wang, Lin, and Zhang}]{diao-etal-2023-active}
Shizhe Diao, Pengcheng Wang, Yong Lin, and Tong Zhang. 2023.
\newblock \href {https://doi.org/10.48550/ARXIV.2302.12246} {Active prompting with chain-of-thought for large language models}.
\newblock \emph{CoRR}, abs/2302.12246.

\bibitem[{Do et~al.(2011)Do, Chan, and Roth}]{do-etal-2011-minimally}
Quang Do, Yee~Seng Chan, and Dan Roth. 2011.
\newblock \href {https://aclanthology.org/D11-1027} {Minimally supervised event causality identification}.
\newblock In \emph{Proceedings of the 2011 Conference on Empirical Methods in Natural Language Processing}, pages 294--303, Edinburgh, Scotland, UK. Association for Computational Linguistics.

\bibitem[{Doan et~al.(2019)Doan, Yang, Tilak, Li, Zisook, and Torii}]{doan-etal-2019-extracting}
Son Doan, Elly~W. Yang, Sameer~S. Tilak, Peter~W. Li, Daniel~S. Zisook, and Manabu Torii. 2019.
\newblock \href {https://doi.org/10.1186/s12911-019-0785-0} {Extracting health-related causality from twitter messages using natural language processing}.
\newblock \emph{{BMC} Medical Informatics Decis. Mak.}, 19-S(3):71--77.

\bibitem[{Drury et~al.(2022)Drury, Oliveira, and de~Andrade~Lopes}]{drury-etal-2022-survey}
Brett Drury, Hugo~Gon{\c{c}}alo Oliveira, and Alneu de~Andrade~Lopes. 2022.
\newblock \href {https://doi.org/10.1017/S135132492100036X} {A survey of the extraction and applications of causal relations}.
\newblock \emph{Nat. Lang. Eng.}, 28(3):361--400.

\bibitem[{Du et~al.(2022)Du, Ding, Xiong, Liu, and Qin}]{du-etal-2022-e}
Li~Du, Xiao Ding, Kai Xiong, Ting Liu, and Bing Qin. 2022.
\newblock \href {https://doi.org/10.18653/v1/2022.acl-long.33} {e-{CARE}: a new dataset for exploring explainable causal reasoning}.
\newblock In \emph{Proceedings of the 60th Annual Meeting of the Association for Computational Linguistics (Volume 1: Long Papers)}, pages 432--446, Dublin, Ireland. Association for Computational Linguistics.

\bibitem[{Dunietz(2018)}]{dunietz-2018-annotating}
Jesse Dunietz. 2018.
\newblock \href {http://reports-archive.adm.cs.cmu.edu/anon/anon/usr/ftp/usr0/ftp/2018/CMU-CS-18-100.pdf} {Annotating and automatically tagging constructions of causal language}.
\newblock \emph{CARNEGIE MELON UNIVERSITY DISSERTATION}.

\bibitem[{Dunietz et~al.(2017)Dunietz, Levin, and Carbonell}]{dunietz-etal-2017-corpus}
Jesse Dunietz, Lori Levin, and Jaime Carbonell. 2017.
\newblock \href {https://doi.org/10.18653/v1/W17-0812} {The {BEC}au{SE} corpus 2.0: Annotating causality and overlapping relations}.
\newblock In \emph{Proceedings of the 11th Linguistic Annotation Workshop}, pages 95--104, Valencia, Spain. Association for Computational Linguistics.

\bibitem[{Dupr{\'e}(1984)}]{dupre-1984-probabilistic}
John Dupr{\'e}. 1984.
\newblock \href {https://onlinelibrary.wiley.com/doi/abs/10.1111/j.1475-4975.1984.tb00058.x} {Probabilistic causality emancipated}.
\newblock \emph{Midwest Studies in Philosophy}, 9:169--175.

\bibitem[{Eells(1991)}]{eells-1991-probabilistic}
Ellery Eells. 1991.
\newblock \href {https://www.cambridge.org/core/books/probabilistic-causality/5AE128920FF91CC726556FFD1E5C5257} {\emph{Probabilistic causality}}, volume~1.
\newblock Cambridge University Press.

\bibitem[{Eronen(2020)}]{eronen-2020-causal}
Markus~I. Eronen. 2020.
\newblock \href {https://doi.org/https://doi.org/10.1016/j.newideapsych.2020.100785} {Causal discovery and the problem of psychological interventions}.
\newblock \emph{New Ideas in Psychology}, 59:100785.

\bibitem[{Feder et~al.(2022)Feder, Keith, Manzoor, Pryzant, Sridhar, Wood-Doughty, Eisenstein, Grimmer, Reichart, Roberts, Stewart, Veitch, and Yang}]{feder-etal-2022-causal}
Amir Feder, Katherine~A. Keith, Emaad Manzoor, Reid Pryzant, Dhanya Sridhar, Zach Wood-Doughty, Jacob Eisenstein, Justin Grimmer, Roi Reichart, Margaret~E. Roberts, Brandon~M. Stewart, Victor Veitch, and Diyi Yang. 2022.
\newblock \href {https://doi.org/10.1162/tacl_a_00511} {Causal inference in natural language processing: Estimation, prediction, interpretation and beyond}.
\newblock \emph{Transactions of the Association for Computational Linguistics}, 10:1138--1158.

\bibitem[{Feng et~al.(2021)Feng, Zhang, He, Zhang, and Chua}]{feng-etal-2021-empowering}
Fuli Feng, Jizhi Zhang, Xiangnan He, Hanwang Zhang, and Tat-Seng Chua. 2021.
\newblock \href {https://doi.org/10.18653/v1/2021.findings-acl.196} {Empowering language understanding with counterfactual reasoning}.
\newblock In \emph{Findings of the Association for Computational Linguistics: ACL-IJCNLP 2021}, pages 2226--2236, Online. Association for Computational Linguistics.

\bibitem[{Ferguson and Sanford(2008)}]{ferguson-sanford-2008-anomalies}
Heather~J Ferguson and Anthony~J Sanford. 2008.
\newblock \href {https://www.sciencedirect.com/science/article/pii/S0749596X07000770} {Anomalies in real and counterfactual worlds: An eye-movement investigation}.
\newblock \emph{Journal of Memory and Language}, 58(3):609--626.

\bibitem[{Fisher(1936)}]{fisher-1936-design}
Richard Fisher. 1936.
\newblock \href {https://www.ncbi.nlm.nih.gov/pmc/articles/PMC2458144/} {Design of experiments}.
\newblock \emph{British Medical Journal}, 1(3923):554.

\bibitem[{Fitelson and Hitchcock(2011)}]{fitelson-hitchcock-2011-probabilistic}
Branden Fitelson and Christopher Hitchcock. 2011.
\newblock \href {https://philpapers.org/rec/FITPMO} {Probabilistic measures of causal strength}.
\newblock \emph{Causality in the Sciences}, pages 600--627.

\bibitem[{Forbes et~al.(2020)Forbes, Hwang, Shwartz, Sap, and Choi}]{forbes-etal-2020-social}
Maxwell Forbes, Jena~D. Hwang, Vered Shwartz, Maarten Sap, and Yejin Choi. 2020.
\newblock \href {https://doi.org/10.18653/v1/2020.emnlp-main.48} {Social chemistry 101: Learning to reason about social and moral norms}.
\newblock In \emph{Proceedings of the 2020 Conference on Empirical Methods in Natural Language Processing (EMNLP)}, pages 653--670, Online. Association for Computational Linguistics.

\bibitem[{Frohberg and Binder(2022)}]{frohberg-binder-2022-crass}
J{\"o}rg Frohberg and Frank Binder. 2022.
\newblock \href {https://aclanthology.org/2022.lrec-1.229} {{CRASS}: A novel data set and benchmark to test counterfactual reasoning of large language models}.
\newblock In \emph{Proceedings of the Thirteenth Language Resources and Evaluation Conference}, pages 2126--2140, Marseille, France. European Language Resources Association.

\bibitem[{Gao et~al.(2019)Gao, Choubey, and Huang}]{gao-etal-2019-modeling}
Lei Gao, Prafulla~Kumar Choubey, and Ruihong Huang. 2019.
\newblock \href {https://doi.org/10.18653/v1/N19-1179} {Modeling document-level causal structures for event causal relation identification}.
\newblock In \emph{Proceedings of the 2019 Conference of the North {A}merican Chapter of the Association for Computational Linguistics: Human Language Technologies, Volume 1 (Long and Short Papers)}, pages 1808--1817, Minneapolis, Minnesota. Association for Computational Linguistics.

\bibitem[{Garcia(1997)}]{garcia-1997-coatis}
Daniela Garcia. 1997.
\newblock \href {https://doi.org/10.1007/BFb0026799} {Coatis, an {NLP} system to locate expressions of actions connected by causality links}.
\newblock In \emph{Knowledge Acquisition, Modeling and Management, 10th European Workshop, EKAW'97, Sant Feliu de Guixols, Catalonia, Spain, October 15-18, 1997, Proceedings}, volume 1319 of \emph{Lecture Notes in Computer Science}, pages 347--352. Springer.

\bibitem[{Geiger et~al.(1989)Geiger, Verma, and Pearl}]{geiger-etal-1989-d}
Dan Geiger, Thomas Verma, and Judea Pearl. 1989.
\newblock \href {https://dslpitt.org/uai/displayArticleDetails.jsp?mmnu=1\&smnu=2\&article\_id=1872\&proceeding\_id=1005} {d-separation: From theorems to algorithms}.
\newblock In \emph{{UAI} '89: Proceedings of the Fifth Annual Conference on Uncertainty in Artificial Intelligence, Windsor, Ontario, Canada, August 18-20, 1989}, pages 139--148. North-Holland.

\bibitem[{Girju(2003)}]{girju-2003-automatic}
Roxana Girju. 2003.
\newblock \href {https://doi.org/10.3115/1119312.1119322} {Automatic detection of causal relations for question answering}.
\newblock In \emph{Proceedings of the {ACL} 2003 Workshop on Multilingual Summarization and Question Answering}, pages 76--83, Sapporo, Japan. Association for Computational Linguistics.

\bibitem[{Girju et~al.(2007)Girju, Nakov, Nastase, Szpakowicz, Turney, and Yuret}]{girju-etal-2007-semeval}
Roxana Girju, Preslav Nakov, Vivi Nastase, Stan Szpakowicz, Peter Turney, and Deniz Yuret. 2007.
\newblock \href {https://aclanthology.org/S07-1003} {{S}em{E}val-2007 task 04: Classification of semantic relations between nominals}.
\newblock In \emph{Proceedings of the Fourth International Workshop on Semantic Evaluations ({S}em{E}val-2007)}, pages 13--18, Prague, Czech Republic. Association for Computational Linguistics.

\bibitem[{Glymour et~al.(2019)Glymour, Zhang, and Spirtes}]{glymour-etal-2019-review}
Clark Glymour, Kun Zhang, and Peter Spirtes. 2019.
\newblock \href {https://doi.org/10.3389/fgene.2019.00524} {Review of causal discovery methods based on graphical models}.
\newblock \emph{Frontiers in genetics}, 10:524.

\bibitem[{Goffrier et~al.(2023)Goffrier, Maystre, and Gilligan{-}Lee}]{van-etal-2023-estimating}
Graham~Van Goffrier, Lucas Maystre, and Ciar{\'{a}}n~Mark Gilligan{-}Lee. 2023.
\newblock \href {https://proceedings.mlr.press/v213/goffrier23a.html} {Estimating long-term causal effects from short-term experiments and long-term observational data with unobserved confounding}.
\newblock In \emph{Conference on Causal Learning and Reasoning, CLeaR 2023, 11-14 April 2023, Amazon Development Center, T{\"{u}}bingen, Germany, April 11-14, 2023}, volume 213 of \emph{Proceedings of Machine Learning Research}, pages 791--813. {PMLR}.

\bibitem[{Good(1961)}]{good-1961-causal}
Irving~J Good. 1961.
\newblock \href {https://www.jstor.org/stable/685131} {A causal calculus (i)}.
\newblock \emph{The British journal for the philosophy of science}, 11(44):305--318.

\bibitem[{Goodman(1947)}]{goodman-1947-problem}
Nelson Goodman. 1947.
\newblock \href {https://www.jstor.org/stable/2019988} {The problem of counterfactual conditionals}.
\newblock \emph{The Journal of Philosophy}, 44(5):113--128.

\bibitem[{Granger(1988)}]{granger-1988-some}
Clive Granger. 1988.
\newblock \href {https://www.sciencedirect.com/science/article/pii/0304407688900450} {Some recent development in a concept of causality}.
\newblock \emph{Journal of Econometrics}, 39(1-2):199--211.

\bibitem[{Grunbaum(1952)}]{grunbaum-1952-causality}
Adolf Grunbaum. 1952.
\newblock \href {http://www.jstor.org/stable/27826461} {Causality and the science of human behavior}.
\newblock \emph{American Scientist}, 40(4):665--689.

\bibitem[{Gurulingappa et~al.(2012)Gurulingappa, Rajput, Roberts, Fluck, Hofmann{-}Apitius, and Toldo}]{gurulingappa-etal-2012-development}
Harsha Gurulingappa, Abdul~Mateen Rajput, Angus Roberts, Juliane Fluck, Martin Hofmann{-}Apitius, and Luca Toldo. 2012.
\newblock \href {https://doi.org/10.1016/j.jbi.2012.04.008} {Development of a benchmark corpus to support the automatic extraction of drug-related adverse effects from medical case reports}.
\newblock \emph{J. Biomed. Informatics}, 45(5):885--892.

\bibitem[{Hair~Jr et~al.(2021)Hair~Jr, Hult, Ringle, Sarstedt, Danks, Ray, Hair, Hult, Ringle, Sarstedt et~al.}]{hair-etal-2021-introduction}
Joseph~F Hair~Jr, G~Tomas~M Hult, Christian~M Ringle, Marko Sarstedt, Nicholas~P Danks, Soumya Ray, Joseph~F Hair, G~Tomas~M Hult, Christian~M Ringle, Marko Sarstedt, et~al. 2021.
\newblock \href {https://link.springer.com/chapter/10.1007/978-3-030-80519-7_1} {An introduction to structural equation modeling}.
\newblock \emph{Partial least squares structural equation modeling (PLS-SEM) using R: a workbook}, pages 1--29.

\bibitem[{Hartshorne(2014)}]{hartshorne-2014-implicit}
Joshua~K. Hartshorne. 2014.
\newblock \href {https://doi.org/10.1080/01690965.2013.796396} {What is implicit causality?}
\newblock \emph{Language, Cognition and Neuroscience}, 29(7):804--824.

\bibitem[{Hassanzadeh et~al.(2020)Hassanzadeh, Bhattacharjya, Feblowitz, Srinivas, Perrone, Sohrabi, and Katz}]{hassanzadeh-etal-2020-causal}
Oktie Hassanzadeh, Debarun Bhattacharjya, Mark Feblowitz, Kavitha Srinivas, Michael Perrone, Shirin Sohrabi, and Michael Katz. 2020.
\newblock \href {https://doi.org/10.1609/aaai.v34i09.7092} {Causal knowledge extraction through large-scale text mining}.
\newblock In \emph{The Thirty-Fourth {AAAI} Conference on Artificial Intelligence, {AAAI} 2020, The Thirty-Second Innovative Applications of Artificial Intelligence Conference, {IAAI} 2020, The Tenth {AAAI} Symposium on Educational Advances in Artificial Intelligence, {EAAI} 2020, New York, NY, USA, February 7-12, 2020}, pages 13610--13611. {AAAI} Press.

\bibitem[{Hayduk et~al.(2003)Hayduk, Cummings, Stratkotter, Nimmo, Grygoryev, Dosman, Gillespie, Pazderka-Robinson, and Boadu}]{hayduk-etal-2003-structural}
Leslie Hayduk, Greta Cummings, Rainer Stratkotter, Melanie Nimmo, Kostyantyn Grygoryev, Donna Dosman, Michael Gillespie, Hannah Pazderka-Robinson, and Kwame Boadu. 2003.
\newblock \href {https://doi.org/10.1207/S15328007SEM1002\_8} {Pearl's d-separation: One more step into causal thinking}.
\newblock \emph{Structural Equation Modeling: A Multidisciplinary Journal}, 10(2):289--311.

\bibitem[{He et~al.(2021)He, Liu, Gao, and Chen}]{he-etal-2021-deberta}
Pengcheng He, Xiaodong Liu, Jianfeng Gao, and Weizhu Chen. 2021.
\newblock \href {https://openreview.net/forum?id=XPZIaotutsD} {Deberta: decoding-enhanced bert with disentangled attention}.
\newblock In \emph{9th International Conference on Learning Representations, {ICLR} 2021, Virtual Event, Austria, May 3-7, 2021}. OpenReview.net.

\bibitem[{Heckerman(2008)}]{heckerman-2008-tutorial}
David Heckerman. 2008.
\newblock \href {https://doi.org/10.1007/978-3-540-85066-3\_3} {A tutorial on learning with bayesian networks}.
\newblock In Dawn~E. Holmes and Lakhmi~C. Jain, editors, \emph{Innovations in Bayesian Networks: Theory and Applications}, volume 156 of \emph{Studies in Computational Intelligence}, pages 33--82. Springer.

\bibitem[{Heindorf et~al.(2020)Heindorf, Scholten, Wachsmuth, Ngomo, and Potthast}]{heindorf-etal-2020-causenet}
Stefan Heindorf, Yan Scholten, Henning Wachsmuth, Axel{-}Cyrille~Ngonga Ngomo, and Martin Potthast. 2020.
\newblock \href {https://doi.org/10.1145/3340531.3412763} {Causenet: Towards a causality graph extracted from the web}.
\newblock In \emph{{CIKM} '20: The 29th {ACM} International Conference on Information and Knowledge Management, Virtual Event, Ireland, October 19-23, 2020}, pages 3023--3030. {ACM}.

\bibitem[{Hendrickx et~al.(2010)Hendrickx, Kim, Kozareva, Nakov, {\'O}~S{\'e}aghdha, Pad{\'o}, Pennacchiotti, Romano, and Szpakowicz}]{hendrickx-etal-2010-semeval}
Iris Hendrickx, Su~Nam Kim, Zornitsa Kozareva, Preslav Nakov, Diarmuid {\'O}~S{\'e}aghdha, Sebastian Pad{\'o}, Marco Pennacchiotti, Lorenza Romano, and Stan Szpakowicz. 2010.
\newblock \href {https://aclanthology.org/S10-1006} {{S}em{E}val-2010 task 8: Multi-way classification of semantic relations between pairs of nominals}.
\newblock In \emph{Proceedings of the 5th International Workshop on Semantic Evaluation}, pages 33--38, Uppsala, Sweden. Association for Computational Linguistics.

\bibitem[{Hern{\'a}n and Robins(2010)}]{hernan-robins-2010-causal}
Miguel~A Hern{\'a}n and James~M Robins. 2010.
\newblock \href {https://www.hsph.harvard.edu/miguel-hernan/causal-inference-book/} {Causal inference}.

\bibitem[{Hidey and McKeown(2016)}]{hidey-mckeown-2016-identifying}
Christopher Hidey and Kathy McKeown. 2016.
\newblock \href {https://doi.org/10.18653/v1/P16-1135} {Identifying causal relations using parallel {W}ikipedia articles}.
\newblock In \emph{Proceedings of the 54th Annual Meeting of the Association for Computational Linguistics (Volume 1: Long Papers)}, pages 1424--1433, Berlin, Germany. Association for Computational Linguistics.

\bibitem[{Hitchcock(1997)}]{hitchcock-1997-probabilistic}
Christopher Hitchcock. 1997.
\newblock \href {https://plato.stanford.edu/entries/causation-probabilistic/} {Probabilistic causation}.

\bibitem[{Hochreiter and Schmidhuber(1997)}]{hochreiter-etal-1997-long}
Sepp Hochreiter and J{\"{u}}rgen Schmidhuber. 1997.
\newblock \href {https://doi.org/10.1162/NECO.1997.9.8.1735} {Long short-term memory}.
\newblock \emph{Neural Comput.}, 9(8):1735--1780.

\bibitem[{Hocutt(1974)}]{hocutt-1974-aristotle}
Max Hocutt. 1974.
\newblock \href {https://www.jstor.org/stable/3750237} {Aristotle's four becauses}.
\newblock \emph{Philosophy}, 49(190):385--399.

\bibitem[{Hoekstra and Breuker(2007)}]{hoekstra-breuker-2007-commonsense}
Rinke Hoekstra and Joost Breuker. 2007.
\newblock \href {https://doi.org/10.1007/S10506-007-9033-5} {Commonsense causal explanation in a legal domain}.
\newblock \emph{Artif. Intell. Law}, 15(3):281--299.

\bibitem[{H{\"o}fer et~al.(2004)H{\"o}fer, Przyrembel, and Verleger}]{hofer-etal-2004-new}
Thomas H{\"o}fer, Hildegard Przyrembel, and Silvia Verleger. 2004.
\newblock \href {https://doi.org/10.1111/j.1365-3016.2003.00534.x} {New evidence for the theory of the stork}.
\newblock \emph{Paediatric and perinatal epidemiology}, 18(1):88--92.

\bibitem[{Holland(1986)}]{holland-1986-statistics}
Paul~W Holland. 1986.
\newblock \href {https://www.jstor.org/stable/2289064} {Statistics and causal inference}.
\newblock \emph{Journal of the American statistical Association}, 81(396):945--960.

\bibitem[{Hoover(2006)}]{hoover-2006-causality}
Kevin~D. Hoover. 2006.
\newblock \href {https://api.semanticscholar.org/CorpusID:45489459} {Causality in economics and econometrics}.
\newblock \emph{History of Finance eJournal}.

\bibitem[{Ikuta et~al.(2014)Ikuta, Styler, Hamang, O{'}Gorman, and Palmer}]{ikuta-etal-2014-challenges}
Rei Ikuta, Will Styler, Mariah Hamang, Tim O{'}Gorman, and Martha Palmer. 2014.
\newblock \href {https://doi.org/10.3115/v1/W14-2903} {Challenges of adding causation to richer event descriptions}.
\newblock In \emph{Proceedings of the Second Workshop on {EVENTS}: Definition, Detection, Coreference, and Representation}, pages 12--20, Baltimore, Maryland, USA. Association for Computational Linguistics.

\bibitem[{Imbens et~al.(2022)Imbens, Kallus, Mao, and Wang}]{imbens-etal-2022-long}
Guido Imbens, Nathan Kallus, Xiaojie Mao, and Yuhao Wang. 2022.
\newblock \href {https://arxiv.org/abs/2202.07234} {Long-term causal inference under persistent confounding via data combination}.
\newblock \emph{arXiv preprint arXiv:2202.07234}.

\bibitem[{Inui et~al.(2003)Inui, Inui, and Matsumoto}]{inui-etal-2003-kinds}
Takashi Inui, Kentaro Inui, and Yuji Matsumoto. 2003.
\newblock \href {https://doi.org/10.1007/978-3-540-39644-4\_16} {What kinds and amounts of causal knowledge can be acquired from text by using connective markers as clues?}
\newblock In \emph{Discovery Science, 6th International Conference, {DS} 2003, Sapporo, Japan, October 17-19,2003, Proceedings}, volume 2843 of \emph{Lecture Notes in Computer Science}, pages 180--193. Springer.

\bibitem[{Inui et~al.(2005)Inui, Inui, and Matsumoto}]{inui-etal-2005-acquiring}
Takashi Inui, Kentaro Inui, and Yuji Matsumoto. 2005.
\newblock \href {https://doi.org/10.1145/1113308.1113313} {Acquiring causal knowledge from text using the connective marker \emph{tame}}.
\newblock \emph{{ACM} Trans. Asian Lang. Inf. Process.}, 4(4):435--474.

\bibitem[{Ishii et~al.(2010)Ishii, Ma, and Yoshikawa}]{ishii-etal-2010-causal}
Hiroshi Ishii, Qiang Ma, and Masatoshi Yoshikawa. 2010.
\newblock \href {https://doi.org/10.1109/HICSS.2010.97} {Causal network construction to support understanding of news}.
\newblock In \emph{43rd Hawaii International International Conference on Systems Science {(HICSS-43} 2010), Proceedings, 5-8 January 2010, Koloa, Kauai, HI, {USA}}, pages 1--10. {IEEE} Computer Society.

\bibitem[{Ittoo and Bouma(2011)}]{ittoo-bouma-2011-extracting}
Ashwin Ittoo and Gosse Bouma. 2011.
\newblock \href {https://doi.org/10.1007/978-3-642-22327-3\_6} {Extracting explicit and implicit causal relations from sparse, domain-specific texts}.
\newblock In \emph{Natural Language Processing and Information Systems - 16th International Conference on Applications of Natural Language to Information Systems, {NLDB} 2011, Alicante, Spain, June 28-30, 2011. Proceedings}, volume 6716 of \emph{Lecture Notes in Computer Science}, pages 52--63. Springer.

\bibitem[{Jiang et~al.(2023)Jiang, Sablayrolles, Mensch, Bamford, Chaplot, de~Las~Casas, Bressand, Lengyel, Lample, Saulnier, Lavaud, Lachaux, Stock, Scao, Lavril, Wang, Lacroix, and Sayed}]{jiang-etal-2023-mistral}
Albert~Q. Jiang, Alexandre Sablayrolles, Arthur Mensch, Chris Bamford, Devendra~Singh Chaplot, Diego de~Las~Casas, Florian Bressand, Gianna Lengyel, Guillaume Lample, Lucile Saulnier, L{\'{e}}lio~Renard Lavaud, Marie{-}Anne Lachaux, Pierre Stock, Teven~Le Scao, Thibaut Lavril, Thomas Wang, Timoth{\'{e}}e Lacroix, and William~El Sayed. 2023.
\newblock \href {https://doi.org/10.48550/ARXIV.2310.06825} {Mistral 7b}.
\newblock \emph{CoRR}, abs/2310.06825.

\bibitem[{Jin et~al.(2020)Jin, Wang, Luo, Huang, and Gu}]{jin-etal-2020-inter}
Xianxian Jin, Xinzhi Wang, Xiangfeng Luo, Subin Huang, and Shengwei Gu. 2020.
\newblock \href {https://doi.org/10.1007/978-3-030-47426-3\_57} {Inter-sentence and implicit causality extraction from chinese corpus}.
\newblock In \emph{Advances in Knowledge Discovery and Data Mining - 24th Pacific-Asia Conference, {PAKDD} 2020, Singapore, May 11-14, 2020, Proceedings, Part {I}}, volume 12084 of \emph{Lecture Notes in Computer Science}, pages 739--751. Springer.

\bibitem[{Jin et~al.(2023{\natexlab{a}})Jin, Chen, Leeb, Gresele, Kamal, LYU, Blin, Adauto, Kleiman-Weiner, Sachan, and Sch{\"o}lkopf}]{jin2023cladder}
Zhijing Jin, Yuen Chen, Felix Leeb, Luigi Gresele, Ojasv Kamal, Zhiheng LYU, Kevin Blin, Fernando~Gonzalez Adauto, Max Kleiman-Weiner, Mrinmaya Sachan, and Bernhard Sch{\"o}lkopf. 2023{\natexlab{a}}.
\newblock \href {https://openreview.net/forum?id=e2wtjx0Yqu} {{CL}adder: A benchmark to assess causal reasoning capabilities of language models}.
\newblock In \emph{Thirty-seventh Conference on Neural Information Processing Systems}.

\bibitem[{Jin et~al.(2023{\natexlab{b}})Jin, Chen, Leeb, Gresele, Kamal, Lyu, Blin, Gonzalez, Kleiman-Weiner, Sachan, and Sch{\"{o}}lkopf}]{jin-etal-2023-cladder}
Zhijing Jin, Yuen Chen, Felix Leeb, Luigi Gresele, Ojasv Kamal, Zhiheng Lyu, Kevin Blin, Fernando Gonzalez, Max Kleiman-Weiner, Mrinmaya Sachan, and Bernhard Sch{\"{o}}lkopf. 2023{\natexlab{b}}.
\newblock \href {https://openreview.net/forum?id=e2wtjx0Yqu} {{CL}adder: {A}ssessing causal reasoning in language models}.
\newblock In \emph{NeurIPS}.

\bibitem[{Jin et~al.(2022)Jin, Lalwani, Vaidhya, Shen, Ding, Lyu, Sachan, Mihalcea, and Sch{\"{o}}lkopf}]{jin-etal-2022-logical}
Zhijing Jin, Abhinav Lalwani, Tejas Vaidhya, Xiaoyu Shen, Yiwen Ding, Zhiheng Lyu, Mrinmaya Sachan, Rada Mihalcea, and Bernhard Sch{\"{o}}lkopf. 2022.
\newblock \href {https://arxiv.org/abs/2202.13758} {Logical fallacy detection}.
\newblock In \emph{Findings of the Association for Computational Linguistics: EMNLP 2022}, pages 7180--7198, Abu Dhabi, United Arab Emirates. Association for Computational Linguistics.

\bibitem[{Jin et~al.(2024)Jin, Liu, Lyu, Poff, Sachan, Mihalcea, Diab, and Sch{\"{o}}lkopf}]{jin2024large}
Zhijing Jin, Jiarui Liu, Zhiheng Lyu, Spencer Poff, Mrinmaya Sachan, Rada Mihalcea, Mona~T. Diab, and Bernhard Sch{\"{o}}lkopf. 2024.
\newblock \href {https://openreview.net/pdf?id=vqIH0ObdqL} {Can large language models infer causation from correlation?}
\newblock In \emph{The Twelfth International Conference on Learning Representations, {ICLR} 2024}. OpenReview.net.

\bibitem[{Kaplan et~al.(2020)Kaplan, McCandlish, Henighan, Brown, Chess, Child, Gray, Radford, Wu, and Amodei}]{kaplan-etal-2020-scaling}
Jared Kaplan, Sam McCandlish, Tom Henighan, Tom~B. Brown, Benjamin Chess, Rewon Child, Scott Gray, Alec Radford, Jeffrey Wu, and Dario Amodei. 2020.
\newblock \href {http://arxiv.org/abs/2001.08361} {Scaling laws for neural language models}.
\newblock \emph{CoRR}, abs/2001.08361.

\bibitem[{Khoo et~al.(2000)Khoo, Chan, and Niu}]{khoo-etal-2000-extracting}
Christopher S.~G. Khoo, Syin Chan, and Yun Niu. 2000.
\newblock \href {https://doi.org/10.3115/1075218.1075261} {Extracting causal knowledge from a medical database using graphical patterns}.
\newblock In \emph{Proceedings of the 38th Annual Meeting of the Association for Computational Linguistics}, pages 336--343, Hong Kong. Association for Computational Linguistics.

\bibitem[{Khoo et~al.(1998)Khoo, Kornfilt, Oddy, and Myaeng}]{khoo-etal-1998-automatic}
Christopher~SG Khoo, Jaklin Kornfilt, Robert~N Oddy, and Sung~Hyon Myaeng. 1998.
\newblock \href {https://jglobal.jst.go.jp/en/detail?JGLOBAL_ID=200902119293860954} {Automatic extraction of cause-effect information from newspaper text without knowledge-based inferencing}.
\newblock \emph{Literary and linguistic computing}, 13(4):177--186.

\bibitem[{Kiciman et~al.(2023)Kiciman, Ness, Sharma, and Tan}]{kiciman-etal-2023-causal}
Emre Kiciman, Robert Ness, Amit Sharma, and Chenhao Tan. 2023.
\newblock \href {https://doi.org/10.48550/arXiv.2305.00050} {Causal reasoning and large language models: Opening a new frontier for causality}.
\newblock \emph{CoRR}, abs/2305.00050.

\bibitem[{Kim et~al.(2022)Kim, Zala, and Bansal}]{kim-etal-2022-cosim}
Hyounghun Kim, Abhay Zala, and Mohit Bansal. 2022.
\newblock \href {https://doi.org/10.18653/v1/2022.naacl-main.66} {{C}o{SI}m: Commonsense reasoning for counterfactual scene imagination}.
\newblock In \emph{Proceedings of the 2022 Conference of the North American Chapter of the Association for Computational Linguistics: Human Language Technologies}, pages 911--923, Seattle, United States. Association for Computational Linguistics.

\bibitem[{Kim et~al.(2023)Kim, Hessel, Jiang, West, Lu, Yu, Zhou, Bras, Alikhani, Kim, Sap, and Choi}]{kim-etal-2023-soda}
Hyunwoo Kim, Jack Hessel, Liwei Jiang, Peter West, Ximing Lu, Youngjae Yu, Pei Zhou, Ronan Bras, Malihe Alikhani, Gunhee Kim, Maarten Sap, and Yejin Choi. 2023.
\newblock \href {https://doi.org/10.18653/v1/2023.emnlp-main.799} {{SODA}: Million-scale dialogue distillation with social commonsense contextualization}.
\newblock In \emph{Proceedings of the 2023 Conference on Empirical Methods in Natural Language Processing}, pages 12930--12949, Singapore. Association for Computational Linguistics.

\bibitem[{Kim(2014)}]{kim-2014-convolutional}
Yoon Kim. 2014.
\newblock \href {https://doi.org/10.3115/v1/D14-1181} {Convolutional neural networks for sentence classification}.
\newblock In \emph{Proceedings of the 2014 Conference on Empirical Methods in Natural Language Processing ({EMNLP})}, pages 1746--1751, Doha, Qatar. Association for Computational Linguistics.

\bibitem[{Kratenko(2022)}]{kratenko-2022-problem}
MV~Kratenko. 2022.
\newblock \href {https://doi.org/10.17116/sudmed20226501162} {The problem of uncertainty of causality in" medical cases" and ways to solve it.(regarding the evidence level of expert opinion)}.
\newblock \emph{Sudebno-meditsinskaia Ekspertiza}, 65(1):62--66.

\bibitem[{Kruengkrai et~al.(2017)Kruengkrai, Torisawa, Hashimoto, Kloetzer, Oh, and Tanaka}]{kruengkrai-etal-2017-improving}
Canasai Kruengkrai, Kentaro Torisawa, Chikara Hashimoto, Julien Kloetzer, Jong{-}Hoon Oh, and Masahiro Tanaka. 2017.
\newblock \href {https://doi.org/10.1609/AAAI.V31I1.11005} {Improving event causality recognition with multiple background knowledge sources using multi-column convolutional neural networks}.
\newblock In \emph{Proceedings of the Thirty-First {AAAI} Conference on Artificial Intelligence, February 4-9, 2017, San Francisco, California, {USA}}, pages 3466--3473. {AAAI} Press.

\bibitem[{Kyriakakis et~al.(2019)Kyriakakis, Androutsopoulos, Saudabayev, and Gin{\'e}s~i Ametll{\'e}}]{kyriakakis-etal-2019-transfer}
Manolis Kyriakakis, Ion Androutsopoulos, Artur Saudabayev, and Joan Gin{\'e}s~i Ametll{\'e}. 2019.
\newblock \href {https://doi.org/10.18653/v1/W19-5031} {Transfer learning for causal sentence detection}.
\newblock In \emph{Proceedings of the 18th BioNLP Workshop and Shared Task}, pages 292--297, Florence, Italy. Association for Computational Linguistics.

\bibitem[{Lan et~al.(2020)Lan, Chen, Goodman, Gimpel, Sharma, and Soricut}]{lan-etal-2020-albert}
Zhenzhong Lan, Mingda Chen, Sebastian Goodman, Kevin Gimpel, Piyush Sharma, and Radu Soricut. 2020.
\newblock \href {https://openreview.net/forum?id=H1eA7AEtvS} {{ALBERT:} {A} lite {BERT} for self-supervised learning of language representations}.
\newblock In \emph{8th International Conference on Learning Representations, {ICLR} 2020, Addis Ababa, Ethiopia, April 26-30, 2020}. OpenReview.net.

\bibitem[{Lee et~al.(2019)Lee, Yoon, Kim, Kim, Kim, So, and Kang}]{lee-etal-2019-biobert}
Jinhyuk Lee, Wonjin Yoon, Sungdong Kim, Donghyeon Kim, Sunkyu Kim, Chan~Ho So, and Jaewoo Kang. 2019.
\newblock \href {https://api.semanticscholar.org/CorpusID:59291975} {Biobert: a pre-trained biomedical language representation model for biomedical text mining}.
\newblock \emph{Bioinformatics}, 36:1234 -- 1240.

\bibitem[{Lewis(1973)}]{lewis-1973-counterfactuals}
David~K. Lewis. 1973.
\newblock \href {https://philpapers.org/rec/LEWC-19} {\emph{Counterfactuals}}.
\newblock Blackwell, Malden, Mass.

\bibitem[{Lewis et~al.(2020)Lewis, Liu, Goyal, Ghazvininejad, Mohamed, Levy, Stoyanov, and Zettlemoyer}]{lewis-etal-2020-bart}
Mike Lewis, Yinhan Liu, Naman Goyal, Marjan Ghazvininejad, Abdelrahman Mohamed, Omer Levy, Veselin Stoyanov, and Luke Zettlemoyer. 2020.
\newblock \href {https://doi.org/10.18653/v1/2020.acl-main.703} {{BART}: Denoising sequence-to-sequence pre-training for natural language generation, translation, and comprehension}.
\newblock In \emph{Proceedings of the 58th Annual Meeting of the Association for Computational Linguistics}, pages 7871--7880, Online. Association for Computational Linguistics.

\bibitem[{Li et~al.(2023)Li, Yu, and Ettinger}]{li-etal-2023-counterfactual}
Jiaxuan Li, Lang Yu, and Allyson Ettinger. 2023.
\newblock \href {https://doi.org/10.18653/v1/2023.acl-short.70} {Counterfactual reasoning: Testing language models{'} understanding of hypothetical scenarios}.
\newblock In \emph{Proceedings of the 61st Annual Meeting of the Association for Computational Linguistics (Volume 2: Short Papers)}, pages 804--815, Toronto, Canada. Association for Computational Linguistics.

\bibitem[{Li and Mao(2019)}]{li-mao-2019-knowledge}
Pengfei Li and Kezhi Mao. 2019.
\newblock \href {https://doi.org/10.1016/J.ESWA.2018.08.009} {Knowledge-oriented convolutional neural network for causal relation extraction from natural language texts}.
\newblock \emph{Expert Syst. Appl.}, 115:512--523.

\bibitem[{Li et~al.(2020{\natexlab{a}})Li, Yin, Li, Zhang, Hu, Zhang, Wang, Hu, Dong, Wei, Choi, and Gao}]{li-etal-2020-oscar}
Xiujun Li, Xi~Yin, Chunyuan Li, Pengchuan Zhang, Xiaowei Hu, Lei Zhang, Lijuan Wang, Houdong Hu, Li~Dong, Furu Wei, Yejin Choi, and Jianfeng Gao. 2020{\natexlab{a}}.
\newblock \href {https://doi.org/10.1007/978-3-030-58577-8\_8} {Oscar: Object-semantics aligned pre-training for vision-language tasks}.
\newblock In \emph{Computer Vision - {ECCV} 2020 - 16th European Conference, Glasgow, UK, August 23-28, 2020, Proceedings, Part {XXX}}, volume 12375 of \emph{Lecture Notes in Computer Science}, pages 121--137. Springer.

\bibitem[{Li et~al.(2021)Li, Li, Zou, and Ren}]{li-etal-2021-causality}
Zhaoning Li, Qi~Li, Xiaotian Zou, and Jiangtao Ren. 2021.
\newblock \href {https://doi.org/10.1016/j.neucom.2020.08.078} {Causality extraction based on self-attentive bilstm-crf with transferred embeddings}.
\newblock \emph{Neurocomputing}, 423:207--219.

\bibitem[{Li et~al.(2020{\natexlab{b}})Li, Ding, Liu, Hu, and Durme}]{li-etal-2021-guided}
Zhongyang Li, Xiao Ding, Ting Liu, J.~Edward Hu, and Benjamin~Van Durme. 2020{\natexlab{b}}.
\newblock \href {https://doi.org/10.24963/ijcai.2020/502} {Guided generation of cause and effect}.
\newblock In \emph{Proceedings of the Twenty-Ninth International Joint Conference on Artificial Intelligence, {IJCAI} 2020}, pages 3629--3636. ijcai.org.

\bibitem[{Liang et~al.(2022)Liang, Bommasani, Lee, Tsipras, Soylu, Yasunaga, Zhang, Narayanan, Wu, Kumar, Newman, Yuan, Yan, Zhang, Cosgrove, Manning, R{\'{e}}, Acosta{-}Navas, Hudson, Zelikman, Durmus, Ladhak, Rong, Ren, Yao, Wang, Santhanam, Orr, Zheng, Y{\"{u}}ksekg{\"{o}}n{\"{u}}l, Suzgun, Kim, Guha, Chatterji, Khattab, Henderson, Huang, Chi, Xie, Santurkar, Ganguli, Hashimoto, Icard, Zhang, Chaudhary, Wang, Li, Mai, Zhang, and Koreeda}]{liang-etal-2022-holistic}
Percy Liang, Rishi Bommasani, Tony Lee, Dimitris Tsipras, Dilara Soylu, Michihiro Yasunaga, Yian Zhang, Deepak Narayanan, Yuhuai Wu, Ananya Kumar, Benjamin Newman, Binhang Yuan, Bobby Yan, Ce~Zhang, Christian Cosgrove, Christopher~D. Manning, Christopher R{\'{e}}, Diana Acosta{-}Navas, Drew~A. Hudson, Eric Zelikman, Esin Durmus, Faisal Ladhak, Frieda Rong, Hongyu Ren, Huaxiu Yao, Jue Wang, Keshav Santhanam, Laurel~J. Orr, Lucia Zheng, Mert Y{\"{u}}ksekg{\"{o}}n{\"{u}}l, Mirac Suzgun, Nathan Kim, Neel Guha, Niladri~S. Chatterji, Omar Khattab, Peter Henderson, Qian Huang, Ryan Chi, Sang~Michael Xie, Shibani Santurkar, Surya Ganguli, Tatsunori Hashimoto, Thomas Icard, Tianyi Zhang, Vishrav Chaudhary, William Wang, Xuechen Li, Yifan Mai, Yuhui Zhang, and Yuta Koreeda. 2022.
\newblock \href {https://doi.org/10.48550/ARXIV.2211.09110} {Holistic evaluation of language models}.
\newblock \emph{CoRR}, abs/2211.09110.

\bibitem[{Liu et~al.(2019)Liu, Ott, Goyal, Du, Joshi, Chen, Levy, Lewis, Zettlemoyer, and Stoyanov}]{liu-etal-2019-roberta}
Yinhan Liu, Myle Ott, Naman Goyal, Jingfei Du, Mandar Joshi, Danqi Chen, Omer Levy, Mike Lewis, Luke Zettlemoyer, and Veselin Stoyanov. 2019.
\newblock \href {http://arxiv.org/abs/1907.11692} {Roberta: {A} robustly optimized {BERT} pretraining approach}.
\newblock \emph{CoRR}, abs/1907.11692.

\bibitem[{Lu et~al.(2019)Lu, Batra, Parikh, and Lee}]{lu-etal-2019-vilbert}
Jiasen Lu, Dhruv Batra, Devi Parikh, and Stefan Lee. 2019.
\newblock \href {https://proceedings.neurips.cc/paper/2019/hash/c74d97b01eae257e44aa9d5bade97baf-Abstract.html} {Vilbert: Pretraining task-agnostic visiolinguistic representations for vision-and-language tasks}.
\newblock In \emph{Advances in Neural Information Processing Systems 32: Annual Conference on Neural Information Processing Systems 2019, NeurIPS 2019, December 8-14, 2019, Vancouver, BC, Canada}, pages 13--23.

\bibitem[{Luo et~al.(2016)Luo, Sha, Zhu, Hwang, and Wang}]{luo-etal-2016-commonsense}
Zhiyi Luo, Yuchen Sha, Kenny~Q. Zhu, Seung{-}won Hwang, and Zhongyuan Wang. 2016.
\newblock \href {http://www.aaai.org/ocs/index.php/KR/KR16/paper/view/12818} {Commonsense causal reasoning between short texts}.
\newblock In \emph{Principles of Knowledge Representation and Reasoning: Proceedings of the Fifteenth International Conference, {KR} 2016, Cape Town, South Africa, April 25-29, 2016}, pages 421--431. {AAAI} Press.

\bibitem[{Madaan et~al.(2021)Madaan, Padhi, Panwar, and Saha}]{madaan-etal-2021-generate}
Nishtha Madaan, Inkit Padhi, Naveen Panwar, and Diptikalyan Saha. 2021.
\newblock \href {https://doi.org/10.1609/aaai.v35i15.17594} {Generate your counterfactuals: Towards controlled counterfactual generation for text}.
\newblock In \emph{Thirty-Fifth {AAAI} Conference on Artificial Intelligence, {AAAI} 2021, Thirty-Third Conference on Innovative Applications of Artificial Intelligence, {IAAI} 2021, The Eleventh Symposium on Educational Advances in Artificial Intelligence, {EAAI} 2021, Virtual Event, February 2-9, 2021}, pages 13516--13524. {AAAI} Press.

\bibitem[{Mann and Thompson(1988)}]{mann-thompson-1988-rhetorical}
William~C Mann and Sandra~A Thompson. 1988.
\newblock \href {https://www.degruyter.com/document/doi/10.1515/text.1.1988.8.3.243/pdf} {Rhetorical structure theory: Toward a functional theory of text organization}.
\newblock \emph{Text-interdisciplinary Journal for the Study of Discourse}, 8(3):243--281.

\bibitem[{Marcos(2021)}]{marcos-2021-study}
Henrique Marcos. 2021.
\newblock \href {https://papers.ssrn.com/sol3/papers.cfm?abstract_id=3918384} {A study on defeasibility and defeaters in international law: Process or procedure distinction against the non-discrimination rule}.
\newblock \emph{International Courts and the Guarantee of Social Rights}.

\bibitem[{Martin(2018)}]{martin-2018-time}
Fabienne Martin. 2018.
\newblock Time in probabilistic causation: Direct vs. indirect uses of lexical causative verbs.
\newblock In \emph{Proceedings of Sinn und Bedeutung}, volume~22, pages 107--124.

\bibitem[{Matute et~al.(2015)Matute, Blanco, Yarritu, D{\'i}az-Lago, Vadillo, and Barberia}]{matute-etal-2015-illusions}
Helena Matute, Fernando Blanco, Ion Yarritu, Marcos D{\'i}az-Lago, Miguel~A. Vadillo, and Itxaso Barberia. 2015.
\newblock \href {https://api.semanticscholar.org/CorpusID:17888149} {Illusions of causality: how they bias our everyday thinking and how they could be reduced}.
\newblock \emph{Frontiers in Psychology}, 6.

\bibitem[{Menzies(1989)}]{menzies-1989-probabilistic}
Peter Menzies. 1989.
\newblock \href {https://api.semanticscholar.org/CorpusID:120916896} {Probabilistic causation and causal processes: A critique of lewis}.
\newblock \emph{Philosophy of Science}, 56:642 -- 663.

\bibitem[{Menzies and Beebee(2001)}]{menzies-beebee-2001-counterfactual}
Peter Menzies and Helen Beebee. 2001.
\newblock \href {https://api.semanticscholar.org/CorpusID:15752795} {Counterfactual theories of causation}.
\newblock In \emph{The Stanford Encyclopedia of Philosophy}.

\bibitem[{Mesnard et~al.(2024)Mesnard, Hardin, Dadashi, Bhupatiraju, Pathak, Sifre, Rivi{\`{e}}re, Kale, Love, Tafti, Hussenot, Chowdhery, Roberts, Barua, Botev, Castro{-}Ros, Slone, H{\'{e}}liou, Tacchetti, Bulanova, Paterson, Tsai, Shahriari, Lan, Choquette{-}Choo, Crepy, Cer, Ippolito, Reid, Buchatskaya, Ni, Noland, Yan, Tucker, Muraru, Rozhdestvenskiy, Michalewski, Tenney, Grishchenko, Austin, Keeling, Labanowski, Lespiau, Stanway, Brennan, Chen, Ferret, Chiu, and et~al.}]{mesnard-etal-2024-gemma}
Thomas Mesnard, Cassidy Hardin, Robert Dadashi, Surya Bhupatiraju, Shreya Pathak, Laurent Sifre, Morgane Rivi{\`{e}}re, Mihir~Sanjay Kale, Juliette Love, Pouya Tafti, L{\'{e}}onard Hussenot, Aakanksha Chowdhery, Adam Roberts, Aditya Barua, Alex Botev, Alex Castro{-}Ros, Ambrose Slone, Am{\'{e}}lie H{\'{e}}liou, Andrea Tacchetti, Anna Bulanova, Antonia Paterson, Beth Tsai, Bobak Shahriari, Charline~Le Lan, Christopher~A. Choquette{-}Choo, Cl{\'{e}}ment Crepy, Daniel Cer, Daphne Ippolito, David Reid, Elena Buchatskaya, Eric Ni, Eric Noland, Geng Yan, George Tucker, George{-}Cristian Muraru, Grigory Rozhdestvenskiy, Henryk Michalewski, Ian Tenney, Ivan Grishchenko, Jacob Austin, James Keeling, Jane Labanowski, Jean{-}Baptiste Lespiau, Jeff Stanway, Jenny Brennan, Jeremy Chen, Johan Ferret, Justin Chiu, and et~al. 2024.
\newblock \href {https://doi.org/10.48550/ARXIV.2403.08295} {Gemma: Open models based on gemini research and technology}.
\newblock \emph{CoRR}, abs/2403.08295.

\bibitem[{Mih{\u{a}}il{\u{a}} and Ananiadou(2013)}]{mihaila-ananiadou-2013-causes}
Claudiu Mih{\u{a}}il{\u{a}} and Sophia Ananiadou. 2013.
\newblock \href {https://aclanthology.org/P13-3006} {What causes a causal relation? detecting causal triggers in biomedical scientific discourse}.
\newblock In \emph{51st Annual Meeting of the Association for Computational Linguistics Proceedings of the Student Research Workshop}, pages 38--45, Sofia, Bulgaria. Association for Computational Linguistics.

\bibitem[{Mihaila et~al.(2013)Mihaila, Ohta, Pyysalo, and Ananiadou}]{mihuailua-etal-2013-biocause}
Claudiu Mihaila, Tomoko Ohta, Sampo Pyysalo, and Sophia Ananiadou. 2013.
\newblock \href {https://doi.org/10.1186/1471-2105-14-2} {Biocause: Annotating and analysing causality in the biomedical domain}.
\newblock \emph{{BMC} Bioinform.}, 14:2.

\bibitem[{Mikolov et~al.(2013)Mikolov, Chen, Corrado, and Dean}]{mikolov-etal-2013-efficient}
Tom{\'{a}}s Mikolov, Kai Chen, Greg Corrado, and Jeffrey Dean. 2013.
\newblock \href {http://arxiv.org/abs/1301.3781} {Efficient estimation of word representations in vector space}.
\newblock In \emph{1st International Conference on Learning Representations, {ICLR} 2013, Scottsdale, Arizona, USA, May 2-4, 2013, Workshop Track Proceedings}.

\bibitem[{Mirza et~al.(2014)Mirza, Sprugnoli, Tonelli, and Speranza}]{mirza-etal-2014-annotating}
Paramita Mirza, Rachele Sprugnoli, Sara Tonelli, and Manuela Speranza. 2014.
\newblock \href {https://doi.org/10.3115/v1/W14-0702} {Annotating causality in the {T}emp{E}val-3 corpus}.
\newblock In \emph{Proceedings of the {EACL} 2014 Workshop on Computational Approaches to Causality in Language ({CA}to{CL})}, pages 10--19, Gothenburg, Sweden. Association for Computational Linguistics.

\bibitem[{Mirza and Tonelli(2016)}]{mirza-tonelli-2016-catena}
Paramita Mirza and Sara Tonelli. 2016.
\newblock \href {https://aclanthology.org/C16-1007} {{CATENA}: {CA}usal and {TE}mporal relation extraction from {NA}tural language texts}.
\newblock In \emph{Proceedings of {COLING} 2016, the 26th International Conference on Computational Linguistics: Technical Papers}, pages 64--75, Osaka, Japan. The COLING 2016 Organizing Committee.

\bibitem[{Mooij et~al.(2016)Mooij, Peters, Janzing, Zscheischler, and Sch{\"{o}}lkopf}]{mooij-etal-2016-distinguishing}
Joris~M. Mooij, Jonas Peters, Dominik Janzing, Jakob Zscheischler, and Bernhard Sch{\"{o}}lkopf. 2016.
\newblock \href {http://jmlr.org/papers/v17/14-518.html} {Distinguishing cause from effect using observational data: Methods and benchmarks}.
\newblock \emph{J. Mach. Learn. Res.}, 17:32:1--32:102.

\bibitem[{Mostafazadeh et~al.(2016{\natexlab{a}})Mostafazadeh, Chambers, He, Parikh, Batra, Vanderwende, Kohli, and Allen}]{mostafazadeh-etal-2016-corpus}
Nasrin Mostafazadeh, Nathanael Chambers, Xiaodong He, Devi Parikh, Dhruv Batra, Lucy Vanderwende, Pushmeet Kohli, and James Allen. 2016{\natexlab{a}}.
\newblock \href {https://doi.org/10.18653/v1/N16-1098} {A corpus and cloze evaluation for deeper understanding of commonsense stories}.
\newblock In \emph{Proceedings of the 2016 Conference of the North {A}merican Chapter of the Association for Computational Linguistics: Human Language Technologies}, pages 839--849, San Diego, California. Association for Computational Linguistics.

\bibitem[{Mostafazadeh et~al.(2016{\natexlab{b}})Mostafazadeh, Grealish, Chambers, Allen, and Vanderwende}]{mostafazadeh-etal-2016-caters}
Nasrin Mostafazadeh, Alyson Grealish, Nathanael Chambers, James Allen, and Lucy Vanderwende. 2016{\natexlab{b}}.
\newblock \href {https://doi.org/10.18653/v1/W16-1007} {{C}a{T}e{RS}: Causal and temporal relation scheme for semantic annotation of event structures}.
\newblock In \emph{Proceedings of the Fourth Workshop on Events}, pages 51--61, San Diego, California. Association for Computational Linguistics.

\bibitem[{Mostafazadeh et~al.(2020)Mostafazadeh, Kalyanpur, Moon, Buchanan, Berkowitz, Biran, and Chu-Carroll}]{mostafazadeh-etal-2020-glucose}
Nasrin Mostafazadeh, Aditya Kalyanpur, Lori Moon, David Buchanan, Lauren Berkowitz, Or~Biran, and Jennifer Chu-Carroll. 2020.
\newblock \href {https://doi.org/10.18653/v1/2020.emnlp-main.370} {{GLUCOSE}: {G}enera{L}ized and {CO}ntextualized story explanations}.
\newblock In \emph{Proceedings of the 2020 Conference on Empirical Methods in Natural Language Processing (EMNLP)}, pages 4569--4586, Online. Association for Computational Linguistics.

\bibitem[{Mulkar{-}Mehta et~al.(2011)Mulkar{-}Mehta, Welty, Hobbs, and Hovy}]{mulkar-etal-2011-using}
Rutu Mulkar{-}Mehta, Christopher~A. Welty, Jerry~R. Hobbs, and Eduard~H. Hovy. 2011.
\newblock \href {http://aaai.org/ocs/index.php/FLAIRS/FLAIRS11/paper/view/2511} {Using part-of relations for discovering causality}.
\newblock In \emph{Proceedings of the Twenty-Fourth International Florida Artificial Intelligence Research Society Conference, May 18-20, 2011, Palm Beach, Florida, {USA}}. {AAAI} Press.

\bibitem[{Nadathur and Lauer(2020)}]{nadathur-lauer-2020-causal}
Prerna Nadathur and Sven Lauer. 2020.
\newblock \href {https://api.semanticscholar.org/CorpusID:2431703} {Causal necessity, causal sufficiency, and the implications of causative verbs}.
\newblock \emph{Glossa}, 5:49.

\bibitem[{Ning et~al.(2018)Ning, Feng, Wu, and Roth}]{ning-etal-2018-joint}
Qiang Ning, Zhili Feng, Hao Wu, and Dan Roth. 2018.
\newblock \href {https://doi.org/10.18653/v1/P18-1212} {Joint reasoning for temporal and causal relations}.
\newblock In \emph{Proceedings of the 56th Annual Meeting of the Association for Computational Linguistics (Volume 1: Long Papers)}, pages 2278--2288, Melbourne, Australia. Association for Computational Linguistics.

\bibitem[{Oh et~al.(2013)Oh, Torisawa, Hashimoto, Sano, De~Saeger, and Ohtake}]{oh-etal-2013-question}
Jong-Hoon Oh, Kentaro Torisawa, Chikara Hashimoto, Motoki Sano, Stijn De~Saeger, and Kiyonori Ohtake. 2013.
\newblock \href {https://aclanthology.org/P13-1170} {Why-question answering using intra- and inter-sentential causal relations}.
\newblock In \emph{Proceedings of the 51st Annual Meeting of the Association for Computational Linguistics (Volume 1: Long Papers)}, pages 1733--1743, Sofia, Bulgaria. Association for Computational Linguistics.

\bibitem[{OpenAI et~al.(2023)OpenAI, :, Achiam, Adler, Agarwal, Ahmad, Akkaya, Aleman, Almeida, Altenschmidt, Altman, Anadkat, Avila, Babuschkin, Balaji, Balcom, Baltescu, Bao, Bavarian, Belgum, Bello, Berdine, Bernadett-Shapiro, Berner, Bogdonoff, Boiko, Boyd, Brakman, Brockman, Brooks, Brundage, Button, Cai, Campbell, Cann, Carey, Carlson, Carmichael, Chan, Chang, Chantzis, Chen, Chen, Chen, Chen, Chen, Chess, Cho, Chu, Chung, Cummings, Currier, Dai, Decareaux, Degry, Deutsch, Deville, Dhar, Dohan, Dowling, Dunning, Ecoffet, Eleti, Eloundou, Farhi, Fedus, Felix, Fishman, Forte, Fulford, Gao, Georges, Gibson, Goel, Gogineni, Goh, Gontijo-Lopes, Gordon, Grafstein, Gray, Greene, Gross, Gu, Guo, Hallacy, Han, Harris, He, Heaton, Heidecke, Hesse, Hickey, Hickey, Hoeschele, Houghton, Hsu, Hu, Hu, Huizinga, Jain, Jain, Jang, Jiang, Jiang, Jin, Jin, Jomoto, Jonn, Jun, Kaftan, Łukasz Kaiser, Kamali, Kanitscheider, Keskar, Khan, Kilpatrick, Kim, Kim, Kim, Kirchner, Kiros, Knight, Kokotajlo, Łukasz Kondraciuk,
  Kondrich, Konstantinidis, Kosic, Krueger, Kuo, Lampe, Lan, Lee, Leike, Leung, Levy, Li, Lim, Lin, Lin, Litwin, Lopez, Lowe, Lue, Makanju, Malfacini, Manning, Markov, Markovski, Martin, Mayer, Mayne, McGrew, McKinney, McLeavey, McMillan, McNeil, Medina, Mehta, Menick, Metz, Mishchenko, Mishkin, Monaco, Morikawa, Mossing, Mu, Murati, Murk, Mély, Nair, Nakano, Nayak, Neelakantan, Ngo, Noh, Ouyang, O'Keefe, Pachocki, Paino, Palermo, Pantuliano, Parascandolo, Parish, Parparita, Passos, Pavlov, Peng, Perelman, de~Avila Belbute~Peres, Petrov, de~Oliveira~Pinto, Michael, Pokorny, Pokrass, Pong, Powell, Power, Power, Proehl, Puri, Radford, Rae, Ramesh, Raymond, Real, Rimbach, Ross, Rotsted, Roussez, Ryder, Saltarelli, Sanders, Santurkar, Sastry, Schmidt, Schnurr, Schulman, Selsam, Sheppard, Sherbakov, Shieh, Shoker, Shyam, Sidor, Sigler, Simens, Sitkin, Slama, Sohl, Sokolowsky, Song, Staudacher, Such, Summers, Sutskever, Tang, Tezak, Thompson, Tillet, Tootoonchian, Tseng, Tuggle, Turley, Tworek, Uribe, Vallone,
  Vijayvergiya, Voss, Wainwright, Wang, Wang, Wang, Ward, Wei, Weinmann, Welihinda, Welinder, Weng, Weng, Wiethoff, Willner, Winter, Wolrich, Wong, Workman, Wu, Wu, Wu, Xiao, Xu, Yoo, Yu, Yuan, Zaremba, Zellers, Zhang, Zhang, Zhao, Zheng, Zhuang, Zhuk, and Zoph}]{openai-etal-2023-gpt4}
OpenAI, :, Josh Achiam, Steven Adler, Sandhini Agarwal, Lama Ahmad, Ilge Akkaya, Florencia~Leoni Aleman, Diogo Almeida, Janko Altenschmidt, Sam Altman, Shyamal Anadkat, Red Avila, Igor Babuschkin, Suchir Balaji, Valerie Balcom, Paul Baltescu, Haiming Bao, Mo~Bavarian, Jeff Belgum, Irwan Bello, Jake Berdine, Gabriel Bernadett-Shapiro, Christopher Berner, Lenny Bogdonoff, Oleg Boiko, Madelaine Boyd, Anna-Luisa Brakman, Greg Brockman, Tim Brooks, Miles Brundage, Kevin Button, Trevor Cai, Rosie Campbell, Andrew Cann, Brittany Carey, Chelsea Carlson, Rory Carmichael, Brooke Chan, Che Chang, Fotis Chantzis, Derek Chen, Sully Chen, Ruby Chen, Jason Chen, Mark Chen, Ben Chess, Chester Cho, Casey Chu, Hyung~Won Chung, Dave Cummings, Jeremiah Currier, Yunxing Dai, Cory Decareaux, Thomas Degry, Noah Deutsch, Damien Deville, Arka Dhar, David Dohan, Steve Dowling, Sheila Dunning, Adrien Ecoffet, Atty Eleti, Tyna Eloundou, David Farhi, Liam Fedus, Niko Felix, Simón~Posada Fishman, Juston Forte, Isabella Fulford, Leo Gao,
  Elie Georges, Christian Gibson, Vik Goel, Tarun Gogineni, Gabriel Goh, Rapha Gontijo-Lopes, Jonathan Gordon, Morgan Grafstein, Scott Gray, Ryan Greene, Joshua Gross, Shixiang~Shane Gu, Yufei Guo, Chris Hallacy, Jesse Han, Jeff Harris, Yuchen He, Mike Heaton, Johannes Heidecke, Chris Hesse, Alan Hickey, Wade Hickey, Peter Hoeschele, Brandon Houghton, Kenny Hsu, Shengli Hu, Xin Hu, Joost Huizinga, Shantanu Jain, Shawn Jain, Joanne Jang, Angela Jiang, Roger Jiang, Haozhun Jin, Denny Jin, Shino Jomoto, Billie Jonn, Heewoo Jun, Tomer Kaftan, Łukasz Kaiser, Ali Kamali, Ingmar Kanitscheider, Nitish~Shirish Keskar, Tabarak Khan, Logan Kilpatrick, Jong~Wook Kim, Christina Kim, Yongjik Kim, Hendrik Kirchner, Jamie Kiros, Matt Knight, Daniel Kokotajlo, Łukasz Kondraciuk, Andrew Kondrich, Aris Konstantinidis, Kyle Kosic, Gretchen Krueger, Vishal Kuo, Michael Lampe, Ikai Lan, Teddy Lee, Jan Leike, Jade Leung, Daniel Levy, Chak~Ming Li, Rachel Lim, Molly Lin, Stephanie Lin, Mateusz Litwin, Theresa Lopez, Ryan Lowe,
  Patricia Lue, Anna Makanju, Kim Malfacini, Sam Manning, Todor Markov, Yaniv Markovski, Bianca Martin, Katie Mayer, Andrew Mayne, Bob McGrew, Scott~Mayer McKinney, Christine McLeavey, Paul McMillan, Jake McNeil, David Medina, Aalok Mehta, Jacob Menick, Luke Metz, Andrey Mishchenko, Pamela Mishkin, Vinnie Monaco, Evan Morikawa, Daniel Mossing, Tong Mu, Mira Murati, Oleg Murk, David Mély, Ashvin Nair, Reiichiro Nakano, Rajeev Nayak, Arvind Neelakantan, Richard Ngo, Hyeonwoo Noh, Long Ouyang, Cullen O'Keefe, Jakub Pachocki, Alex Paino, Joe Palermo, Ashley Pantuliano, Giambattista Parascandolo, Joel Parish, Emy Parparita, Alex Passos, Mikhail Pavlov, Andrew Peng, Adam Perelman, Filipe de~Avila Belbute~Peres, Michael Petrov, Henrique~Ponde de~Oliveira~Pinto, Michael, Pokorny, Michelle Pokrass, Vitchyr Pong, Tolly Powell, Alethea Power, Boris Power, Elizabeth Proehl, Raul Puri, Alec Radford, Jack Rae, Aditya Ramesh, Cameron Raymond, Francis Real, Kendra Rimbach, Carl Ross, Bob Rotsted, Henri Roussez, Nick Ryder,
  Mario Saltarelli, Ted Sanders, Shibani Santurkar, Girish Sastry, Heather Schmidt, David Schnurr, John Schulman, Daniel Selsam, Kyla Sheppard, Toki Sherbakov, Jessica Shieh, Sarah Shoker, Pranav Shyam, Szymon Sidor, Eric Sigler, Maddie Simens, Jordan Sitkin, Katarina Slama, Ian Sohl, Benjamin Sokolowsky, Yang Song, Natalie Staudacher, Felipe~Petroski Such, Natalie Summers, Ilya Sutskever, Jie Tang, Nikolas Tezak, Madeleine Thompson, Phil Tillet, Amin Tootoonchian, Elizabeth Tseng, Preston Tuggle, Nick Turley, Jerry Tworek, Juan Felipe~Cerón Uribe, Andrea Vallone, Arun Vijayvergiya, Chelsea Voss, Carroll Wainwright, Justin~Jay Wang, Alvin Wang, Ben Wang, Jonathan Ward, Jason Wei, CJ~Weinmann, Akila Welihinda, Peter Welinder, Jiayi Weng, Lilian Weng, Matt Wiethoff, Dave Willner, Clemens Winter, Samuel Wolrich, Hannah Wong, Lauren Workman, Sherwin Wu, Jeff Wu, Michael Wu, Kai Xiao, Tao Xu, Sarah Yoo, Kevin Yu, Qiming Yuan, Wojciech Zaremba, Rowan Zellers, Chong Zhang, Marvin Zhang, Shengjia Zhao, Tianhao
  Zheng, Juntang Zhuang, William Zhuk, and Barret Zoph. 2023.
\newblock \href {http://arxiv.org/abs/2303.08774} {Gpt-4 technical report}.

\bibitem[{Palmer et~al.(2005)Palmer, Gildea, and Kingsbury}]{palmer-etal-2005-proposition}
Martha Palmer, Daniel Gildea, and Paul Kingsbury. 2005.
\newblock \href {https://doi.org/10.1162/0891201053630264} {The {P}roposition {B}ank: An annotated corpus of semantic roles}.
\newblock \emph{Computational Linguistics}, 31(1):71--106.

\bibitem[{Pearl(2000)}]{pearl-2000-causality}
Judea Pearl. 2000.
\newblock \href {https://api.semanticscholar.org/CorpusID:12575481} {Causality: Models, reasoning and inference}.

\bibitem[{Pearl(2009)}]{pearl-2009-causality}
Judea Pearl. 2009.
\newblock \href {http://bayes.cs.ucla.edu/BOOK-2K/} {\emph{Causality}}.
\newblock Cambridge university press.

\bibitem[{Pearl(2012)}]{pearl-2012-do-calculus}
Judea Pearl. 2012.
\newblock \href {https://dslpitt.org/uai/displayArticleDetails.jsp?mmnu=1\&smnu=2\&article\_id=2330\&proceeding\_id=28} {The do-calculus revisited}.
\newblock In \emph{Proceedings of the Twenty-Eighth Conference on Uncertainty in Artificial Intelligence, Catalina Island, CA, USA, August 14-18, 2012}, pages 3--11. {AUAI} Press.

\bibitem[{Pearl et~al.(2016)Pearl, Glymour, and Jewell}]{pearl-etal-2016-causal}
Judea Pearl, Madelyn Glymour, and Nicholas~P Jewell. 2016.
\newblock \href {http://bayes.cs.ucla.edu/PRIMER/} {\emph{Causal inference in statistics: A primer}}.
\newblock John Wiley \& Sons.

\bibitem[{Pearl and Mackenzie(2018)}]{pearl-mackenzie-2018-book}
Judea Pearl and Dana Mackenzie. 2018.
\newblock \href {https://dl.acm.org/doi/10.5555/3238230} {\emph{The book of why: the new science of cause and effect}}.
\newblock Basic books.

\bibitem[{Pennington et~al.(2014)Pennington, Socher, and Manning}]{pennington-etal-2014-glove}
Jeffrey Pennington, Richard Socher, and Christopher Manning. 2014.
\newblock \href {https://doi.org/10.3115/v1/D14-1162} {{G}lo{V}e: Global vectors for word representation}.
\newblock In \emph{Proceedings of the 2014 Conference on Empirical Methods in Natural Language Processing ({EMNLP})}, pages 1532--1543, Doha, Qatar. Association for Computational Linguistics.

\bibitem[{Pesaran and Smith(2016)}]{pesaran-smith-2016-counterfactual}
M~Hashem Pesaran and Ron~P Smith. 2016.
\newblock \href {https://docs.iza.org/dp6618.pdf} {Counterfactual analysis in macroeconometrics: An empirical investigation into the effects of quantitative easing}.
\newblock \emph{Research in Economics}, 70(2):262--280.

\bibitem[{Peters et~al.(2017)Peters, Janzing, and Sch{\"o}lkopf}]{peters-etal-2017-elements}
Jonas Peters, Dominik Janzing, and Bernhard Sch{\"o}lkopf. 2017.
\newblock \href {https://mitpress.mit.edu/9780262037310/elements-of-causal-inference/} {\emph{Elements of causal inference: foundations and learning algorithms}}.
\newblock The MIT Press.

\bibitem[{Ponti et~al.(2020)Ponti, Glava{\v{s}}, Majewska, Liu, Vuli{\'c}, and Korhonen}]{ponti-etal-2020-xcopa}
Edoardo~Maria Ponti, Goran Glava{\v{s}}, Olga Majewska, Qianchu Liu, Ivan Vuli{\'c}, and Anna Korhonen. 2020.
\newblock \href {https://doi.org/10.18653/v1/2020.emnlp-main.185} {{XCOPA}: A multilingual dataset for causal commonsense reasoning}.
\newblock In \emph{Proceedings of the 2020 Conference on Empirical Methods in Natural Language Processing (EMNLP)}, pages 2362--2376, Online. Association for Computational Linguistics.

\bibitem[{Prasad et~al.(2008)Prasad, Dinesh, Lee, Miltsakaki, Robaldo, Joshi, and Webber}]{prasad-etal-2008-penn}
Rashmi Prasad, Nikhil Dinesh, Alan Lee, Eleni Miltsakaki, Livio Robaldo, Aravind Joshi, and Bonnie Webber. 2008.
\newblock \href {http://www.lrec-conf.org/proceedings/lrec2008/pdf/754_paper.pdf} {The {P}enn {D}iscourse {T}ree{B}ank 2.0.}
\newblock In \emph{Proceedings of the Sixth International Conference on Language Resources and Evaluation ({LREC}'08)}, Marrakech, Morocco. European Language Resources Association (ELRA).

\bibitem[{Pyysalo et~al.(2007)Pyysalo, Ginter, Heimonen, Bj{\"{o}}rne, Boberg, J{\"{a}}rvinen, and Salakoski}]{pyysalo-etal-2007-bioinfer}
Sampo Pyysalo, Filip Ginter, Juho Heimonen, Jari Bj{\"{o}}rne, Jorma Boberg, Jouni J{\"{a}}rvinen, and Tapio Salakoski. 2007.
\newblock \href {https://doi.org/10.1186/1471-2105-8-50} {Bioinfer: a corpus for information extraction in the biomedical domain}.
\newblock \emph{{BMC} Bioinform.}, 8.

\bibitem[{Qiao et~al.(2023)Qiao, Ou, Zhang, Chen, Yao, Deng, Tan, Huang, and Chen}]{qiao-etal-2023-reasoning}
Shuofei Qiao, Yixin Ou, Ningyu Zhang, Xiang Chen, Yunzhi Yao, Shumin Deng, Chuanqi Tan, Fei Huang, and Huajun Chen. 2023.
\newblock \href {https://doi.org/10.18653/v1/2023.acl-long.294} {Reasoning with language model prompting: A survey}.
\newblock In \emph{Proceedings of the 61st Annual Meeting of the Association for Computational Linguistics (Volume 1: Long Papers)}, pages 5368--5393, Toronto, Canada. Association for Computational Linguistics.

\bibitem[{Qin et~al.(2019)Qin, Bosselut, Holtzman, Bhagavatula, Clark, and Choi}]{qin-etal-2019-counterfactual}
Lianhui Qin, Antoine Bosselut, Ari Holtzman, Chandra Bhagavatula, Elizabeth Clark, and Yejin Choi. 2019.
\newblock \href {https://doi.org/10.18653/v1/D19-1509} {Counterfactual story reasoning and generation}.
\newblock In \emph{Proceedings of the 2019 Conference on Empirical Methods in Natural Language Processing and the 9th International Joint Conference on Natural Language Processing (EMNLP-IJCNLP)}, pages 5043--5053, Hong Kong, China. Association for Computational Linguistics.

\bibitem[{Quinlan(1986)}]{quinlan-1986-induction}
J.~Ross Quinlan. 1986.
\newblock \href {https://doi.org/10.1023/A:1022643204877} {Induction of decision trees}.
\newblock \emph{Mach. Learn.}, 1(1):81--106.

\bibitem[{Radinsky et~al.(2012)Radinsky, Davidovich, and Markovitch}]{radinsky-etal-2012-learning}
Kira Radinsky, Sagie Davidovich, and Shaul Markovitch. 2012.
\newblock \href {https://doi.org/10.1145/2187836.2187958} {Learning causality for news events prediction}.
\newblock In \emph{Proceedings of the 21st World Wide Web Conference 2012, {WWW} 2012, Lyon, France, April 16-20, 2012}, pages 909--918. {ACM}.

\bibitem[{Raffel et~al.(2020)Raffel, Shazeer, Roberts, Lee, Narang, Matena, Zhou, Li, and Liu}]{raffel-etal-2020-exploring}
Colin Raffel, Noam Shazeer, Adam Roberts, Katherine Lee, Sharan Narang, Michael Matena, Yanqi Zhou, Wei Li, and Peter~J. Liu. 2020.
\newblock \href {http://jmlr.org/papers/v21/20-074.html} {Exploring the limits of transfer learning with a unified text-to-text transformer}.
\newblock \emph{J. Mach. Learn. Res.}, 21:140:1--140:67.

\bibitem[{Rashkin et~al.(2018)Rashkin, Sap, Allaway, Smith, and Choi}]{rashkin-etal-2018-event2mind}
Hannah Rashkin, Maarten Sap, Emily Allaway, Noah~A. Smith, and Yejin Choi. 2018.
\newblock \href {https://doi.org/10.18653/v1/P18-1043} {{E}vent2{M}ind: Commonsense inference on events, intents, and reactions}.
\newblock In \emph{Proceedings of the 56th Annual Meeting of the Association for Computational Linguistics (Volume 1: Long Papers)}, pages 463--473, Melbourne, Australia. Association for Computational Linguistics.

\bibitem[{Reichenbach(1956)}]{reichenbach-1956-direction}
Hans Reichenbach. 1956.
\newblock \href {https://philpapers.org/rec/REITDO-2} {\emph{The Direction of Time}}.
\newblock Dover Publications, Mineola, N.Y.

\bibitem[{Richens et~al.(2020)Richens, Lee, and Johri}]{richens-etal-2020-improving}
Jonathan~G. Richens, Ciar{\'a}n~M. Lee, and Saurabh Johri. 2020.
\newblock \href {https://api.semanticscholar.org/CorpusID:221108473} {Improving the accuracy of medical diagnosis with causal machine learning}.
\newblock \emph{Nature Communications}, 11.

\bibitem[{Robeer et~al.(2021)Robeer, Bex, and Feelders}]{robeer-etal-2021-generating-realistic}
Marcel Robeer, Floris Bex, and Ad~Feelders. 2021.
\newblock \href {https://doi.org/10.18653/v1/2021.findings-emnlp.306} {Generating realistic natural language counterfactuals}.
\newblock In \emph{Findings of the Association for Computational Linguistics: EMNLP 2021}, pages 3611--3625, Punta Cana, Dominican Republic. Association for Computational Linguistics.

\bibitem[{Roemmele et~al.(2011)Roemmele, Bejan, and Gordon}]{roemmele-etal-2011-choice}
Melissa Roemmele, Cosmin~Adrian Bejan, and Andrew~S. Gordon. 2011.
\newblock \href {http://www.aaai.org/ocs/index.php/SSS/SSS11/paper/view/2418} {Choice of plausible alternatives: An evaluation of commonsense causal reasoning}.
\newblock In \emph{Logical Formalizations of Commonsense Reasoning, Papers from the 2011 {AAAI} Spring Symposium, Technical Report SS-11-06, Stanford, California, USA, March 21-23, 2011}. {AAAI}.

\bibitem[{Roese(1994)}]{roese-1994-functional}
Neal~J Roese. 1994.
\newblock The functional basis of counterfactual thinking.
\newblock \emph{Journal of personality and Social Psychology}, 66(5):805.

\bibitem[{Roese and Morrison(2009)}]{roese-morrison-2009-psychology}
Neal~J Roese and Mike Morrison. 2009.
\newblock \href {https://www.jstor.org/stable/20762352} {The psychology of counterfactual thinking}.
\newblock \emph{Historical Social Research/Historische Sozialforschung}, pages 16--26.

\bibitem[{Rohrer(2018)}]{rohrer-2018-thinking}
Julia~M Rohrer. 2018.
\newblock \href {https://doi.org/10.1177/2515245917745629} {Thinking clearly about correlations and causation: Graphical causal models for observational data}.
\newblock \emph{Advances in methods and practices in psychological science}, 1(1):27--42.

\bibitem[{Rubin(1974)}]{rubin-1974-estimating}
Donald~B Rubin. 1974.
\newblock \href {https://onlinelibrary.wiley.com/doi/abs/10.1002/j.2333-8504.1972.tb00631.x} {Estimating causal effects of treatments in randomized and nonrandomized studies.}
\newblock \emph{Journal of educational Psychology}, 66(5):688.

\bibitem[{Ruppenhofer et~al.(2016)Ruppenhofer, Ellsworth, Schwarzer-Petruck, Johnson, and Scheffczyk}]{ruppenhofer-etal-2016-framenet}
Josef Ruppenhofer, Michael Ellsworth, Myriam Schwarzer-Petruck, Christopher~R Johnson, and Jan Scheffczyk. 2016.
\newblock \href {https://framenet2.icsi.berkeley.edu/docs/r1.7/book.pdf} {Framenet ii: Extended theory and practice}.
\newblock Technical report, International Computer Science Institute.

\bibitem[{Russo and Williamson(2007)}]{russo-williamson-2007-interpreting}
Federica Russo and Jon Williamson. 2007.
\newblock \href {https://doi.org/10.1080/02698590701498084} {Interpreting causality in the health sciences}.
\newblock \emph{International Studies in the Philosophy of Science}, 21(2):157--170.

\bibitem[{Sakaji et~al.(2008)Sakaji, Sekine, and Masuyama}]{sakaji-etal-2008-extracting}
Hiroki Sakaji, Satoshi Sekine, and Shigeru Masuyama. 2008.
\newblock \href {https://doi.org/10.1007/978-3-540-89447-6\_12} {Extracting causal knowledge using clue phrases and syntactic patterns}.
\newblock In \emph{Practical Aspects of Knowledge Management, 7th International Conference, {PAKM} 2008, Yokohama, Japan, November 22-23, 2008. Proceedings}, volume 5345 of \emph{Lecture Notes in Computer Science}, pages 111--122. Springer.

\bibitem[{Saki and Faghihi(2022)}]{saki-faghihi-2022-fundamental}
Amir Saki and Usef Faghihi. 2022.
\newblock \href {https://doi.org/10.48550/ARXIV.2205.15016} {A fundamental probabilistic fuzzy logic framework suitable for causal reasoning}.
\newblock \emph{CoRR}, abs/2205.15016.

\bibitem[{Sanh et~al.(2019)Sanh, Debut, Chaumond, and Wolf}]{sanh-etal-2019-distilbert}
Victor Sanh, Lysandre Debut, Julien Chaumond, and Thomas Wolf. 2019.
\newblock \href {http://arxiv.org/abs/1910.01108} {Distilbert, a distilled version of {BERT:} smaller, faster, cheaper and lighter}.
\newblock \emph{CoRR}, abs/1910.01108.

\bibitem[{Sap et~al.(2019{\natexlab{a}})Sap, Bras, Allaway, Bhagavatula, Lourie, Rashkin, Roof, Smith, and Choi}]{sap-etal-2019-atomic}
Maarten Sap, Ronan~Le Bras, Emily Allaway, Chandra Bhagavatula, Nicholas Lourie, Hannah Rashkin, Brendan Roof, Noah~A. Smith, and Yejin Choi. 2019{\natexlab{a}}.
\newblock \href {https://doi.org/10.1609/aaai.v33i01.33013027} {{ATOMIC:} an atlas of machine commonsense for if-then reasoning}.
\newblock In \emph{The Thirty-Third {AAAI} Conference on Artificial Intelligence, {AAAI} 2019, The Thirty-First Innovative Applications of Artificial Intelligence Conference, {IAAI} 2019, The Ninth {AAAI} Symposium on Educational Advances in Artificial Intelligence, {EAAI} 2019, Honolulu, Hawaii, USA, January 27 - February 1, 2019}, pages 3027--3035. {AAAI} Press.

\bibitem[{Sap et~al.(2019{\natexlab{b}})Sap, Rashkin, Chen, Le~Bras, and Choi}]{sap-etal-2019-social}
Maarten Sap, Hannah Rashkin, Derek Chen, Ronan Le~Bras, and Yejin Choi. 2019{\natexlab{b}}.
\newblock \href {https://doi.org/10.18653/v1/D19-1454} {Social {IQ}a: Commonsense reasoning about social interactions}.
\newblock In \emph{Proceedings of the 2019 Conference on Empirical Methods in Natural Language Processing and the 9th International Joint Conference on Natural Language Processing (EMNLP-IJCNLP)}, pages 4463--4473, Hong Kong, China. Association for Computational Linguistics.

\bibitem[{Shalit et~al.(2017)Shalit, Johansson, and Sontag}]{shalit-etal-2017-estimating}
Uri Shalit, Fredrik~D. Johansson, and David~A. Sontag. 2017.
\newblock \href {http://proceedings.mlr.press/v70/shalit17a.html} {Estimating individual treatment effect: generalization bounds and algorithms}.
\newblock In \emph{Proceedings of the 34th International Conference on Machine Learning, {ICML} 2017, Sydney, NSW, Australia, 6-11 August 2017}, volume~70 of \emph{Proceedings of Machine Learning Research}, pages 3076--3085. {PMLR}.

\bibitem[{Skyrms(1981)}]{skyrms-1981-causal}
Brian Skyrms. 1981.
\newblock \href {https://doi.org/10.1086/289003} {Causal necessity}.
\newblock \emph{Philosophy of Science}, 48(2):329--335.

\bibitem[{Son et~al.(2017)Son, Buffone, Raso, Larche, Janocko, Zembroski, Schwartz, and Ungar}]{son-etal-2017-recognizing}
Youngseo Son, Anneke Buffone, Joe Raso, Allegra Larche, Anthony Janocko, Kevin Zembroski, H~Andrew Schwartz, and Lyle Ungar. 2017.
\newblock \href {https://doi.org/10.18653/v1/P17-2103} {Recognizing counterfactual thinking in social media texts}.
\newblock In \emph{Proceedings of the 55th Annual Meeting of the Association for Computational Linguistics (Volume 2: Short Papers)}, pages 654--658, Vancouver, Canada. Association for Computational Linguistics.

\bibitem[{Speer et~al.(2017)Speer, Chin, and Havasi}]{speer-etal-2017-conceptnet}
Robyn Speer, Joshua Chin, and Catherine Havasi. 2017.
\newblock \href {https://doi.org/10.1609/aaai.v31i1.11164} {Conceptnet 5.5: An open multilingual graph of general knowledge}.
\newblock In \emph{Proceedings of the Thirty-First {AAAI} Conference on Artificial Intelligence, February 4-9, 2017, San Francisco, California, {USA}}, pages 4444--4451. {AAAI} Press.

\bibitem[{Storks et~al.(2019)Storks, Gao, and Chai}]{storks-etal-2019-commonsense}
Shane Storks, Qiaozi Gao, and Joyce~Y. Chai. 2019.
\newblock \href {http://arxiv.org/abs/1904.01172} {Commonsense reasoning for natural language understanding: {A} survey of benchmarks, resources, and approaches}.
\newblock \emph{CoRR}, abs/1904.01172.

\bibitem[{Summers(2018)}]{summers-2018-common}
Andrew Summers. 2018.
\newblock \href {https://doi.org/10.1093/ojls/gqy028} {{Common-Sense Causation in the Law}}.
\newblock \emph{Oxford Journal of Legal Studies}, 38(4):793--821.

\bibitem[{Sun et~al.(2019)Sun, Wang, Li, Feng, Chen, Zhang, Tian, Zhu, Tian, and Wu}]{sun-etal-2019-ernie}
Yu~Sun, Shuohuan Wang, Yu{-}Kun Li, Shikun Feng, Xuyi Chen, Han Zhang, Xin Tian, Danxiang Zhu, Hao Tian, and Hua Wu. 2019.
\newblock \href {http://arxiv.org/abs/1904.09223} {{ERNIE:} enhanced representation through knowledge integration}.
\newblock \emph{CoRR}, abs/1904.09223.

\bibitem[{Suppes(1973)}]{suppes-1973-probabilistic}
Patrick Suppes. 1973.
\newblock \href {https://philpapers.org/rec/SUPAPT} {A probabilistic theory of causality}.
\newblock \emph{British Journal for the Philosophy of Science}, 24(4).

\bibitem[{Sutskever et~al.(2014)Sutskever, Vinyals, and Le}]{sutskever-etal-2014-sequence}
Ilya Sutskever, Oriol Vinyals, and Quoc~V. Le. 2014.
\newblock \href {https://proceedings.neurips.cc/paper/2014/hash/a14ac55a4f27472c5d894ec1c3c743d2-Abstract.html} {Sequence to sequence learning with neural networks}.
\newblock In \emph{Advances in Neural Information Processing Systems 27: Annual Conference on Neural Information Processing Systems 2014, December 8-13 2014, Montreal, Quebec, Canada}, pages 3104--3112.

\bibitem[{Sutton and McCallum(2012)}]{sutton-etal-2012-introduction}
Charles Sutton and Andrew McCallum. 2012.
\newblock \href {https://doi.org/10.1561/2200000013} {An introduction to conditional random fields}.
\newblock \emph{Found. Trends Mach. Learn.}, 4(4):267--373.

\bibitem[{Swampillai and Stevenson(2011)}]{swampillai-stevenson-2011-extracting}
Kumutha Swampillai and Mark Stevenson. 2011.
\newblock \href {https://aclanthology.org/R11-1004} {Extracting relations within and across sentences}.
\newblock In \emph{Proceedings of the International Conference Recent Advances in Natural Language Processing 2011}, pages 25--32, Hissar, Bulgaria. Association for Computational Linguistics.

\bibitem[{Talmy(1988)}]{talmy-1988-force}
Leonard Talmy. 1988.
\newblock \href {https://doi.org/10.1207/S15516709COG1201\_2} {Force dynamics in language and cognition}.
\newblock \emph{Cogn. Sci.}, 12(1):49--100.

\bibitem[{Tetlock and Belkin(1996)}]{tetlock-belkin-1996-counterfactual}
Philip~E Tetlock and Aaron Belkin. 1996.
\newblock \href {https://www.jstor.org/stable/j.ctv10vm1bn} {\emph{Counterfactual thought experiments in world politics: Logical, methodological, and psychological perspectives}}.
\newblock Princeton University Press.

\bibitem[{Thoppilan et~al.(2022)Thoppilan, Freitas, Hall, Shazeer, Kulshreshtha, Cheng, Jin, Bos, Baker, Du, Li, Lee, Zheng, Ghafouri, Menegali, Huang, Krikun, Lepikhin, Qin, Chen, Xu, Chen, Roberts, Bosma, Zhou, Chang, Krivokon, Rusch, Pickett, Meier{-}Hellstern, Morris, Doshi, Santos, Duke, Soraker, Zevenbergen, Prabhakaran, Diaz, Hutchinson, Olson, Molina, Hoffman{-}John, Lee, Aroyo, Rajakumar, Butryna, Lamm, Kuzmina, Fenton, Cohen, Bernstein, Kurzweil, y~Arcas, Cui, Croak, Chi, and Le}]{thoppilan-etal-2022-lamda}
Romal Thoppilan, Daniel~De Freitas, Jamie Hall, Noam Shazeer, Apoorv Kulshreshtha, Heng{-}Tze Cheng, Alicia Jin, Taylor Bos, Leslie Baker, Yu~Du, YaGuang Li, Hongrae Lee, Huaixiu~Steven Zheng, Amin Ghafouri, Marcelo Menegali, Yanping Huang, Maxim Krikun, Dmitry Lepikhin, James Qin, Dehao Chen, Yuanzhong Xu, Zhifeng Chen, Adam Roberts, Maarten Bosma, Yanqi Zhou, Chung{-}Ching Chang, Igor Krivokon, Will Rusch, Marc Pickett, Kathleen~S. Meier{-}Hellstern, Meredith~Ringel Morris, Tulsee Doshi, Renelito~Delos Santos, Toju Duke, Johnny Soraker, Ben Zevenbergen, Vinodkumar Prabhakaran, Mark Diaz, Ben Hutchinson, Kristen Olson, Alejandra Molina, Erin Hoffman{-}John, Josh Lee, Lora Aroyo, Ravi Rajakumar, Alena Butryna, Matthew Lamm, Viktoriya Kuzmina, Joe Fenton, Aaron Cohen, Rachel Bernstein, Ray Kurzweil, Blaise~Ag{\"{u}}era y~Arcas, Claire Cui, Marian Croak, Ed~H. Chi, and Quoc Le. 2022.
\newblock \href {http://arxiv.org/abs/2201.08239} {Lamda: Language models for dialog applications}.
\newblock \emph{CoRR}, abs/2201.08239.

\bibitem[{Touvron et~al.(2023)Touvron, Lavril, Izacard, Martinet, Lachaux, Lacroix, Rozi{\`{e}}re, Goyal, Hambro, Azhar, Rodriguez, Joulin, Grave, and Lample}]{touvron-etal-2023-llama}
Hugo Touvron, Thibaut Lavril, Gautier Izacard, Xavier Martinet, Marie{-}Anne Lachaux, Timoth{\'{e}}e Lacroix, Baptiste Rozi{\`{e}}re, Naman Goyal, Eric Hambro, Faisal Azhar, Aur{\'{e}}lien Rodriguez, Armand Joulin, Edouard Grave, and Guillaume Lample. 2023.
\newblock \href {https://doi.org/10.48550/ARXIV.2302.13971} {Llama: Open and efficient foundation language models}.
\newblock \emph{CoRR}, abs/2302.13971.

\bibitem[{Tucci(2013)}]{tucci-2013-introduction}
Robert~R. Tucci. 2013.
\newblock \href {http://arxiv.org/abs/1305.5506} {Introduction to judea pearl's do-calculus}.
\newblock \emph{CoRR}, abs/1305.5506.

\bibitem[{Vaswani et~al.(2017)Vaswani, Shazeer, Parmar, Uszkoreit, Jones, Gomez, Kaiser, and Polosukhin}]{vaswani-etal-2017-attention}
Ashish Vaswani, Noam Shazeer, Niki Parmar, Jakob Uszkoreit, Llion Jones, Aidan~N. Gomez, Lukasz Kaiser, and Illia Polosukhin. 2017.
\newblock \href {https://proceedings.neurips.cc/paper/2017/hash/3f5ee243547dee91fbd053c1c4a845aa-Abstract.html} {Attention is all you need}.
\newblock In \emph{Advances in Neural Information Processing Systems 30: Annual Conference on Neural Information Processing Systems 2017, December 4-9, 2017, Long Beach, CA, {USA}}, pages 5998--6008.

\bibitem[{Venzke(2018)}]{venzke-2018-if}
Ingo Venzke. 2018.
\newblock \href {https://papers.ssrn.com/sol3/papers.cfm?abstract_id=2881226} {What if? counterfactual (hi) stories of international law}.
\newblock \emph{Asian journal of international law}, 8(2):403--431.

\bibitem[{Vigen(2015)}]{vigen-2015-spurious}
T.~Vigen. 2015.
\newblock \href {https://books.google.ch/books?id=0uDrBQAAQBAJ} {\emph{Spurious Correlations}}.
\newblock Hachette Books.

\bibitem[{Volchok(2015)}]{volchok-2015-three}
Edward Volchok. 2015.
\newblock {T}hree {L}evels of {C}ausation --- media.acc.qcc.cuny.edu.
\newblock \url{http://media.acc.qcc.cuny.edu/faculty/volchok/causalMR/CausalMR3.html}.
\newblock [Accessed 26-06-2024].

\bibitem[{Wang et~al.(2023)Wang, Do, Zhang, Zhang, Wang, Fang, Song, Wong, and See}]{wang-etal-2023-cola}
Zhaowei Wang, Quyet~V. Do, Hongming Zhang, Jiayao Zhang, Weiqi Wang, Tianqing Fang, Yangqiu Song, Ginny Wong, and Simon See. 2023.
\newblock \href {https://doi.org/10.18653/v1/2023.acl-long.288} {{COLA}: Contextualized commonsense causal reasoning from the causal inference perspective}.
\newblock In \emph{Proceedings of the 61st Annual Meeting of the Association for Computational Linguistics (Volume 1: Long Papers)}, pages 5253--5271, Toronto, Canada. Association for Computational Linguistics.

\bibitem[{Webber et~al.(2019)Webber, Prasad, Lee, and Joshi}]{webber-etal-2019-penn}
Bonnie Webber, Rashmi Prasad, Alan Lee, and Aravind Joshi. 2019.
\newblock \href {https://catalog.ldc.upenn.edu/docs/LDC2019T05/PDTB3-Annotation-Manual.pdf} {The penn discourse treebank 3.0 annotation manual}.
\newblock \emph{Philadelphia, University of Pennsylvania}, 35:108.

\bibitem[{Wei et~al.(2022)Wei, Wang, Schuurmans, Bosma, Ichter, Xia, Chi, Le, and Zhou}]{wei-etal-2022-chain}
Jason Wei, Xuezhi Wang, Dale Schuurmans, Maarten Bosma, Brian Ichter, Fei Xia, Ed~H. Chi, Quoc~V. Le, and Denny Zhou. 2022.
\newblock \href {http://papers.nips.cc/paper\_files/paper/2022/hash/9d5609613524ecf4f15af0f7b31abca4-Abstract-Conference.html} {Chain-of-thought prompting elicits reasoning in large language models}.
\newblock In \emph{NeurIPS}.

\bibitem[{Weinrib(2016)}]{weinrib-2016-causal}
Ernest~J Weinrib. 2016.
\newblock \href {https://www.jstor.org/stable/26363959} {Causal uncertainty}.
\newblock \emph{Oxford Journal of Legal Studies}, 36(1):135--164.

\bibitem[{Williams(1961)}]{williams-1961-causation}
Glanville~Llewelyn Williams. 1961.
\newblock \href {https://api.semanticscholar.org/CorpusID:144104729} {Causation in the law}.
\newblock \emph{The Cambridge Law Journal}, 19:62 -- 85.

\bibitem[{Williamson(2009)}]{williamson-2009-probabilistic}
Jon Williamson. 2009.
\newblock \href {https://api.semanticscholar.org/CorpusID:2409003} {Probabilistic theories of causality}.

\bibitem[{Wolff(2007)}]{wolff-2007-representing}
Phillip Wolff. 2007.
\newblock \href {https://doi.org/10.1037/0096-3445.136.1.82} {Representing causation}.
\newblock \emph{Journal of experimental psychology: General}, 136(1):82.

\bibitem[{Wolff and Shepard(2013)}]{wollf-shepard-2013-chapter}
Phillip Wolff and Jason Shepard. 2013.
\newblock \href {https://doi.org/https://doi.org/10.1016/B978-0-12-407237-4.00005-0} {Causation, touch, and the perception of force}.
\newblock In Brian~H. Ross, editor, \emph{Psychology of learning and motivation}, volume~58 of \emph{Psychology of Learning and Motivation}, pages 167--202. Academic Press.

\bibitem[{Wolff and Thorstad(2017)}]{wolff-thorstad-2017-force}
Phillip Wolff and Robert Thorstad. 2017.
\newblock Force dynamics.
\newblock \emph{The Oxford handbook of causal reasoning}, pages 147--168.

\bibitem[{Wu et~al.(2012)Wu, Yu, and Chang}]{wu-etal-2012-detecting}
Jheng{-}Long Wu, Liang{-}Chih Yu, and Pei{-}Chann Chang. 2012.
\newblock \href {https://doi.org/10.1186/1472-6947-12-72} {Detecting causality from online psychiatric texts using inter-sentential language patterns}.
\newblock \emph{{BMC} Medical Informatics Decis. Mak.}, 12:72.

\bibitem[{Wu et~al.(2021)Wu, Ribeiro, Heer, and Weld}]{wu-etal-2021-polyjuice}
Tongshuang Wu, Marco~Tulio Ribeiro, Jeffrey Heer, and Daniel Weld. 2021.
\newblock \href {https://doi.org/10.18653/v1/2021.acl-long.523} {Polyjuice: Generating counterfactuals for explaining, evaluating, and improving models}.
\newblock In \emph{Proceedings of the 59th Annual Meeting of the Association for Computational Linguistics and the 11th International Joint Conference on Natural Language Processing (Volume 1: Long Papers)}, pages 6707--6723, Online. Association for Computational Linguistics.

\bibitem[{Xu et~al.(2020)Xu, Zuo, Liang, and Zuo}]{xu-etal-2020-review}
Jinghang Xu, Wanli Zuo, Shining Liang, and Xianglin Zuo. 2020.
\newblock \href {https://doi.org/10.18653/v1/2020.coling-main.133} {A review of dataset and labeling methods for causality extraction}.
\newblock In \emph{Proceedings of the 28th International Conference on Computational Linguistics}, pages 1519--1531, Barcelona, Spain (Online). International Committee on Computational Linguistics.

\bibitem[{Xu et~al.(2016)Xu, Wu, Zhang, Wang, Lee, and Xu}]{xu-etal-2016-cd}
Jun Xu, Yonghui Wu, Yaoyun Zhang, Jingqi Wang, Hee{-}Jin Lee, and Hua Xu. 2016.
\newblock \href {https://doi.org/10.1093/DATABASE/BAW036} {{CD-REST:} a system for extracting chemical-induced disease relation in literature}.
\newblock \emph{Database J. Biol. Databases Curation}, 2016.

\bibitem[{Yang et~al.(2022)Yang, Han, and Poon}]{yang-etal-2022-survey}
Jie Yang, Soyeon~Caren Han, and Josiah Poon. 2022.
\newblock \href {https://doi.org/10.1007/s10115-022-01665-w} {A survey on extraction of causal relations from natural language text}.
\newblock \emph{Knowl. Inf. Syst.}, 64(5):1161--1186.

\bibitem[{Yang et~al.(2020)Yang, Obadinma, Zhao, Zhang, Matwin, and Zhu}]{yang-etal-2020-semeval}
Xiaoyu Yang, Stephen Obadinma, Huasha Zhao, Qiong Zhang, Stan Matwin, and Xiaodan Zhu. 2020.
\newblock \href {https://doi.org/10.18653/v1/2020.semeval-1.40} {{S}em{E}val-2020 task 5: Counterfactual recognition}.
\newblock In \emph{Proceedings of the Fourteenth Workshop on Semantic Evaluation}, pages 322--335, Barcelona (online). International Committee for Computational Linguistics.

\bibitem[{Yang et~al.(2019)Yang, Dai, Yang, Carbonell, Salakhutdinov, and Le}]{yang-etal-2019-xlnet}
Zhilin Yang, Zihang Dai, Yiming Yang, Jaime~G. Carbonell, Ruslan Salakhutdinov, and Quoc~V. Le. 2019.
\newblock \href {https://proceedings.neurips.cc/paper/2019/hash/dc6a7e655d7e5840e66733e9ee67cc69-Abstract.html} {Xlnet: Generalized autoregressive pretraining for language understanding}.
\newblock In \emph{Advances in Neural Information Processing Systems 32: Annual Conference on Neural Information Processing Systems 2019, NeurIPS 2019, December 8-14, 2019, Vancouver, BC, Canada}, pages 5754--5764.

\bibitem[{Yao et~al.(2021)Yao, Chu, Li, Li, Gao, and Zhang}]{yao-etal-2021-survey}
Liuyi Yao, Zhixuan Chu, Sheng Li, Yaliang Li, Jing Gao, and Aidong Zhang. 2021.
\newblock \href {https://doi.org/10.1145/3444944} {A survey on causal inference}.
\newblock \emph{{ACM} Trans. Knowl. Discov. Data}, 15(5):74:1--74:46.

\bibitem[{Yao et~al.(2023)Yao, Yu, Zhao, Shafran, Griffiths, Cao, and Narasimhan}]{yao-etal-2023-tree}
Shunyu Yao, Dian Yu, Jeffrey Zhao, Izhak Shafran, Thomas~L. Griffiths, Yuan Cao, and Karthik Narasimhan. 2023.
\newblock \href {https://doi.org/10.48550/ARXIV.2305.10601} {Tree of thoughts: Deliberate problem solving with large language models}.
\newblock \emph{CoRR}, abs/2305.10601.

\bibitem[{Yarlett and Ramscar(2019)}]{yarlett-ramscar-2019-uncertainty}
Daniel Yarlett and Michael Ramscar. 2019.
\newblock \href {https://escholarship.org/content/qt2p34t5bt/qt2p34t5bt_noSplash_5d10218cba568d7bcf96df42e978b40c.pdf?t=op2jsp} {Uncertainty in causal and counterfactual inference}.
\newblock In \emph{Proceedings of the Twenty-fourth Annual Conference of the Cognitive Science Society}, pages 956--961. Routledge.

\bibitem[{Yu et~al.(2019)Yu, Li, and Wang}]{yu-etal-2019-detecting}
Bei Yu, Yingya Li, and Jun Wang. 2019.
\newblock \href {https://doi.org/10.18653/v1/D19-1473} {Detecting causal language use in science findings}.
\newblock In \emph{Proceedings of the 2019 Conference on Empirical Methods in Natural Language Processing and the 9th International Joint Conference on Natural Language Processing (EMNLP-IJCNLP)}, pages 4664--4674, Hong Kong, China. Association for Computational Linguistics.

\bibitem[{Yu et~al.(2023{\natexlab{a}})Yu, Wang, Golovneva, AlKhamissi, Verma, Jin, Ghosh, Diab, and Celikyilmaz}]{yu-etal-2023-alert}
Ping Yu, Tianlu Wang, Olga Golovneva, Badr AlKhamissi, Siddharth Verma, Zhijing Jin, Gargi Ghosh, Mona Diab, and Asli Celikyilmaz. 2023{\natexlab{a}}.
\newblock \href {https://doi.org/10.48550/arXiv.2212.08286} {{ALERT:} adapting language models to reasoning tasks}.
\newblock In \emph{Proceedings of the 61st Annual Meeting of the Association for Computational Linguistics (Volume 1: Long Papers)}, Toronto, Canada. Association for Computational Linguistics.

\bibitem[{Yu et~al.(2023{\natexlab{b}})Yu, Jiang, Clark, and Sabharwal}]{yu-etal-2023-ifqa}
Wenhao Yu, Meng Jiang, Peter Clark, and Ashish Sabharwal. 2023{\natexlab{b}}.
\newblock \href {https://arxiv.org/abs/2305.14010} {Ifqa: A dataset for open-domain question answering under counterfactual presuppositions}.
\newblock \emph{arXiv preprint arXiv:2305.14010}.

\bibitem[{Zang et~al.(2013)Zang, Cao, Cao, Wu, and Cao}]{zang-etal-2013-survey}
Liangjun Zang, Cong Cao, Yanan Cao, Yuming Wu, and Cungen Cao. 2013.
\newblock \href {https://doi.org/10.1007/s11390-013-1369-6} {A survey of commonsense knowledge acquisition}.
\newblock \emph{J. Comput. Sci. Technol.}, 28(4):689--719.

\bibitem[{Zeng and Wang(2022)}]{zeng-wang-2022-survey}
Jingying Zeng and Run Wang. 2022.
\newblock \href {https://arxiv.org/abs/2209.00869} {A survey of causal inference frameworks}.
\newblock \emph{arXiv preprint arXiv:2209.00869}.

\bibitem[{Zhang and Foo(2001)}]{zhang-foo-2001-epdl}
Dongmo Zhang and Norman~Y. Foo. 2001.
\newblock {EPDL:} {A} logic for causal reasoning.
\newblock In \emph{Proceedings of the Seventeenth International Joint Conference on Artificial Intelligence, {IJCAI} 2001, Seattle, Washington, USA, August 4-10, 2001}, pages 131--138. Morgan Kaufmann.

\bibitem[{Zhang et~al.(2020)Zhang, Liu, Pan, Song, and Leung}]{zhang-etal-2020-aser}
Hongming Zhang, Xin Liu, Haojie Pan, Yangqiu Song, and Cane~Wing{-}Ki Leung. 2020.
\newblock \href {https://doi.org/10.1145/3366423.3380107} {{ASER:} {A} large-scale eventuality knowledge graph}.
\newblock In \emph{{WWW} '20: The Web Conference 2020, Taipei, Taiwan, April 20-24, 2020}, pages 201--211. {ACM} / {IW3C2}.

\bibitem[{Zhang et~al.(2022)Zhang, Zhang, Su, and Roth}]{zhang-etal-2022-rock}
Jiayao Zhang, Hongming Zhang, Weijie~J. Su, and Dan Roth. 2022.
\newblock \href {https://proceedings.mlr.press/v162/zhang22am.html} {{ROCK:} causal inference principles for reasoning about commonsense causality}.
\newblock In \emph{International Conference on Machine Learning, {ICML} 2022, 17-23 July 2022, Baltimore, Maryland, {USA}}, volume 162 of \emph{Proceedings of Machine Learning Research}, pages 26750--26771. {PMLR}.

\bibitem[{Zhang et~al.(2023)Zhang, Zhang, Li, and Smola}]{zhang-etal-2023-automatic}
Zhuosheng Zhang, Aston Zhang, Mu~Li, and Alex Smola. 2023.
\newblock \href {https://openreview.net/pdf?id=5NTt8GFjUHkr} {Automatic chain of thought prompting in large language models}.
\newblock In \emph{The Eleventh International Conference on Learning Representations, {ICLR} 2023, Kigali, Rwanda, May 1-5, 2023}. OpenReview.net.

\bibitem[{Zhao et~al.(2016)Zhao, Liu, Zhao, Chen, and Nie}]{zhao-etal-2016-event}
Sendong Zhao, Ting Liu, Sicheng Zhao, Yiheng Chen, and Jian{-}Yun Nie. 2016.
\newblock \href {https://doi.org/10.1016/j.neucom.2015.09.066} {Event causality extraction based on connectives analysis}.
\newblock \emph{Neurocomputing}, 173:1943--1950.

\bibitem[{Zhou et~al.(2023)Zhou, Muresanu, Han, Paster, Pitis, Chan, and Ba}]{zhou-etal-2023-large}
Yongchao Zhou, Andrei~Ioan Muresanu, Ziwen Han, Keiran Paster, Silviu Pitis, Harris Chan, and Jimmy Ba. 2023.
\newblock \href {https://openreview.net/pdf?id=92gvk82DE-} {Large language models are human-level prompt engineers}.
\newblock In \emph{The Eleventh International Conference on Learning Representations, {ICLR} 2023, Kigali, Rwanda, May 1-5, 2023}. OpenReview.net.

\end{thebibliography}
\bibliographystyle{acl_natbib}
\appendix
\clearpage

\section{Real-World Applications of Commonsense Causality} \label{appendix:applications}
Commonsense causality has a wide range of applications in domains like medical diagnosis~\cite{richens-etal-2020-improving}, psychology~\cite{matute-etal-2015-illusions,eronen-2020-causal}, behavioral science~\cite{grunbaum-1952-causality}, economics~\cite{bronfenbrenner-1981-causality,hoover-2006-causality}, legal systems~\cite{williams-1961-causation,summers-2018-common}. Here we mainly detail two of them which include healthcare assistance~(\Appendix~\ref{appendix:applications:healthcare}) and forensic analysis~(\Appendix~\ref{appendix:applications:legal}). 
\subsection{Healthcare and Medical Assistance} \label{appendix:applications:healthcare}
The cornerstones for medicine or healthcare are the investigation of~\cite{russo-williamson-2007-interpreting}:
\begin{enumerate}
    \item What \textit{causes} diseases and pandemics to develop?
    \item What medicine and policy could \textit{stop} or \textit{prevent} the disease or pandemic?
\end{enumerate}

For these two core objectives, commonsense causality assists in various aspects: 
\begin{itemize}
    \item Medical Diagnosis: Medical professionals use commonsense to interpret symptoms and link them to particular diseases~\cite{richens-etal-2020-improving} 
    \item Disease Treatment and Prevention Program: A deep comprehension of causal relationships between certain lifestyles and diseases helps people to make better treatment and prevention plans. For instance, knowing that a sedentary lifestyle leads to Type 2 Diabetes will motivate people to exercise more to prevent illness. 
    \item Public Health Strategy: Commonsense causality is important for prudent public health strategy making~\cite{chiolero-2019-causality}. For example,  the causal relationship between air pollution and increasing numbers of pulmonary disease patients pushes the government to restrict emissions and promote clean energy. 
\end{itemize}

\subsection{Legal and Forensic Analysis} \label{appendix:applications:legal}
One of the most important applications of commonsense causality is understanding legal causation. 
As mentioned in Section 2.A of \cite{summers-2018-common}, commonsense has been a useful tool in determining legal causation. As Lord Reid put in \textit{Stapley v Gypsum Mines}:
\begin{quote}
To determine what caused an accident from the point of view of legal liability is a most difficult task. If there is any valid logical or scientific theory of causation it is quite irrelevant in this connection … The question must be determined by \underline{applying common sense} to the facts of each particular case. 
\end{quote}
There are various legal scenarios where commonsense causality plays an important role: 
\begin{itemize}
    \item Determining Legal Liability: Establishing causality is crucial for determining legal liability. Commonsense causality is helpful in judging whether a defendant's action leads to the plaintiff's loss~\cite{williams-1961-causation, summers-2018-common, hoekstra-breuker-2007-commonsense}. 
    \item Investigation of Criminal Intent and Motive: A comprehension of the causal relationship helps to understand the criminal motive. This assists judges with the sentencing of defendants and makes fair decisions. For instance, if one driver hits another car parked on the side of the road, commonsense causality helps to attribute the cause of the incident to the driver. 
\end{itemize}

\section{Preliminaries and Definitions} \label{appendix:background} In this section, we mainly introduce the preliminary knowledge about commonsense in \Appendix~\ref{appendix:background:commonsense} and then describe the qualitative reasoning tasks in \Appendix~\ref{appendix:background:reasoning}. 
Other more specific preliminary knowledge such as language models, causal concepts, linguistic causality, and causal inference is described in \Appendix~\ref{appendix:lm}, \ref{appendix:concepts}, \ref{appendix:linguistic}, and \ref{appendix:causal_inference}, respectively. 
To help readers refer back to the main body of the paper, this section corresponds to \S~\ref{sec:taxonomy:type}.

\subsection{Commonsense} \label{appendix:background:commonsense}
\SmallHeadingQuestion{What Is Commonsense} 
Commonsense in the domain of NLP refers to widely accepted knowledge that helps the majority of people understand the real world better like ``water flows from high to low'' and ``rain leads to slippery roads''. There are some aspects of commonsense: 
(i) World Knowledge Reasoning: Information about daily life such as ``When you are hungry you need to eat food''; 
(ii) Commonsense Causal Reasoning: Understanding the cause-effect relationship such as ``rain makes roads slippery''; 
(iii) Commonsense Temporal Reasoning: Understanding sequences of events and the concept of time order, e.g., ``Dessert usually comes after the main course'';
(iv) Commonsense Spatial Reasoning: Understanding the physical concept of space, e.g., ``a ball is placed inside a box instead of a bowl'' and ``a basketball is usually larger than a table tennis ball''; 
(v) Social Context: Comprehending the social norm, i.e., the accepted behaviors, practices, and values within a society. For instance, it's customary to bring a small gift when visiting someone's house; 
(vi) Counterfactual Reasoning: Reasoning over scenarios that didn't happen but could have. For instance, ``Had I noticed the 'Wet Floor' sign, I wouldn't have slipped''. 

\SmallHeading{Characteristics of Commonsense} Commonsense, by its inherent nature and definition, has distinctiveness like intuitiveness and universality. Beyond that, there are some aspects that are commonly ignored:  
(i) Contextual Dependency: The applicability of commonsense varies depending on the context. What is considered as commonsense in one culture may not seen in the same way, e.g., the thumbs-up gesture \thumbsup is viewed as approval in one culture but impoliteness in some other cultures; 
(ii) Time-Sensitiveness: Commonsense is evolving over time. What was perceived as commonsense previously is not commonsense now. A great example is the understanding of the solar system, it was commonsense to posit that Earth was the center around celestial bodies, i.e., the geocentric model. However, the heliocentric model become common nowadays, which believes that the Sun, rather than the Earth, is at the center;   
(iii) Error-Proneness or Inherent Uncertainties: Due to the aforementioned time-sensitiveness and contextual dependency, we can easily tell that there are inherent uncertainties in commonsense causality and it is prone to claim fake commonsense causality.

\SmallHeading{What Is Not Commonsense} In contrast to commonsense, non-commonsense knowledge includes
(i) Specialized Knowledge: Knowledge acquired via specific education, training, or experience is not within the realm of commonsense. For instance, comprehension of complex theories of mathematics or legal principles; 
(ii) Individual Subjectivity: Individual experience on certain cause-and-effect cannot be viewed as commonsense causality. For instance, if a person feels sleepy after drinking milk. Nevertheless, we cannot draw a causal relationship between milk drinking and being sleepy; 
(iii) Counterintuitive Facts: Some scientific facts are not commonsense knowledge during a certain period. For instance, the Earth revolving around the Sun was once a counterintuitive idea before the 14th century.

\subsection{Qualitative Reasoning Tasks Related to Commonsense Causality} \label{appendix:background:reasoning}

\SmallHeading{Causal Reasoning} Commonsense causal reasoning~(CCR) is the task of capturing causal dependencies between one event~(the cause) and the other~(the effect) based on human knowledge. Generally, these events are in textual format. Datasets like COPA~\cite{roemmele-etal-2011-choice}, TCR~\cite{ning-etal-2018-joint}, and e-CARE~\cite{du-etal-2022-e} follows the following format. Each question consists of a premise and two alternatives and the goal is to select the more plausible cause~(or effect) of the given premise. 
\begin{exampleblock}[frametitle={Example of Causal Reasoning}] \noindent Premise: The man broke his toe. What was the CAUSE of this? \par  \noindent Alternative 1: He got a hole in his sock. \par  \noindent Alternative 2: He dropped a hammer on his foot. \end{exampleblock}

\SmallHeading{Counterfactual Reasoning}
Counterfactual reasoning~\cite{goodman-1947-problem, bottou-etal-2013-counterfactual} describe possible outcomes that could happen had certain events happened, e.g., ``Had I brought an umbrella, I would not get wet''. It has been studied in various domains such as Psychology~\cite{roese-1994-functional, roese-morrison-2009-psychology}, Law~\cite{speer-etal-2017-conceptnet,venzke-2018-if}, Economics~\cite{pesaran-smith-2016-counterfactual}, Social Science~\cite{tetlock-belkin-1996-counterfactual}.

\section{Related Survey} \label{appendix:survey}
We provide different lines of surveys related to commonsense causality in Table~\ref{tab:surveys}. The related surveys can be categorized into five types: 
\begin{itemize}
    \item Surveys of Commonsense Reasoning: These surveys cover works from benchmarks~\cite{davis-2023-benchmark} to methods~\cite{bhargava-ng-2022-commonsense,qiao-etal-2023-reasoning} about reasoning with commonsense. 
    \item Surveys of Causal Knowledge Acquisition: Existing works cover datasets, methods, and evaluation metrics of the causality acquisition task. 
    \item Surveys of Causal Reasoning With Language Models: \citet{kiciman-etal-2023-causal} examine the ability of large language models in causal tasks like causal discovery, counterfactual inference, discerning necessary and sufficient causality via solely natural language input. 
    \item Surveys of Causal Inference: Except for textbooks of \citep{hernan-robins-2010-causal,pearl-etal-2016-causal,peters-etal-2017-elements}, there are surveys~\cite{yao-etal-2021-survey,zeng-wang-2022-survey} covering the benchmarks, application, and frameworks of causal inference. 
    \item Surveys of Probabilistic View of Causality: \cite{williamson-2009-probabilistic} review existing probabilistic theories of causality and analyze their failure examples critically. 
\end{itemize}

\begin{table}[htp!]
  \centering
  \resizebox{0.5\textwidth}{!}{
  \begin{tabular}{p{3.0cm}p{6.2cm}}
    \toprule
     \textbf{Citation} & \textbf{Summary} \\
    \midrule
    \rowcolor{mygray} \multicolumn{2}{c}{\textit{Commonsense Reasoning}}\\
    \cite{storks-etal-2019-commonsense} & A survey of existing benchmarks and methods for commonsense reasoning. \\
    \cite{bhargava-ng-2022-commonsense} & Survey about methods of utilizing pre-trained language model for commonsense knowledge reasoning and acquisition. \\
    \cite{qiao-etal-2023-reasoning} & Survey of different prompting methods for commonsense reasoning. \\
    \cite{davis-2023-benchmark} & Survey of 139 commonsense benchmarks: 102 text-based, 18 image-based, 12 video-based, and 7 physical simulation-based. Furthermore, this survey presents the definition and role of commonsense in AI, discusses the desirable nature of a commonsense benchmark, and shows the flaws of existing commonsense benchmarks. \\
    \rowcolor{mygray} \multicolumn{2}{c}{\textit{Causal Knowledge Acquisition}}\\
    \cite{zang-etal-2013-survey} & Survey about the methods and evaluation of existing commonsense knowledge acquisition systems. \\
    \cite{drury-etal-2022-survey} & Survey about extraction of causal relationships from text. \\
    \cite{xu-etal-2020-review} & Survey of datasets and labeling methods for causality extraction from text. \\
    \cite{feder-etal-2022-causal} & Survey for adapting important causal inference concepts into textual format. \\
    \cite{fitelson-hitchcock-2011-probabilistic} & Survey of methods for analyzing causal strength via probability. \\
    \cite{glymour-etal-2019-review} & A brief review of computational methods for causal discovery including constraint-based, score-based, and functional causal model-based methods.\\ 
    \cite{yang-etal-2022-survey} & Survey of causality extraction including taxonomies of causality extraction, benchmark datasets, and extraction techniques.  \\
    \cite{asghar-2016-automatic} & Survey of automatic extraction of causal relationship from natural language. \\
    \rowcolor{mygray} \multicolumn{2}{c}{\textit{Causal Reasoning}} \\
    \cite{kiciman-etal-2023-causal} & Survey of large language models' ability in performing causal discovery, which includes effect inference, attribution, and actual causality,  and understanding actual causality, which includes counterfactual reasoning, identifying necessary and sufficient causes. \\
    \rowcolor{mygray} \multicolumn{2}{c}{\textit{Causal Inference}} \\
    \cite{yao-etal-2021-survey} & A survey about causal inference under the potential outcome framework, benchmarks, and applications.  \\
    \cite{zeng-wang-2022-survey} & A review of past works that focus on outcomes framework and causal graphical models of causal inference. \\
    \rowcolor{mygray} \multicolumn{2}{c}{\textit{Probabilistic View of Causality}} \\
    \cite{williamson-2009-probabilistic} & Survey of probabilistic theories of causality, which includes the theories of Reichenbach~\cite{reichenbach-1956-direction}, Good~\cite{good-1961-causal}, and Suppes~\cite{suppes-1973-probabilistic}. 
    \\ \bottomrule
  \end{tabular}
  }
  \caption{Related surveys.}
  \label{tab:surveys}
\end{table}
Our survey sets itself apart by offering a comprehensive exploration of commonsense causality from a language perspective. Unlike the aforementioned surveys focusing on particular aspects, our works provide an overview of commonsense causality, covering the dimensions of benchmarks, taxonomies, acquisition methods, and both qualitative and quantitative measurements.

\section{More Taxonomies of Commonsense Causality} \label{appendix:taxonomy} 
Different criteria for categorizing commonsense causality lead to the development of distinct taxonomies, each offering a unique perspective on the organization and relationships of commonsense causality. Here we further supplement with taxonomies by skill sets~(\Appendix~\ref{appendix:taxonomy:skills}) and the nature of entities involved~(\Appendix~\ref{appendix:taxonomy:entities}). 
This section refers back to \S~\ref{sec:taxonomy}. 

\subsection{Classification by Skill Sets} \label{appendix:taxonomy:skills}
\begin{figure}[htp!]
    \centering
    \includegraphics[width=\linewidth]{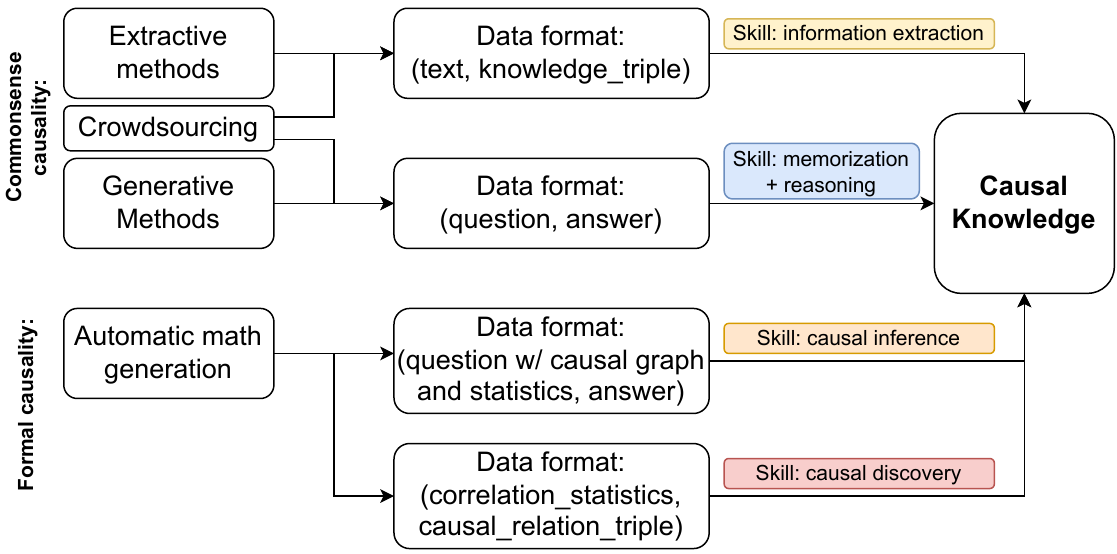}
    \caption{Overview of causal NLP tasks and required skill sets. }
    \label{fig:taxonomy_skill}
\end{figure}

We can classify the skill sets required by causal reasoning into two high-level types:
(1) \textit{Closed book} causality means tasks that can be completed by only looking at the given text, but not recalling external knowledge. This category can test skills such as (a) proper linguistic understanding of the given text, as in information extraction, such as causal relation extraction \cite{do-etal-2011-minimally,hidey-mckeown-2016-identifying}, counterfactual statement identification \cite{hendrickx-etal-2010-semeval}, or (b) formal reasoning on the given conditions and statistics, using skills such as causal inference \cite{jin2023cladder}, and causal discovery \cite{jin2024large,jin-etal-2022-logical}.
(2) \textit{Open book} causality refers to tasks that require external knowledge out of the provided text, which usually includes (a) questions about a causal relation directly, such as asking about the relation between two events, the effect given the cause, or the cause given the effect, or (b) counterfactual reasoning, where an alternative condition is given and asks for the outcome. As indicated in Figure~\ref{fig:taxonomy_skill}, open book causality requires memorization skills and reasoning.

\subsection{Classification by Nature of Entities Involved} \label{appendix:taxonomy:entities}
Based on the nature of the entities involved, commonsense causality can be further classified into physical commonsense causality and social commonsense causality. Physical commonsense causality usually involves non-human entities like inanimate objects or natural phenomena. However, social commonsense causality always involves humans, human behavior, social norms, cultures, etc. 
\begin{itemize}
    \item Physical Commonsense Causality: It usually occurs in the context of the physical or natural world and is governed by the laws and principles of mathematics, physics, biology, and physics. Generally, it is more predictable and context-free. 
    \item Social Commonsense Causality: Different from physical causality, social causality is governed by social background, cultural norms, etc. It is less predictable and relies heavily on social context. It is often observed in the domains of sociology, psychology, and related disciplines. 
\end{itemize}

There are many other taxonomies for commonsense causality, which is beyond the scope of this survey.

\section{Uncertainty in Commonsense Causality} \label{appendix:uncertainty} 
Uncertainty is almost everywhere, no exception for commonsense causality. 
We summarize all sources of uncertainties over commonsense causality into two categories~\cite{yarlett-ramscar-2019-uncertainty}: factual uncertainties~(\Appendix~\ref{appendix:uncertainty:factual}) and causal uncertainties~(\Appendix~\ref{appendix:uncertainty:causal}). 
This section corresponds to \S~\ref{sec:taxonomy:uncertainty}.

\subsection{Factual Uncertainties} \label{appendix:uncertainty:factual}
Factual uncertainties are due to the principle that the observation or description of contextual information of the cause or effect can never be complete. The factual uncertainties can be further classified into the following subcategories: 
\begin{itemize}
    \item Incomplete Observation: The observation of the world is hardly complete. For instance, it is the commonsense knowledge that exercise leads to fatigue. However, a small amount of exercise actually makes people more energetic rather than exhausted. 
    \item Contextual Uncertainty: It arises when the context of the cause or effect introduces ambiguity about the facts. For instance, when determining the cause of certain symptoms, the symptom descriptions heavily depend on the medical diagnosis equipment, which causes uncertainty in the determination of the true cause for diagnosis. 
    \item Temporal Uncertainty: Due to the time-sensitive characteristic of commonsense, commonsense is inherently vulnerable to temporal uncertainty. For instance, historically, the need for light~(the cause) leads to using candles(the effect). However, after the widespread adoption of electricity and bulbs, this causal relationship doesn't hold anymore. 
\end{itemize}

\subsection{Causal Uncertainties} \label{appendix:uncertainty:causal}
Causal uncertainties arise in cases where the cause is not invariably followed by the effect. For instance, we all know that smoking contributes to the occurrence of lung cancers. However, some people smoke a lot but do not suffer from lung cancer. Similar situations can be found in examples like ``clouds lead to rain'', but there are days there are a lot of clouds but no rain at all. 
The causal uncertainties can be further divided into the following subcategories: 
\begin{itemize}
    \item Probabilistic Causation: It refers to the situation wherein causes increase the likelihood of but do not guarantee the occurrence of effects. This is also the focus in the \S~\ref{sec:analysis:quantitative}. Examples include ``not all smokers get lung cancer'', ``a healthy diet does not guarantee longevity'', etc. 
    \item Complex Interaction: Complex causal structures like co-founder, collider, causal chain, triangular causality, and the combination of these basic structures lead to significant complexity and introduce additional uncertainties. 
    \item Causal Loops: Though causal loops can be included in the category of complex interaction, we define them separately, hoping it draws particular attention. There are scenarios where the effect also influences the cause, forming a causal loop. For example, poverty results in poor education opportunities, which in turn aggravates poverty. 
    A similar example in the domain of the environment is the feedback loop between global warming and ice glacier melting. Global warming speeds up the melting of ice glaciers. Without ice to reflect back the sunlight, more solar energy research to the surface of the Earth and thus perpetuate global warming. This phenomenon is also observed in the marketing area: high-quality products reinforce the marketing share, which in turn empowers companies' ability to develop better products. 
\end{itemize}

Besides, the uncertainty of causality in other domains like medical~\cite{kratenko-2022-problem} and legal domains~\cite{weinrib-2016-causal} is also investigated. However, due to the page limit, we will not discuss these topics in this survey.

\section{More Topics on Causality Acquisition} \label{appendix:acquisition}
In this section, we cover some supplementary topics related to commonsense causality acquisition, including extraction methods for implicit and inter-sentential causality~(\Appendix~\ref{appendix:acquisition:other_causality}), and details of manual annotation schemes~(\Appendix~\ref{appendix:acquisition:annotation}). 
This section corresponds to \S~\ref{sec:acquisition} that is about causality acquisition methods. 
\begin{table*}[htp!]
    \centering
    \resizebox{0.95\textwidth}{!}{
    \begin{tabular}{lcccccc} 
        \toprule 
        \textbf{Method} & \textbf{Accuracy/Quality} & \textbf{Cost/Efficiency} & \textbf{Coverage} & \textbf{Adaptability} & \textbf{Scalability} & \textbf{Explainablity}\\
        \midrule
        Extractive &  \fourstars{}  &  \fivestars{} &  \fourstars{} & \fourstars{} & \fivestars{} & \fourstars{} \\
        Generative & \threestars{} & \fourstars{} & \fivestars{} & \threestars{} & \fivestars{} & \onestar{} \\
        Manual Annotation & \fivestars{} & \twostars{} & \twostars{} & \fivestars{} & \onestar{} & \fivestars{} \\
        \bottomrule 
    \end{tabular}
    }
    \caption{Comparison of different commonsense causality acquisition methods. The more solid stars, the better. }
    \label{tab:acquisition_comparison_detailed}
\end{table*}

\subsection{Extraction of Different Kinds of Causality} \label{appendix:acquisition:other_causality}
\SmallHeading{Extraction of Implicit Causality} 
Since causality can be expressed in various ways, the extraction of implicit causality~\cite{hartshorne-2014-implicit,asr-demberg-2012-implicitness}~\footnote{The boundary between explicit causality and implicit causality is unclear. Here, we refer to causality that lacks explicit indicators such as ``because'', ``due to'', etc., as implicit causality. } is even more challenging than the extraction of explicit causality with linguistic indicators such as ``because'', ``due to'', ``lead to'', etc. 
\begin{center}
\begin{minipage}{0.4\textwidth}
\begin{exampleblock}[frametitle={\small Example of implicit causality}]
\small Tom got caught in a heavy rain yesterday and worked with a fever today. 
\end{exampleblock}
\end{minipage}
\end{center}

For implicit causality, it is infeasible to use linguistic patterns to detect the presence of causality. There are two approaches to extracting implicit causality: 
\begin{itemize}
    \item Utilizing External Knowledge Bases: These works~\cite{ittoo-bouma-2011-extracting,kruengkrai-etal-2017-improving} utilize external knowledge to enhance implicit causality extraction and alleviate the need for manually annotated data. \citet{xu-etal-2016-cd} used document-level classifier, 
    \item Learning-Based Approach~\cite{airola-etal-2008-graph, kruengkrai-etal-2017-improving}: They use background knowledge and the original sentences as the features to train models for extracting causality. The key limitation is the lack of supervised learning data for model training. 
\end{itemize}

\SmallHeading{Extraction of Inter-Sentential Causality} 
Besides, different from intra-sentential causality, wherein inter-sentential causality, the cause and the effect lie in two sentences. 
As the following example shows, the inter-sentential causal relation between ```paper deadline'' and ``went to sleep earlier than before'' is difficult to identify due to the lack of causal connectives.  
\begin{center}
\begin{minipage}{0.4\textwidth} 
\begin{exampleblock}[frametitle={\small Example of Inter-Sentential causality}]
\small I was tired last night due to a paper deadline. I went to sleep earlier than before. 
\end{exampleblock}
\end{minipage}
\end{center}

For inter-sentential causality, there are two extraction approaches~\footnote{Most of the intra-sentential causality extraction methods still apply to inter-sentential causality well. Here, we only name several methods specifically designed for inter-sentential causality extraction. }
\begin{itemize}
    \item Linguistic Pattern Matching: \citet{ittoo-bouma-2011-extracting,wu-etal-2012-detecting} extend the pattern matching methods for causality detection to the inter-sentential causality. \citet{jin-etal-2020-inter} propose a cascaded multi-Structure Neural Network~(CSNN) to extract inter-sentential causality without dependency on external knowledge. 
    \item Learning-Based Approach:  \citet{swampillai-stevenson-2011-extracting} propose an approach that works for both intra-sentential and inter-sentential causality extraction. They use adapted features and techniques to deal with the special issues due to the inter-sentential cases. 
\end{itemize}

\subsection{Manual Annotation Schemes of Causation} \label{appendix:acquisition:annotation}
Existing manual annotation schemes can be roughly classified into three types~\cite{cao-etal-2022-cognitive}: 
\begin{itemize}
    \item Trigger Scheme: A manual annotation scheme based on the template of \textit{cause argument, trigger, effect argument}. Inside the template, triggers usually are conjunctions, adverbials, and causation verbs that indicate causation. Manual annotation schemes like BECausSE~\cite{dunietz-2018-annotating}, PDTB~\cite{webber-etal-2019-penn} fall into this category.
    \item CEP Scheme: A manual annotation scheme based on \textsc{Cause, Enable, Prevent}~(CEP) causal relationship. CEP scheme is based on the force dynamics theory of causation~\cite{wolff-2007-representing,wollf-shepard-2013-chapter}. This category covers manual annotation schemes including CCEP~\cite{cao-etal-2022-cognitive}, CaTeRS~\cite{mostafazadeh-etal-2016-caters}, and BECauSE~\cite{dunietz-2018-annotating}.  
    \item Joint Scheme: A manual annotation scheme that jointly annotates causality and temporality. The annotation methods like CaTeRS~\cite{mostafazadeh-etal-2016-caters}, ESL~\cite{caselli-vossen-2017-event} are included in this category. 
\end{itemize}

\begin{figure}[htp!]
\centering
\resizebox{0.85\columnwidth}{!}{
\begin{tikzpicture}[every node/.style={font=\scriptsize}]
  \draw[fill=red, opacity=0.5] (0,0) circle (1.5cm);
  \draw[fill=green, opacity=0.5] (1.5,0) circle (1.5cm);
  \draw[fill=blue, opacity=0.5] (3.0,0) circle (1.5cm);
  \draw (0.8,0) node {BECauSE};  
  \draw (-0.6,0) node {PDTB};  
  \draw (2.2,0) node {CaTeRS};  
  \draw (3.6,0) node {ESL};  
  \draw (1.5,-1) node {CCEP};    
\end{tikzpicture}
}
\caption{Relation of different manual annotation schemes. }
\label{fig:annotation_schemes}
\end{figure}
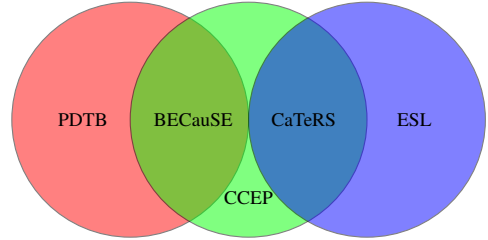

The relations between these three manual annotation schemes can be seen in Figure~\ref{fig:annotation_schemes}. 

\subsection{Strengths and Weaknesses of Different Causality Acquisition Methods} \label{appendix:acquisition:comparison} 
As shown in Table~\ref{tab:acquisition_comparison_detailed}, we analyze the strengths and weaknesses of the extractive methods, generative methods, and manual annotation from different aspects: 
\begin{itemize}
    \item Quality: Generative methods may give poor quality output, even generative LLMs are still suffering from hallucination problems. Extractive methods highly depend on the quality of the source data and are influenced by the extraction methods. However, humans have the capacity to perceive nuanced causal relationships and thus contribute high-quality commonsense causality.  
    \item Collection Cost: Manual annotation is the most labor-intensive and costly due to human labor. The extractive methods can process a large amount of sources. The generative model, however, is a bit more costly than extractive methods due to the training cost of generative models, even invoking the API can become costly if the datasets are large. 
    \item Collection Efficiency: It is self-evident that manual annotation is quite slow. Extractive methods are the most efficient while the generative methods are between the two regarding collection efficiency.  
    \item Coverage: The scale of generative datasets can be very large due to the flexibility of generative methods. The scale of extractive datasets is subject to the size of the source data. Due to the cost and efficiency concerns, the scale of manually annotated datasets is relatively small compared with extractive or generative methods.  
    \item Adaptability: Generative methods are the least adaptive methods as they heavily rely on the domain of training datasets. Extractive methods are more adaptable but are limited by predefined patterns, which can vary across different domains. Manual annotations, however, are the most adaptable as humans can easily adapt to new domains and emerging commonsense knowledge. 
    \item Scalability: It is obvious that the scalability of manual annotation is poor due to the cost and efficiency concerns while both generative and extractive are more scalable and are free from these concerns. 
    \item Explainability: It is well-known that the generative methods lack interpretability and explainability due to the block-box characteristic of large models. Extractive methods are better as the matching patterns are explicit and defined by users. Manual annotation is the most explainable as humans can well explain the causal relationships they create. 
\end{itemize}

\section{Details About Commonsense Causality Benchmarks} \label{appendix:detailed_benchmarks} 
\newcommand{\bcs}[0]{\newline}
We list the details of these benchmarks in Table~\ref{tab:detailed_benchmarks} including the annotation unit, number of causation in the whole dataset, brief introduction, and the license for more responsible research. This section corresponds back to the benchmark introduction in \S~\ref{sec:taxonomy}. 
\onecolumn
{
\begin{center}
\begin{longtable} {p{2.05cm}p{1.4cm}lp{1.0cm}p{0.5cm}p{5.4cm}p{1.5cm}}
    \toprule
     & Annotation Unit & \#Overall & \#Causal  & C.F.~\footnotemark[1]  & Brief introduction & License\\
    \midrule
    \rowcolor{mygray} \multicolumn{7}{c}{\textit{First-Principle Causality}}\\
    CauseEffectPairs\bcs\scriptsize{~\cite{mooij-etal-2016-distinguishing}} & Variable & 108 & 108 & \bempty & 108 different cause-effect pairs selected from 37 datasets which cover domains like meteorology, economy, medicine, engineering, biology.  It focuses on the causal discovery problem whose goal is to decide whether X causes Y or Y causes X, given the co-existence of two variables X and Y. &  FreeBSD \\
    IHDP\bcs\scriptsize{~\cite{shalit-etal-2017-estimating}} & Variable & 2,000 & 2,000 & \bhalf & IHDP, the Infant Health and Development Program dataset, is about the effect of home visit on cognitive test scores for infants.  & Custom Dataset Terms \\
    CRAFT\bcs\scriptsize{~\cite{ates-etal-2022-craft}} & Video & 58,000 & - & \bfull & A new video question answering dataset that needs comprehension of physical forces and object interactions. CRAFT contains descriptive and counterfactual questions. & MIT \\
    \rowcolor{mygray} \multicolumn{7}{c}{\textit{Commonsense Causality in Text Format}}\\
    Temporal-Causal\bcs\scriptsize{\cite{bethard-etal-2008-building}} &  Clause & 1,000  & 271  & \bempty & A corpus of 1,000 event pairs for both temporal and causal relations.  & Missing \\
    CW\bcs\scriptsize{\cite{ferguson-sanford-2008-anomalies}} &  Clause & 128  & 128  & \bfull  & CW, Counterfactual-World, is collected from existing psycholinguistic experiments. & Missing \\
    SemEval07-T4\bcs\scriptsize{\cite{girju-etal-2007-semeval}} &  Phrase & 220  & 114  & \bempty & SemEval07-T4 is not specific for causal relations. It focuses on semantic analysis, i.e., automatic recognition of relations between pairs of words, of which causal relation exists. & Missing\\
    SemEval10-T8\bcs\scriptsize{\cite{hendrickx-etal-2010-semeval}} &  Phrase & 10,717  & 1,331  & \bempty & Similar as the dataset in SemEval07-T4, it focuses on the automatic classification of semantic relations between pairs of nominals, which covers the cause-effect relations. & CC BY 3.0 Unported \\
    COPA\bcs\scriptsize{\cite{roemmele-etal-2011-choice}} &  Sentence & 2,000 & 1,000  & \bempty & Each question consists of a premise and two plausible causes or effect, where the correct one is more plausible than the other.  & BSD 2-Clause\\
    EventCausality\bcs\scriptsize{\cite{do-etal-2011-minimally}} &  Clause & 583  & 583  & \bempty  & \cite{do-etal-2011-minimally} used the discourse connectives and the particular discourse relation to detect causality between events and built a causality corpus.  & Missing \\
    BioCause\bcs\scriptsize{\cite{mihuailua-etal-2013-biocause}} &  Clause & 851  & 851  & \bempty & BioCause contains 851 causal relations collected from 19 biomedical journal articles in the domain of infectious disease. & Creative Commons\\
    CausalTimeBank\bcs\scriptsize{\cite{mirza-etal-2014-annotating}} &  Sentence & 318  & 318  & \bempty & CausalTimeBank is the Timebank corpus with causal samples taken from TempEval-3 corpus.  & CC BY-NC-SA 3.0\\
    CaTeRs\bcs\scriptsize{\cite{mostafazadeh-etal-2016-caters}} &  Sentence & 2,502 & 308  & \bempty  & Annotation of a total of 1,600 sentences in the context of five-sentences storied selected from ROCStories corpus~\cite{mostafazadeh-etal-2016-corpus}. CaTeRS is the name of the annotation scheme. & Missing \\
    AltLex\bcs\scriptsize{\cite{hidey-mckeown-2016-identifying}} &  Clause & 44,240 & 4,595  & \bempty & AltLex is an open class of markers that contains causality. & Missing\\
    BECauSE 2.0\bcs\scriptsize{\cite{dunietz-etal-2017-corpus}} &  Sentence & 729 & 554  & \bempty & BECauSE focus on annotations with causal relations and other relations that co-exists with causation. & MIT  \\
    ESL\bcs\scriptsize{\cite{caselli-vossen-2017-event}} &  Sentence & 2,608 & 2,608  & \bempty  & ESL is a corpus for causal and temporal relation detection. & CC BY 3.0 Unported \\
    TCR\bcs\scriptsize{\cite{ning-etal-2018-joint}} &  Clause & 172 & 172  & \bempty  & Similar as CaTeRS, TCR also focuses on temporal and causal relations. The difference here is that CaTeRs specifically focus on ROCStory dataset. & Missing \\
    PDTB\bcs\scriptsize{\cite{webber-etal-2019-penn}} &  Clause & 7,991 & 7,991  & \bempty & PDTB is more general dataset that marks the discourse relations which include causation. The core idea of PDTB is that discourse relations are grounded in explicit words or phrases or the adjacency of two sentences. & LDC User Agreement for Non-Members\\
    TimeTravel\bcs\scriptsize{~\cite{qin-etal-2019-counterfactual}} & Sentence & 109,964 & 29,849 & \bhalf  & TimeTravel originates from the ROCStories~\cite{mostafazadeh-etal-2016-corpus}. Each sample consists of the original story, a counterfactual fact, and a new storyline compatible with the counterfactual fact. & MIT\\
    GLUCOSE & Clause & 670K & 670K & \bempty & Given a short story as the context and a sentence $X$ in the story, GLUCOSE annotates 10 dimensions of causal explanation for $X$, which concerns events, location, possession, etc. These dimensions are usually implicit causes and effects of $X$. & Creative Commons Attribution-NonCommercial 4.0 International Public License \\
    XCOPA\bcs\scriptsize{~\cite{ponti-etal-2020-xcopa}} & Sentence & 11,000 & 11,000 & \bempty  & The multilingual version~(11 languages) of the COPA dataset. & CC BY 4.0\\
    SemEval20-T5\bcs\scriptsize{\cite{yang-etal-2020-semeval}} &  Clause & 25,501  & 25,501  & \bfull & This is the dataset associated with SemEval-2020 Task 5. There are two subtasks. The first subtask is about determining whether a statement is a counterfactual or not while the second subtask is to extract the antecedent and consequent in a given counterfactual statement. & Missing\\
    CausalBank\bcs\scriptsize{~\cite{li-etal-2021-guided}} & Clause & 314M & 314M & \bempty & CausalBank consists of cause-effect statements collected from the Common Crawl corpus using causal lexical patterns.   \\
    e-CARE\bcs\scriptsize{\cite{du-etal-2022-e}} & Sentence & 21,324 & 21,324  & \bempty  & Each sample in e-CARE consists of a cause-effect pair and a conceptual explanation for the causation. & MIT\\
    CoSIm\bcs\scriptsize{\cite{kim-etal-2022-cosim}} & Image + Text & 3,500 & 3,500 & \bfull & CoSIm is a multimodal counterfactual reasoning dataset. CoSIm focuses on commonsense reasoning for counterfactual scene imagination. Each sample in CoSIM consists of an image, an original question-answer pair, an imagined scene change in text format, and a question-answer pair under this imagined scene change. CoSIm stands as a touchstone for models' multimodal counterfactual ability. & MIT \\
    CRASS\bcs\scriptsize{\cite{frohberg-binder-2022-crass}} & Sentence & 274 & 274 & \bfull & It focuses on counterfactual reasoning setting that is in the format of question-answering.  & Apache 2.0 \\
    COPES\bcs\scriptsize{~\cite{wang-etal-2023-cola}} & Sentence & 1,360 & 1,360 & \bempty & COPES is a corpus collected from ROCStories~\cite{mostafazadeh-etal-2016-corpus} for studying contextualized commonsense causal reasoning. Each sample consists of five chronologically ordered events. Annotators annotate whether the first four events contribute to the occurrence of the last event or not. COPES can study causal reasoning when contextual information is given. & MIT\\
    IfQA\bcs\scriptsize{~\cite{yu-etal-2023-ifqa}} & Sentence & 3,800 & 3,800 & \bfull  & IfQA is a counterfactual open-domain question-answering datasets. Each question is a counterfactual proposition with an \textit{if} clause. A typical example is ``What would be the time difference between Los Angeles and Paris if Los Angeles was on the east coast of the U.S.?'' & Missing\\
    CW-extended\bcs\scriptsize{~\cite{li-etal-2023-counterfactual}} & Sentence & 10,848 & 10,848 & \bfull  & CW-extended is a counterfactual dataset that is the augmentation of the original CW~\cite{ferguson-sanford-2008-anomalies} by words replacement. For instance, in the counterfactual statement that ``If cats were vegetarians, families would feed them with cabbages.'' Replacement of \textit{cats} with \textit{dogs} would obtain a new counterfactual statement ``If dogs were vegetarians, families would feed them with cabbages.'' & Missing\\
    CausalQuest \cite{ceraolo-etal-2024-causalquest} & Sentence & 13,500 & 13,500 & \bhalf & CausalQuest is a dataset comprising natural causal questions collected from social networks, search engines, and AI assistants. It encompasses questions commonly asked by humans seeking to understand causation, categorized into queries about effects, causes, and causal relationships. & Apache 2.0\\
    $\delta$-CAUSAL \cite{cui-etal-2024-exploring} & Sentence & 11,245 & 11,245 & \bhalf & $\delta$-CAUSAL is a causal dataset studying the feasibility in causality. Each sample in $\delta$-CAUSAL consists of a cause-effect pair associated with its supporting and opposing arguments. This dataset introduces the idea of uncertainty/defeasibility inherent in commonsense causality. & MIT\\
    \rowcolor{mygray} \multicolumn{7}{c}{\textit{Commonsense Causality in Knowledge Graph Format}}\\ 
    CausalNet\bcs\scriptsize{\cite{luo-etal-2016-commonsense}} &  Word & 11M & 11M  & \bempty & CausalNet consists of a vast amount of causal relationships from Bing web pages. Each causal relationship is a triple in the format of  \textit{(cause\_word, effect\_word, frequency)}. & Missing \\
    ConceptNet\bcs\scriptsize{~\cite{speer-etal-2017-conceptnet}} & Phrase & 473,000 & - & \bempty  & ConceptNet is the knowledge graph version of the Open Mind Common Sense project. There are causal relations like \textit{causes, entails, derived from and so on} inside ConceptNet relation set.   & CC BY-SA 4.0\\
    Event2Mind\bcs\scriptsize{~\cite{rashkin-etal-2018-event2mind}} & Phrase & 25,000 & - & \bempty & Each sample in Event2Mind consists of a given event, and its intent and reaction. For instance, given the event ``PersonX reads PersonY's diary'', X's intent is ``to be nosey, know secrets'', X's reaction is ``guilty, curious'', and Y's reaction is ``angry, violated, betrayed''. Some of the samples in Event2Mind contain causal relationships. & MIT\\
    ATOMIC\bcs\scriptsize{~\cite{sap-etal-2019-atomic}} & Sentence & 877K & - & \bhalf  & ATOMIC collect commonsense knowledge in the format of \textit{if-then} relations. For instance, ``if X pays Y a compliment, then Y will likely return the compliment''. Obviously, ATOMIC contains event pairs that are in causal relations.  & CC BY 4.0 \\
    ASER\bcs\scriptsize{~\cite{zhang-etal-2020-aser}} & Sentence & 64M & 494K & \bempty  & ASER is an eventuality knowledge graph extracted from more than 11-billion-token unstructured textual data. It contains \textbf{a}ctivities, \textbf{s}tates, \textbf{e}vents, and their \textbf{r}elations. & MIT\\
    CauseNet\bcs\scriptsize{~\cite{heindorf-etal-2020-causenet}} & Word & 11M & 11M & \bempty & CauseNet consists of causal relations extracted from sem- and unstructured web sources(web pages and wikipedia) with precision of 83\%. & CC BY 4.0  \\    
    CEGraph\bcs\scriptsize{~\cite{li-etal-2021-guided}} & Phrase & 89.1M & 89.1M & \bempty &  A large lexical causal knowledge graphs (Cause Effect Graph) associated with CausalBank. Similar as CausalNet~\cite{luo-etal-2016-commonsense}, the nodes in the graph are lemmatized terms, and a directed edge between two terms represents a causal relation. The weights of each edge is cooccurrence frequency.   & Missing\\
    \bottomrule
  \caption{Overview of causal datasets.}
  \label{tab:detailed_benchmarks}
\end{longtable}
\end{center}
}
\twocolumn

\section{Language Models and Prompt Engineering for Large Language Models} \label{appendix:lm}
We mention the important role of language models in qualitative causal reasoning. In this section, we cover preliminary knowledge of language models~(\Appendix~\ref{appendix:lm:lm}), large language models~(\Appendix~\ref{appendix:lm:llms}), and prompting techniques for LLMs~(\Appendix~\ref{appendix:lm:prompting}). 
This section elaborates more details about \S~\ref{sec:analysis:qualitative}.

\subsection{Language Models} \label{appendix:lm:lm}
Before the era of transformer-based pre-trained language models, there are various task-specific models such as CRF~\cite{sutton-etal-2012-introduction}, HMM, LSTM~\cite{hochreiter-etal-1997-long}, GRU~\cite{cho-etal-2014-learning}, SeqSeq~\cite{sutskever-etal-2014-sequence}, Word2Vec~\cite{mikolov-etal-2013-efficient}, GloVe~\cite{pennington-etal-2014-glove}, CNN for text classification~\cite{kim-2014-convolutional}, etc. However, these models are unsatisfactory in performance and are mostly designed for specific tasks.

Starting from BERT~\cite{devlin-etal-2019-bert}, transformer-based~\cite{vaswani-etal-2017-attention} pre-trained models have become the foundation models that have omnipotent capacities in language, vision, robotics, and reasoning. These foundation models handle surprisingly well data in the format of text, images, speech, structured data, 3D signals, etc~\cite{bommasani-etal-2021-opportunities}. There are various kinds of taxonomy for transformer-based pre-trained language models. Here we classify these models by their model structure: 
\begin{itemize}
    \item Unidirectional Models: This series of language models is also known as decoder-based models. It includes the series of GPTs(GPT, GPT-2.0, GPT-3, GPT-3.5, GPT-4.0, GPT-Neo, etc), which is the model where ChatGPT builds upon, XLNet~\cite{yang-etal-2019-xlnet}, LLaMA~\cite{touvron-etal-2023-llama}, etc. 
    \item Bidirectional Models: This series of language models is also known as encoder-based models. It includes the series of models derived from BERT~\cite{devlin-etal-2019-bert} such as RoBERTa~\cite{liu-etal-2019-roberta} and DeBERTa~\cite{he-etal-2021-deberta}, lite BERT like ALBERT~\cite{lan-etal-2020-albert} and DistilBERT~\cite{sanh-etal-2019-distilbert}, and domain-specific models like BioBERT~\cite{lee-etal-2019-biobert}, LegalBERT~\cite{chalkidis-etal-2020-legal}, language specific models like ERNIE~\cite{sun-etal-2019-ernie}, etc. 
    \item Sequence-to-Sequence Models: This series of language models is also known as encoder-decoder-based models. Typical examples are T5~\cite{raffel-etal-2020-exploring}, BART~\cite{lewis-etal-2020-bart}, 
\end{itemize}

\subsection{Large Language Models} \label{appendix:lm:llms}
\begin{figure}[htp!]
\centering
\resizebox{\columnwidth}{!}{
\definecolor{outercolor}{RGB}{102, 178, 255} 
\definecolor{middlecolor}{RGB}{255, 204, 153} 
\definecolor{innercolor}{RGB}{204, 229, 255} 
\begin{tikzpicture}
    \filldraw[fill=outercolor] (0,6) circle (6);
    \filldraw[fill=middlecolor] (0,5) circle (4);
    \filldraw[fill=innercolor] (0,4) circle (2);
    \draw (0,11.5) node {\Large \textcolor{blue}{\textbf{NLP Models}}};
    \draw (0,8.0) node {\Large \textcolor{blue}{\textbf{Pretrained LMs}}};
    \draw (0,5.5) node {\Large \textcolor{blue}{\textbf{LLMs}}};
    \draw (-3,9.5) node {Word2Vec};
    \draw (-1,9.5) node {GloVe};
    \draw (-3,10.5) node {GRU};
    \draw (-1,10.5) node {LSTM};    
    \draw (2,9.5) node {FastText};  
    \draw (2,10.5) node {Seq2Seq};    
    \draw (5,7) node {\rotatebox{-60}{Attention}};  
    \draw (-5,7) node {\rotatebox{60}{CNN for Text}}; 
    \draw (-4.5,4) node {\rotatebox{-60}{Memory Network}};    
    \draw (-3.0,1.5) node {\rotatebox{-60}{CRF}};    
    \draw (3.0,1.5) node {\rotatebox{60}{HMM}};    

    \draw (0,7.4) node {Transformer};    
    \draw (-2,6.8) node {BERT}; 
    \draw (0,6.8) node {T5};    
    \draw (2,6.8) node {GPT};    
    \draw (3,5.5) node {XLNet};   
    \draw (-3,5.5) node {ERNIE};   
    \draw (-3,4.5) node {RoBERTa};    
    \draw (3,4.5) node {BART};  
    
    \draw (0,4.8) node {ChatGPT};  
    \draw (-1,4.0) node {*-Large};   
    \draw (0.9,4.0) node {GPT-3.5/4.0};      
    \draw (-1,3.3) node {LaMDA};      
    \draw (0.7,3.3) node {LLaMA};      
    \draw (0,2.7) node {PaLM};      

\end{tikzpicture}
}
\caption{Relation of NLP models. }
\label{fig:model_relation}
\end{figure}
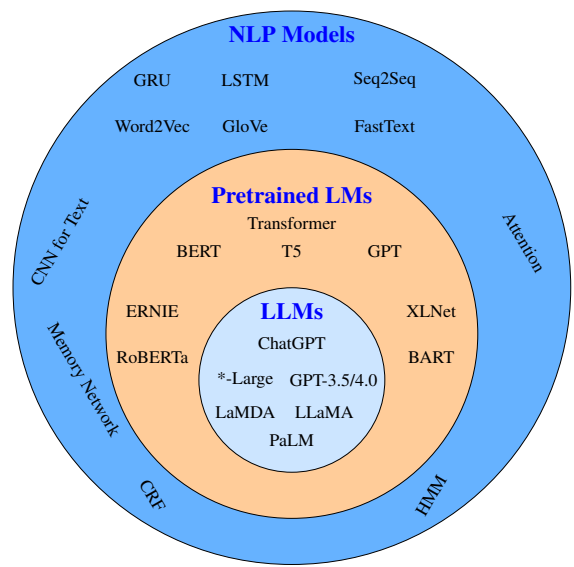

The relation between NLP models, language models, and large language models~(LLMs) is shown in Figure~\ref{fig:model_relation}. Multiple researches~\cite{wei-etal-2022-chain,liang-etal-2022-holistic} have revealed the fact that ``with the increase in model size, the reasoning ability of the model has undergone a change from quantitative to qualitative.'' 
During these two years, various LLMs have been proposed such as GPT-3.5, GPT-4.0, LaMDA~\cite{thoppilan-etal-2022-lamda}, PaLM~\cite{chowdhery-etal-2023-palm}, and LLaMA~\cite{touvron-etal-2023-llama}. 

\subsection{Prompting Techniques}
\label{appendix:lm:prompting}
As LLMs have become the foundation models of NLP tasks, how to fully utilize the power of these LLMs is a key point. Prompting is one of the key techniques. A prompt usually contains one or more of the following elements: 
\begin{itemize}
    \item Instruction: The description of what you expect the LLM to perform. 
    \item Context: External or contextual information that provides more detailed guidance besides instruction. 
    \item Input Data: The input question we want to have an answer to. 
    \item Output Expectation: The format and requirement of the expected output. 
\end{itemize}

Useful techniques for prompting include zero-shot prompting, few-shot prompting~\cite{brown-etal-2020-language,kaplan-etal-2020-scaling}, chain-of-thought prompting~\cite{wei-etal-2022-chain,zhang-etal-2023-automatic}, automatic or active prompting~\cite{zhou-etal-2023-large,diao-etal-2023-active}. Here we mainly elaborate on the prompting process of chain-of-thought prompting. The following example shows the key idea of chain-of-thought prompting~\footnote{The example follow the same idea of the example provided in the original paper of CoT~\cite{wei-etal-2022-chain}}. Standard few-shot prompting is like:
\begin{center}
\mybox[gray!20]{
Q: A farmer collects 56 eggs from his hens. If he packs them in trays that hold 6 eggs each, how many trays does he need? \\
A: The answer is 10 trays. \\
Q: A classroom has 24 students. If 3 students were absent today, and 2 new students joined the class, how many students are in the class now? 
}
\end{center}
The model gives an incorrect answer. 
\begin{center}
\mybox[blue!10]{
A: The answer is 25 students.
}
\end{center}

However, chain-of-thought prompting is a series of intermediate reasoning steps: 
\begin{center}
\mybox[gray!20]{
Q: A farmer collects 56 eggs from his hens. If he packs them in trays that hold 6 eggs each, how many trays does he need? \\
A: There are 56 eggs. Each tray holds 6 eggs. To find out how many trays, divide 56 by 6, which gives 9 remainder 2. So he needs 9 full trays and 1 more tray for the remainder, making 10 trays. \\
Q: A classroom has 24 students. If 3 students were absent today, and 2 new students joined the class, how many students are in the class now?
}
\end{center}
With the techniques of chain-of-thought prompting, the LLM's output is
\begin{center}
\mybox[blue!10]{
A: The classroom originally had 24 students. With 3 students absent today, there were only 21 students present. However, 2 new students joined, bringing the total back to 23 students in class today. The total number of students enrolled now is 24 (original) - 3 (absent) + 2 (new) = 23 students. The answer is 23 students.
}
\end{center}

This simple technique has seen great success in complex reasoning tasks such as arithmetic reasoning, commonsense reasoning, symbolic reasoning, and so on. 
Due to the page limit, we omit the other topics such as alignment techniques for Large Language Models. 

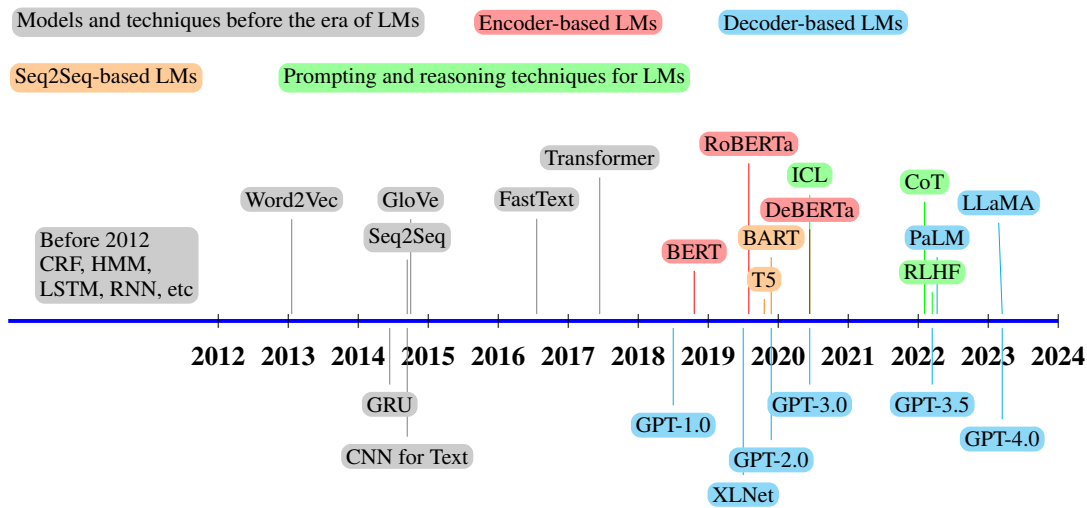
\begin{figure*}  
  \centering
  \resizebox{0.95\textwidth}{!}{
  \usetikzlibrary{decorations.pathreplacing, positioning}

\tikzset{
  event/.style={font=\small, rectangle, rounded corners, fill=#1!40, inner sep=2pt, outer sep=3pt},
  yearlabel/.style={font=\bfseries},
  timeline/.style={ultra thick, blue},
  yeartick/.style={thin, black}
}

\def\beforelm{gray}
\def\encoderlm{red}
\def\seq2seqlm{orange}
\def\decoderlm{cyan}
\def\lmtechniques{green}

\begin{tikzpicture}
  \draw[timeline] (-3,0) -- (12,0); 

  \foreach \year [count=\i] in {2012, 2013, ..., 2024} {
    \node[yearlabel] at (\i-1, -0.5) {\year};
    \draw[yeartick] (\i-1,0.1) -- (\i-1,-0.1);
  }

  \node[event=\beforelm, align=left] at (-1.45,0.8) (before) {Before 2012\\
  CRF, HMM, \\
  LSTM, RNN, etc};


  
  \node[event=\beforelm] at (1.05,1.75) (word2vec) {Word2Vec};
  \draw[\beforelm] (word2vec.south) -- (1.05,0.1);

  \node[event=\beforelm] at (2.70,-1.95) (cnn4text) {CNN for Text};
  \draw[\beforelm] (cnn4text.north) -- (2.70,-0.1);

  \node[event=\beforelm] at (2.45,-1.2) (gru) {GRU};
  \draw[\beforelm] (gru.north) -- (2.45,-0.1);
  \node[event=\beforelm] at (2.75,1.75) (glove) {GloVe};
  \draw[\beforelm] (glove.south) -- (2.75,0.1);
  
  \node[event=\beforelm] at (2.70, 1.2) (seq2seq) {Seq2Seq};
  \draw[\beforelm] (seq2seq.south) -- (2.70,0.1);

  
  \node[event=\beforelm] at (4.55,1.75) (fasttext) {FastText};
  \draw[\beforelm] (fasttext.south) -- (4.55,0.1);

  \node[event=\beforelm] at (5.45,2.35) (transformer) {Transformer};
  \draw[\beforelm] (transformer.south) -- (5.45,0.1);

  \node[event=\decoderlm] at (6.5,-1.50) (gpt1) {GPT-1.0};
  \draw[\decoderlm] (gpt1.north) -- (6.5,-0.1);  
  \node[event=\encoderlm] at (6.8,1.00) (bert) {BERT};
  \draw[\encoderlm] (bert.south) -- (6.8,0.1);
  
  \node[event=\decoderlm] at (7.5,-2.50) (xlnet) {XLNet};
  \draw[\decoderlm] (xlnet.north) -- (7.5,-0.1);  
  \node[event=\encoderlm] at (7.58,2.55) (roberta) {RoBERTa};
  \draw[\encoderlm] (roberta.south) -- (7.58,0.1);
  \node[event=\decoderlm] at (7.9,-2.00) (gpt2) {GPT-2.0};
  \draw[\decoderlm] (gpt2.north) -- (7.9,-0.1);
  \node[event=\seq2seqlm] at (7.9,1.20) (bart) {BART};
  \draw[\seq2seqlm] (bart.south) -- (7.9,0.1);
  \node[event=\seq2seqlm] at (7.8,0.60) (t5) {T5};
  \draw[\seq2seqlm] (t5.south) -- (7.8,0.1);

  \node[event=\decoderlm] at (8.45,-1.20) (gpt3) {GPT-3.0};
  \draw[\decoderlm] (gpt3.north) -- (8.45,-0.1);
  \node[event=\lmtechniques] at (8.45,2.10) (icl) {ICL};
  \draw[\lmtechniques] (icl.south) -- (8.45,0.1);  
  \node[event=\encoderlm] at (8.45,1.60) (deberta) {DeBERTa};
  \draw[\encoderlm] (deberta.south) -- (8.45,0.1);
  
  \node[event=\lmtechniques] at (10.09,2.0) (cot) {CoT};
  \draw[\lmtechniques] (cot.south) -- (10.09,0.1);
  \node[event=\decoderlm] at (10.2,-1.20) (gpt35) {GPT-3.5};
  \draw[\decoderlm] (gpt35.north) -- (10.2,-0.1);
  \node[event=\decoderlm] at (10.27,1.20) (palm) {PaLM};
  \draw[\decoderlm] (palm.south) -- (10.27, 0.1);
  \node[event=\lmtechniques] at (10.2,0.7) (rlhf) {RLHF};
  \draw[\lmtechniques] (rlhf.south) -- (10.2,0.1);  
  
  \node[event=\decoderlm] at (11.15,1.70) (llama) {LLaMA};
  \draw[\decoderlm] (llama.south) -- (11.2,0.1); 
  
  \node[event=\decoderlm] at (11.2,-1.70) (gpt4) {GPT-4.0};
  \draw[\decoderlm] (gpt4.north) -- (11.2,-0.1);  

\node[event=\beforelm] at (0.0,4.30) (beforelmlegend) {Models and techniques before the era of LMs};
\node[event=\encoderlm] at (5,4.30) (encoderlmlegend) {Encoder-based LMs};
\node[event=\decoderlm] at (8.5,4.30) (decoderlmlegend) {Decoder-based LMs};
\node[event=\seq2seqlm] at (-1.6,3.5) (seq2seqlmlegend) {Seq2Seq-based LMs};
\node[event=\lmtechniques] at (3.8,3.5) (lmtechniqueslegend) {Prompting and reasoning techniques for LMs};

\end{tikzpicture}
  }
  \caption{Timeline of major NLP models and techniques for reasoning. We take the publication(or arxiv) time of these works as the date of their appearance(best viewed in color). }
\label{fig:nlp_timeline}
\end{figure*}

\section{Causality in Depth: From Concepts to Theories} \label{appendix:concepts}
When we mention uncertainty in \S~\ref{sec:taxonomy:uncertainty} and contextual nuance in \S~\ref{sec:future}, it involves the understanding of causal sufficiency and necessity~\cite{nadathur-lauer-2020-causal}, which we detail in \Appendix~\ref{appendix:concepts:sufficiency_necessity}. Then, we talk about different theories about causation, which are related to the quantitative measurement of causal strength in \S~\ref{sec:analysis:quantitative} and annotation schemes in \S~\ref{sec:acquisition:manual}. 
\subsection{Causal Sufficiency and Necessity} \label{appendix:concepts:sufficiency_necessity}
\SmallHeading{Causal Sufficiency}
Causal sufficiency refers to a situation where a cause is sufficient enough to lead to a particular effect. Namely, if a cause is present, it is capable of bringing about the effect on its own, without the need for any additional causes or factors.

\SmallHeading{Causal Necessity}
Causal necessity refers to the idea that a cause is necessary for the occurrence of a specific effect. Namely, without the cause, the effect would not happen. 

\subsection{Three Levels of Causality Based on Causal Sufficiency and Necessity} \label{appendix:concepts:three_levels}
From the perspectives of causal sufficiency and necessity, there are three levels of causality~\cite{volchok-2015-three}.  

\SmallHeading{Absolute Causality} Absolute causality refers to the fact that the cause is necessary and sufficient to the happening of the effect. Typically, this kind of causality is only found in the physics domain, such as ``Water flows from high places to low places on Earth''.  However, causality in other domains like law, economics, ecology, social science, and so on is not absolute. For example, ``the increase of carbon dioxide emission'' leads to ``rise of sea level''. Most causality falls in the categories of \textit{conditional causality} and \textit{contributory causality}.

\SmallHeading{Conditional Causality} Conditional causality suggests that a cause is necessary, but not sufficient to trigger an effect. For instance, receiving a vaccine is a necessary but not sufficient cause for preventing a particular disease. 

\SmallHeading{Contributory Causality} Contributory causality means that a cause is not the primary or sole cause of the effect. However, combined with other causes, they collaboratively lead to the occurrence of the effect. For instance, the presence of ``a pedestrian jaywalks at a red light'' and ``a high-speed moving car'' contributes to the effect that ``the car hit the pedestrian'', each cause on its own wouldn't lead to the occurrence of the accident.

\subsection{Theories of Causation} \label{appendix:concepts:causation_theories}
Generally speaking, there are three theories related to causation: 
\begin{itemize}
    \item Counterfactual Theory~\cite{lewis-1973-counterfactuals,menzies-beebee-2001-counterfactual}: The basic idea of the counterfactual theory of causation is that the causation between the cause and the effect can be supported by the argument that ``Had \textit{cause} not occurred, \textit{effect} would not have occurred''. The counterfactual theory is a revision of the regularity theory of causation~\cite{andreas-guenther-2021-regularity}.  
    \item Probabilistic Theory~\cite{skyrms-1981-causal,eells-1991-probabilistic,hitchcock-1997-probabilistic}: Probabilistic causation theories assume that causes change the probabilities of their effects. They use the tools of probability theory to characterize the relationships between causes and effects. 
    \item Force Dynamics Model~\cite{talmy-1988-force}: This theory decomposes the causation into various types of force formulations including the imparting of force, resistance to force, overcoming resistance, and removal of a force. 
    It analyzes causing into finer primitives including ``letting'', ``hindering'',  ``helping'', and ``intending''~\cite{wolff-thorstad-2017-force}. This force dynamic approach accounts for cognitive causation phenomena including causation in language, causal chains, cognitive causation, etc.  
    It is the theoretical background for CEP annotation schemes~(\Appendix~\ref{appendix:acquisition:annotation}). 
\end{itemize}
These theories provide the theory ground for annotations, acquisition, and reasoning of causality.  

\section{Linguistic Causality} \label{appendix:linguistic}
The acquisition methods for causality in \S~\ref{sec:analysis} need linguistic expertise.For instance, the linguistic pattern matching methods need linguistic expertise in causal connectives and verbs, which we elaborate on in \Appendix~\ref{appendix:linguistic:connectives} and \ref{appendix:linguistic:verbs} respectively. 

\subsection{Explicit Causal Connectives} \label{appendix:linguistic:connectives}
Penn Discourse Treebank~(PDTB)~\cite{webber-etal-2019-penn} focuses on the annotation of discourse cues and does not postulate any structure constraints on discourse relations. There are 30 explicit causal connectives under the label of \textit{Contingency.Cause.Result} in PDTB. The explicit causal connectives include \textit
{accordingly, and, as, as a consequence, as a result,
as it turns out, as such, because of that, but, consequently, 
finally, for that reason, furthermore, hence, in fact, 
in other words, in response, in short, in the end, indeed, 
so, so as, so that, that is, then, 
therefore, thus, thus being, to this end, ultimately}. 
\subsection{Causation Verbs} \label{appendix:linguistic:verbs}
We summarize the most common verbs that indicate causation~\cite{martin-2018-time} in Table~\ref{tab:verbs}. We classify these causal verbs into different types based on their causal roles including
\begin{itemize}
    \item Direct: It means the subject directly leads to or causes the occurrence of the object. Examples are ``cause'', ``lead to'', etc.  
    \item Preventative: It means that the subject prevents or stops the happening of the object. Typical examples include verbs like ``prevent'', ``stop'', and ``prohibit''. 
    \item Facilitative: This indicates that the subject facilitates or contributes to the occurrence of the effect. Examples are ``enable'', ``faciliatate'', etc. 
    \item Consequential: These verbs emphasize on the consequences of the subject's action brings to the object. Examples includes ``drive'', ``force'', etc. 
    \item Influential: These influential verbs imply that the subjects influence the object. Examples include ``influence'', ``swing'', ``sway'', etc. 
\end{itemize}

\newcommand{\fwidth}{2.3}
\newcommand{\swidth}{5.3}
\newcommand{\eheight}{-6}
\setlength{\extrarowheight}{1mm} 

\begin{table}[htb!]
  \centering
  \resizebox{0.50\textwidth}{!}{
  \begin{tabular}{p{\fwidth cm}p{\swidth cm}}
    \toprule
    \toprule
     \textbf{Type} & \textbf{Examples} \\
    \midrule \midrule
    Direct & \parbox{\swidth cm}{\vspace{\extrarowheight} Make:  Hard work makes success.\\ 
                       Cause: The drugs tend to cause slight drowsiness. \\ 
                       Lead to: A poor diet will lead to severe illness.  \\[\eheight pt]} \\\hline
    Preventative & \parbox{\swidth cm}{ \vspace{\extrarowheight} Prevent: The vaccine prevents the disease. \\ 
                           Stop: The medicine stops the pain. \\ 
                           Prohibit: The law prohibits the liquor trade. \\[\eheight pt]} \\ \hline
    Facilitative & \parbox{\swidth cm}{ \vspace{\extrarowheight} Enable: The software enables the \\team to process the data three times faster, speeding up the project's progress.  \\ 
                           Facilitate: The peace talks facilitated a ceasefire, reducing the hostilities between these two countries that have hated each other for a long time.  \\ 
                           Assist: The teachers' detailed slides assist the students in grasping mathematical concepts. \\[\eheight pt] } \\ \hline
    Resultative & \parbox{\swidth cm}{\vspace{\extrarowheight} Drive: Ambition drives people to succeed in their careers. \\ 
                           Force: The strong winds forced the tree to topple. \\ 
                           Compell: Peer pressure compels one to follow group norms. \\[\eheight pt] } \\ \hline
    Influential & \parbox{\swidth cm} {\vspace{\extrarowheight} Influence: His family background influences his career choice. \\ 
                           Swing: The judge's final decision swung the verdict in favor of the defendant.  \\ 
                           Sway: The compelling argument presented by the defense swayed the jury. \\[\eheight pt]} \\ \hline
    \bottomrule
  \end{tabular}
  }
  \caption{Taxonomy of causation verbs and examples.}
  \label{tab:verbs}
\end{table}

\section{Preliminary of Causal Inference} 
 \label{appendix:causal_inference}
As mentioned in \S~\ref{sec:analysis:qualitative}, causal inference is a useful tool for neuro-symbolic approaches. 
In this section, we give a brief preliminary of causal inference, including key concepts~(\Appendix~\ref{appendix:causal_inference:concepts}), tools~(\Appendix~\ref{appendix:causal_inference:tools}), and structural causal causal models~(SCM, \Appendix~\ref{appendix:causal_inference:scms}).

\subsection{Concepts} \label{appendix:causal_inference:concepts}
In this subsection, we introduce some fundamental concepts regarding causal inference, which include the distinction between causation and correlation, complex causality structures like confounder and collider, and causal modeling concepts such as D-separation, average treatment effect, etc. These concepts are crucial for understanding the intricate nature of cause-and-effect relationships and for developing a robust framework for causal analysis.

\SmallHeading{Correlation and Causation} 
We may all heard the maxim that ``correlation does not imply causation''~\cite{rohrer-2018-thinking}. We present several examples to better illustrate this concept: 
\begin{itemize}
    \item Ice Cream Sales and Drowning Incidents: There is a positive correlation between ice cream sales and the frequency of drowning incidents. These two kinds of events share the confounding factor \textit{hot weather}. Namely, hot weather drives more people to buy more ice cream and swim, which leads to more drowning incidents. 
    \item Stork Population Around Cities and the Deliveries Outside City Hospitals~\cite{hofer-etal-2004-new}: There is a positive correlation between the number of storks in the German state of Lower Saxony between 1970 and 1985 and the deliveries in that area. Both human birth rates and the population of storks are influenced by the availability of suitable nesting areas in rural areas.
    \item Margarine Consumption and Divorce Rates: The correlation between the per capita consumption of margarine and the diverse rate in Maine State is spurious. This is more coincidence rather than correlation, as shown in \cite{vigen-2015-spurious}. 
\end{itemize}

From these examples, we can tell that correlation suggests a potential causal relationship but does not establish a deterministic one. To determine a causal relationship, spurious correlation, controlled experiments, and confounding variables should be considered. 

\SmallHeading{Confounder} 
Confounders or confounding variables are external factors affecting both the cause and the effect, which is one of the major reasons for false causal inferences between an independent variable and a dependent variable. The following are several examples of confounding variables in causal inference: 
\begin{itemize}
    \item The causal relationship between vaccination status and infection rate may be overridden by a confounder of health awareness. Individuals who are more mentally aware of health are more likely to get vaccinated and take other activities to reduce infection rates. 
    \item The causal relationship between college degree attainment and lifetime earnings is spurious due to the confounder of family background. The background of a wealthy family increases both the likelihood of college degree attainment and high earnings due to networking and social capital. 
    \item The correlation between social media usage and depression levels may be confounded by the factor of age. Younger people tend to use social media more and are vulnerable to depression.  
\end{itemize}

\SmallHeading{Collider} By contrast, colliders are variables causal by at least two other variables. Namely, a collider has more than one arrow pointing to it in a causal graph. What follows are several collider examples: 
\begin{itemize}
    \item Hypertension usually is the collider node of genetic predisposition and a high-salt diet. 
    \item The admission to a prestigious university is the collider node of parental socioeconomic background and the quality of primary education.
    \item The state of one's health is the collider node of genetic factors and postnatal factors such as exercise and nutrition.  
\end{itemize}

\SmallHeading{D-Separation} 
D-separation~\cite{geiger-etal-1989-d,hayduk-etal-2003-structural} is a criterion used in a directed acyclic graph (DAG) to determine whether a set of variables X is independent of another set of variables Y, given a third set of variables Z. 
Two (sets of) nodes X and Y are d-separated by a set of nodes Z if all of the paths between (any node in) X and (any node in) Y are blocked by Z.

\SmallHeading{Treatment Group and Control Group}
In a causal study, we usually divide individuals into two groups: the treatment group and the control group. The treatment group receives the intervention or treatment while the control group does not. 

\SmallHeading{Average Treatment Effect} In a causal study, we choose an outcome variable influenced by the treatment. The outcome is measured in both the treatment and control groups. The Average Treatment Effect~(ATE) is defined as the average difference between the outcomes of the control group and the control group. Mathematically, 
\begin{equation}
    \text{ATE} = \mathbb{E}[Y_1 - Y_0]
\end{equation}
where the $Y_1$ is the outcome of the treatment group while $Y_0$ is the outcome of the control group. The expectation is taken over all individuals in the groups. 

\subsection{Tools for Causal Inference Analysis} \label{appendix:causal_inference:tools}
\SmallHeading{Directed Acyclic Graphs~(DAGs)} DAGs provide a structured and visual representation of causal relationships between variables. Nodes in DAGs represent the variables of interest and a directed edge from A to B implies that A causally impacts B. The acyclicity ensures that there are no loops in causality. With DAGs, it is easier to detect special nodes like common ancestral nodes and collider nodes in a complex causal system. Besides, DAGs also help to identify confounding variables and analyze counterfactuals and interventions. 

\SmallHeading{\textit{Do}-Calculus} The do-calculus~\cite{pearl-2000-causality,pearl-2009-causality,pearl-2012-do-calculus,tucci-2013-introduction} is an axiomatic system for replacing probability formulas containing the do operator with ordinary conditional probabilities. 
\textit{Do}-calculus is to estimate the causal effect. Namely,
\begin{equation}
    P(Y \vert do(X=x)). 
\end{equation}
It returns the probability of $Y$ if we intervene $X$ as $x$.  

\subsection{Structural Causal Models} \label{appendix:causal_inference:scms}
\SmallHeading{Structural Equation} 
Before we start, we distinguish two different kinds of variables within a causal system. Endogenous variables are influenced by other variables within the same system.  They are the variables of interest in a structural causal model. Exogenous variables, however, are not influenced by other variables within the same system. Namely,  there is no causal parent for the exogenous variables in the system. In a DAG, exogenous variables are nodes without incoming edges while endogenous variables are nodes with incoming edges. 

Let us suppose that $X$ causally influences $Y$ and $X$ is an exogenous variable. 
The structural equation is
\begin{equation}
    X = U_X
\end{equation}
where $U_X$ is the unobserved factors affecting $X$. 
For endogenous variable $Y$(it is influenced by other variables within the same system), the structural equation becomes
\begin{equation}
    Y \coloneqq f(X) + U_Y
\end{equation}
where $f(X)$ is the causal function between $X$ and $Y$. $U_Y$ represents the unobserved factors $Y$.

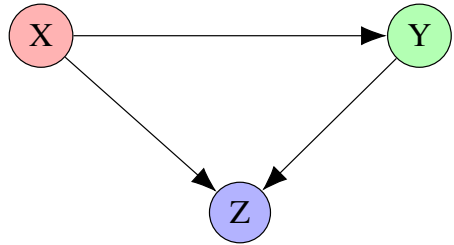
\begin{figure} 
  \centering
  \resizebox{0.8\columnwidth}{!}{
  \begin{tikzpicture}[auto]
  \node (X) [draw, circle, fill=red!30] at (0, 3) {X};
  \node (Y) [right=of X, draw, circle, fill=green!30] at (3, 3) {Y};
  \node (Z) [below=of Y, draw, circle, fill=blue!30] at (2.3, 2.3) {Z};

  \draw[-{Latex[length=3mm]}] (X) -- (Y);
  \draw[-{Latex[length=3mm]}] (Y) -- (Z);
  \draw[-{Latex[length=3mm]}] (X) -- (Z);
\end{tikzpicture}
  }
  \caption{Timeline of major NLP models and techniques for reasoning. We take the publication(or arxiv) time of these works as the date of their appearance(best viewed in color). }
  \label{fig:scm}
\end{figure}

\SmallHeading{Structural Causal Models}
Structural Causal Models~(SCMs)\cite{bollen-pearl-2013-eight,hair-etal-2021-introduction} are mathematical tools to analyze DAGs. 
A SCM consists of a set of exogenous variables~($U$) and a set of endogenous variables~($V$) connected by a set of functions~($F$).  Each SCM is associated with a DAG where a node is a variable in $U$ or $V$ while an edge is a function $f$ in $F$.  
We show an example of a DAG in Figure~\ref{fig:scm}. 

Now structural equations turn into: 
\begin{equation}
    \begin{cases}
        Y = f_{Y}(X) + U_Y \\
        Z = f_{Z}(X, Y) + U_Z
    \end{cases}
\end{equation}
where $f_{Y}$ are causal function between $X$ and $Y$ while $U_Y$ is the unobserved factor affecting $Y$. $f_{Z}$ are causal function between $X$ and $Y$ as a unit and $Z$ while $U_Z$ is the unobserved factor affecting $Z$.

\section{Handbook for Beginners to Learn Commonsense Causality} \label{appendix:handbook}
We believe that a detailed and accessible handbook is invaluable for junior researchers to get familiar with research topics. 
We list step-by-step and comprehensive practices for junior researchers in the domain and hope that it is helpful for this field's future progress.  
\begin{itemize}
    \item First, we kindly suggest researchers interested in this domain start with the background and preliminary knowledge which covers linguistic causality~(\Appendix~\ref{appendix:linguistic}), NLP preliminary including pretrained language models and prompting techniques~(\Appendix~\ref{appendix:lm}), benchmarks of commonsense causality~(\S~\ref{sec:taxonomy} and \Appendix~\ref{appendix:detailed_benchmarks}), the taxonomy of causation~(\S~\ref{sec:taxonomy} and \Appendix~\ref{appendix:taxonomy}), theories of causation~(\Appendix~\ref{appendix:concepts}), probabilistic causation~(\S~\ref{sec:analysis:quantitative}), reading of related surveys~(\Appendix~\ref{appendix:survey}), and causal inference preliminary~(\Appendix~\ref{appendix:causal_inference}).  
    \item Then, there are two branches for selection: causality acquisition~(\S~\ref{sec:acquisition} and \Appendix~\ref{appendix:acquisition}) and reasoning over causality~(\S~\ref{sec:analysis}). For causality acquisition, there are various approaches such as extractive methods~(\S~\ref{sec:acquisition:extraction}), generative methods~(\S~\ref{sec:acquisition:generative}), and manual annotations~(\S~\ref{sec:acquisition:manual}). 
    For causality reasoning, both qualitative reasoning~(causal reasoning and counterfactual reasoning) and quantitative measurement of causal strength~(\S~\ref{sec:analysis:quantitative}, \Appendix~\ref{appendix:uncertainty}) are promising topics for further research.  
    \item Following this, there are some more challenging but significant prospective topics for exploring. We highlight the contextual nuances, complex structures, temporal dynamics, probabilistic approaches, and multimodal data of commonsense causality for future study. 
    Specifically, we first underscore the importance of developing models that address the contextual dependency of cause-effect relationships and intricate causal structures. 
    Additionally, we underline the temporal dynamics of commonsense causality, pointing out the urgency of identifying the most effective timing for interventions and understanding the temporal sequences of causal impacts for future study. 
    Furthermore, we emphasize the necessity of employing probabilistic perspectives to manage uncertainties in commonsense causality. 
    Finally, we outline several challenging topics for multimodal causality analysis including the reasoning of multimodal commonsense causality and the alignment of cross-modal commonsense causality. For a comprehensive discussion of these promising future directions, please refer to Section~\ref{sec:future}.
    \item Finally, for researchers who are more interested in the application side of commonsense causality~(\Appendix~\ref{appendix:applications}), the combination of commonsense causality and fields like medical diagnosis~\cite{richens-etal-2020-improving}, psychology~\cite{matute-etal-2015-illusions,eronen-2020-causal}, behavioral science~\cite{grunbaum-1952-causality}, economics~\cite{bronfenbrenner-1981-causality,hoover-2006-causality}, legal systems~\cite{williams-1961-causation,summers-2018-common} are optimistic topics. 
\end{itemize}

\begin{figure*}[htb]
\begin{adjustwidth}{-0.5in}{-0.5in}
\centering
\resizebox{\linewidth}{!}{\input{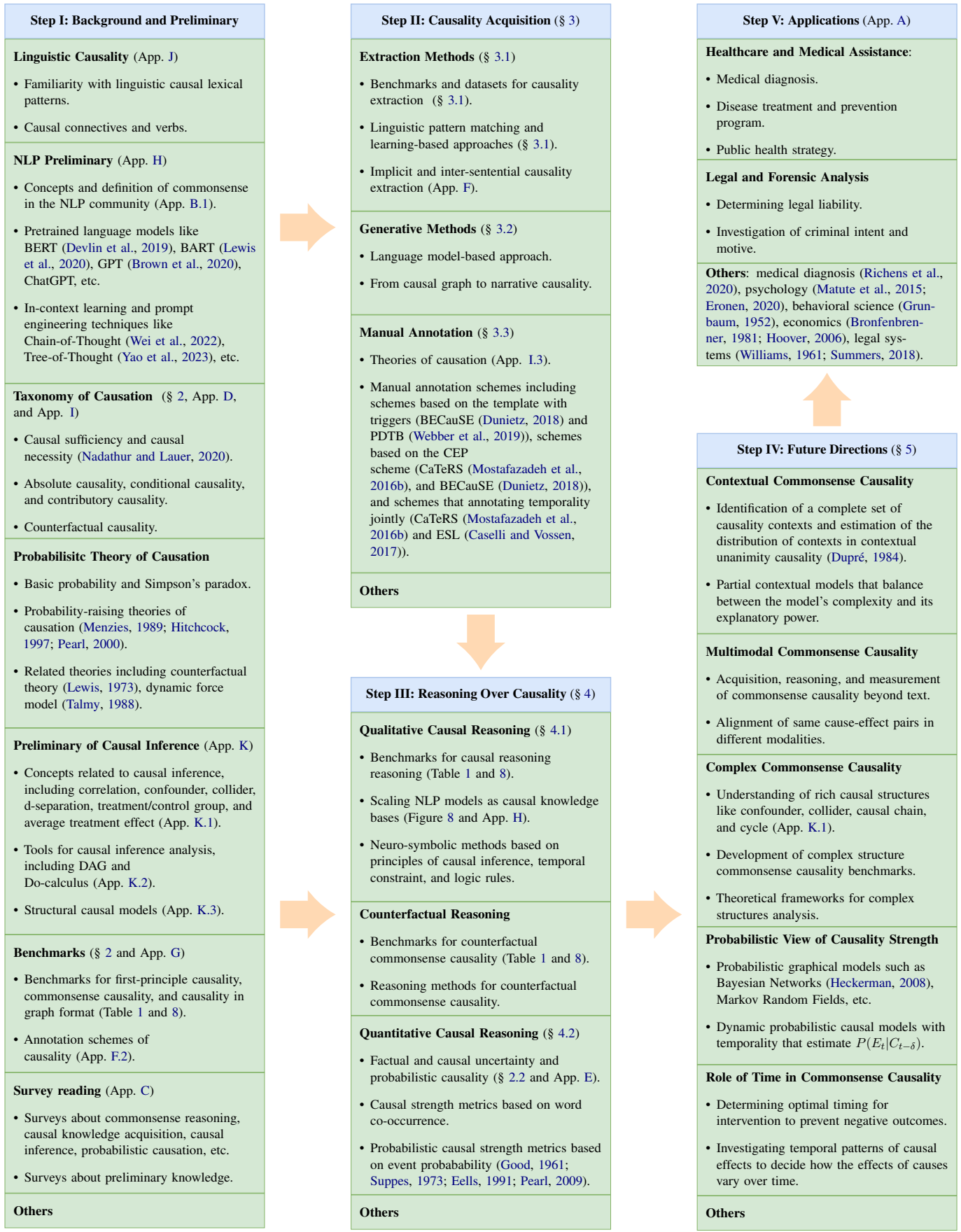}}
\end{adjustwidth}
\caption{Handbook for beginner researchers interested in commonsense causality.}
\label{fig:handbook}
\end{figure*}

More details can be found in Figure~\ref{fig:handbook}. Through this detailed handbook, we hope that the topic of commonsense causality is more accessible to researchers who are interested in this domain. 

\end{document}